\documentclass{LMCS}

\usepackage{calc}         
\usepackage{color}        
\usepackage{graphicx}     
\usepackage{ifthen}       
\usepackage{pst-all}      
\usepackage{pst-poly}     
\usepackage{multido}      
\usepackage{epsfig}

\def\eqalign#1{\null\,\vcenter{\openup\jot\mathsurround=0 pt
  \ialign{\strut\hfil$\displaystyle{##}$&$\displaystyle{{}##}$\hfil
      \crcr#1\crcr}}\,}

\newcommand{\tamara}[1]{{\bf Tamara: #1}}

\newcommand{\smitem}{\item}
\newcommand{\cal}{\mathcal}

\newcommand{\true}[0]{{\it true}}

\newcommand{\aname}[0]{{\cal N}}
\newcommand{\precond}[0]{{\cal P}}

\newcommand{\assertlist}[0]{{\cal A}}

\newcommand{\closure}[1]{#1{}^{\circ}}

\newcommand{\main}[0]{{\cal M}}
\newcommand{\vars}[0]{{\cal V}}
\newcommand{\exc}[0]{{\cal E}}

\newcommand{\exci}[1]{{\cal E}_{#1}}
\newcommand{\subst}[0]{\Sigma}

\newcommand{\bigo}[0]{{\it O}}
\newcommand{\widewedge}[0]{\;\land\;}

\newcommand{\ourexists}[1]{\exists #1\,.\, }
\newcommand{\ourforall}[1]{\forall #1\,.\, }
\newcommand{\ourst}[0]{\,|\,}

\newcommand{\m}[1]{$#1$}
\newcommand{\md}[1]{$$#1$$}

\newcommand{\entails}[0]{\models} 

\newcommand{\psiplan}[0]{{PSIPLAN}} 
\newcommand{\psipop}[0]{{PSIPOP}} 
 
\newcommand{\psipopse}[0]{{PSIPOP-SE}} 
\newcommand{\psigraph}[0]{{PSIGraph}} 
\newcommand{\psiform}[0]{\m{\psi}-form}
\newcommand{\psiforms}[0]{\m{\psi}-forms}

\newcommand{\ourmd}[1]{\md{#1}} 

\newcommand{\I}[1]{I_{#1}}
\newcommand{\ignore}[1]{}

\ignore{

\newenvironment{proof}{{\bf Proof:} }{\hfill\m{\Box}}
}

\newcommand {\image}[2]{Im_{#2}^{#1}}

\newcommand{\mgusubset}[0]{MGU_{\scriptscriptstyle{\subseteq}}} 
\newcommand{\mgu}[0]{MGU_{\scriptscriptstyle{\equiv}}}

\renewcommand {\image}[2]{#1\triangleright{}#2}

\newtheorem{preourdefinition}{Definition}
{\begin{preourdefinition}\rm}{\end{preourdefinition}}

\newenvironment{ourexample}%
{\begin{exa}}{\end{exa}}

\newenvironment{ourlemma}%
{\begin{lem}}{ \end{lem}}

\newenvironment{ourtheorem}%
{\begin{thm}}{{\theend} \end{thm}}

{\begin{cor}}{{\theend} \end{cor}}

\newenvironment{ourobservation}%
{\begin{obs}}{{\theend}\end{obs}}

\usepackage{latexsym}
\newcommand{\theend}[0]{\relax}
\newcommand{\set}[1]{\{{#1}\}} 

\newcommand{\psiset}[1]{[{#1}]} 
\newcommand{\spsiset}[1]{[{#1}]} 
\newcommand{\spsi}[0]{\spsiset} 
\newcommand{\apsi}[2]{[{#1}\mbox{ except }\set{#2}]} 
\newcommand{\setpsi}[0]{\phi} 

\newcommand{\Ms}[0]{{I}} %

\newcommand{\diff}[0]{\mathop{^{\underbar{\small$\,e\,$}}}}

\newcommand{\trans}[0]{Trans}

\newcommand{\norm}[1]{\left\|{#1}\right\|}

\newcommand{\revised}[0]{}
\newcommand{\nb}[1]{#1}

\newcommand{\edit}[2]{{#2}}
\newcommand{\corr}[2]{{#2}}

\newcommand{\mydefi}[1]{{\bf #1\index{#1}}}    

\newcommand{\infruletop}[2]{\begin{array}{c} \underline{#1}\\{#2}\end{array}}

\newcommand{\ccc}[0]{\psi_1^i\in\Psi_1} 

\providecommand{\thrmFour}{
{\bf \psiform{} Entailment.}
  Let \m{\psi_1,\ldots,\psi_n} and \m{\psi} be arbitrary \psiforms{}.

\noindent \m{\set{\psi_1,\ldots,\psi_n}\entails\psi} if and only if there
  exists a \m{k},\m{1\le k\le n}, such that:
\begin{itemize}
\item \m{\spsi{\main(\psi_k)}\entails\spsi{\main(\psi)}} (i.e.,
  the main part of \m{\psi_k} entails the main part of \m{\psi}), and
\item
  \m{\set{\psi_1,\ldots,\psi_{k-1},\psi_{k+1},\ldots,\psi_n}\entails
    \psi\diff\psi_k}.
\end{itemize}
}

\providecommand{\thrmFive}{
  Let \m{\psi_1} and \m{\psi_2} be simple fixed length \psiforms{}.
  \md{\hbox to 89 pt{\hfil}{\image {\psi_1} {\psi_2}} =
    \set{\spsi{\main(\psi_2)\sigma}\ourst \sigma\in {
        \mgusubset(\main(\psi_1),\main(\psi_2))}}\hbox to 88 pt{\hfil}}
}

\providecommand{\thrmSix}{
\begin{sloppypar}
    Let \m{\psi_2} be an arbitrary fixed length \psiform{} and
    \m{\psi_1} be a simple fixed length \psiform{}.  \md{ \psi_2 \diff
      \psi_1 = \left\{\begin{array}{ll}
          \emptyset, &\mbox{if \m{\psi_1\entails\psi_2},}\\
          \set{[\main(\psi_2)\mbox{ except } \Sigma(\psi_2)\cup \Sigma']},
          &\mbox{otherwise,}\\
                             \end{array}
                           \right.  }
 where \nb{\m{\Sigma' = \set{\sigma'\ourst\sigma'=Trans (\sigma, \psi_2),\textit{  where } \sigma\in\mgusubset(\main(\psi_1), \main(\psi_2))}}.} 
(Recall that \m{Trans (\sigma, \psi_2)} defined in Figure \ref{fig:trans} transforms substitution \m{\sigma} to an equivalent one that conforms to the format of exceptions of \m{\psi_2}.)
\end{sloppypar}
}

\providecommand{\thrmSeven}{
  If \m{\mgusubset(\main(\psi_1), \main(\psi_2)) = \emptyset},
  \m{\image{\psi_1}{\psi_2} = \emptyset} and \m{\psi_2 \diff \psi_1 =
    \set{\psi_2}}.
}

\providecommand{\thrmEight}{
When  \m{\spsi{\main(\psi_1)}\entails\spsi{\main(\psi_2)}}
\begin{equation}\label{th:case2Img}
{\image {\psi_1} {\psi_2}} = \psi_2 - H(\psi_1, \psi_2)
\end{equation}
\begin{equation}\label{th:case2Diff}
{{\psi_2}\diff {\psi_1}} =  H(\psi_1, \psi_2) - \exc(\psi_2)
\end{equation}
}
\providecommand{\thrmNine}{
Let \m{\mgusubset(\main(\psi_1),\main(\psi_2))\neq\emptyset}, and let
\m{\Psi} denote the image
\m{\image{\spsi{\main(\psi_1)}}{\spsi{\main(\psi_2)}}}. Then  
\begin{equation}
{\image {\psi_1} {\psi_2}} =({\image {\psi_1} {\Psi}}) - \exc(\psi_2)
\end{equation}
\begin{equation}\label{th:FLDiff}
{{\psi_2}\diff {\psi_1}} = (\psi_2 - \Psi) \cup [(\Psi  - \exc(\psi_2)) \diff {\psi_1}]
\end{equation}}

\providecommand{\thrmTen}{
  Image and e-difference of two fixed length \psiforms{} is equivalent
  to a finite set of fixed length \psiforms{}.
}

\providecommand{\thrmEleven}{
Let  \m{a} be a  domain action and  \m{s} be an agent's SOK before
executing \m{a}, where \m{s} is satisfiable and saturated, and
 \m{a} is executable in \m{s}.
Then the following state of knowledge update rule   
\begin{equation} \label{update}
update(s,a) = ((s \diff \assertlist^-(a)) \diff {\assertlist}(a)) \cup
\assertlist(a),
\end{equation}
is correct and complete. Moreover, the resulting SOK \m{s' =
update(s,a)} is saturated. It is minimal if \m{s} is minimal.
}

\providecommand{\lemmaOne}{When \m{\psi} is a fixed length \psiform, there is a unique \m{\sigma} for each clause \m{c\in\psi} such that \m{\main(\psi)\sigma=c}.}

\providecommand{\lemmaTwo}{Given two non-empty ground clauses of negated literals, \m{c_1} and  \m{c_2}, \m{c_1\entails c_2} if and only if \m{c_1\subseteq c_2}.}

\providecommand{\lemmaThree}{\m{ {\image{\psi_1}{\psi_2}} = (({\image {\spsi{\main(\psi_1)}}      {\spsi{\main(\psi_2)}}}) - H(\psi_1,\psi_2) ) - \exc(\psi_2) }\,.}

\providecommand{\lemmaFour}{ \m{ \psi_2 \diff \psi_1 = (({\spsi{\main(\psi_2)}} \diff    {\spsi{\main(\psi_1)}}) \cup H(\psi_1,\psi_2)) - \exc(\psi_2) }\,.}

\providecommand{\lemmaFive}{ Let \m{\Theta   =\mgusubset(\main(\psi_1),\main(\psi_2),\vars(\psi_1))}
be nonempty. 
Let   \m{\Psi_1} be defined as follows.
\begin{equation}
{\Psi_1 =  \set{\psi_1^i\ourst\psi_1^i =
\spsi{\main(\psi_1)\theta_i},  \theta_i \in \Theta, { 1\leq
    i\leq{\norm{\Theta}} }}}.
\end{equation}
Then, 
\begin{equation}\label{th:case2holes}
H(\psi_1, \psi_2) = \bigcap_{\ccc}\bigcup_{j=0}^{\norm{\Sigma(\psi_1)}} ({\image
{(\spsi{\exci{j}(\psi_1)}\cap \psi_1^i )}
{\spsi{\main(\psi_2)}}}) , 
\end{equation}
}

\providecommand{\lemmaSix}{
Assume \m{\spsi{\main(\psi_1)}\entails\spsi{\main(\psi_2)}} and assume
that none of the exception forms of \m{\psi_1} entails
\m{\spsi{\main(\psi_2)}}.  Then each \psiform{} in \m{\psi_2\diff\psi_1} uses
strictly fewer variables than there are in \m{\psi_2}.
}

\providecommand{\lemmaSeven}{
  Procedure {\bf Saturate}\m{(s)} returns a SOK \m{s_0} that is
  a saturated equivalent of \m{ s}, if \m{  s} is consistent,
  or {\bf fail}, otherwise. 
\edit{A-6}{Assuming \m{s} consists of \m{n} fixed length \psiforms{} and \m{m} atoms, the time complexity of procedure {\bf Saturate(s)} is \m{\bigo(nm^C)}}, where C denotes the maximum length of a \psiform{} clause.
}
\providecommand{\lemmaEight}{
Let \m{\psi_1} and  \m{\psi_2} be fixed
length \psiforms{}. The number of subset-matches of  \m{\main(\psi_1)} onto
\m{\main(\psi_2)} is bounded by \m{e^{C/e}},
i.e. \m{\norm{\mgusubset(\main(\psi_1), \main(\psi_2), \vars(\psi_1))}
\leq(e^{C/e}) }.  
}

\def\doi{2 (3:5) 2006}
\lmcsheading%
{\doi}
{1--39}
{}
{}
{Jan.~21, 2005}
{Sep.~26, 2006}
{}
\begin{document}

\title[Efficient Open World Reasoning for Planning]{Efficient Open World Reasoning for Planning}

\author[T. Babaian]{Tamara Babaian\rsuper a}
\address{{\lsuper a}Department of Computer Information Systems, 
  Bentley College, Waltham, MA 02452  USA}
\email {tbabaian@bentley.edu}

\author[J. G. Schmolze]{ James G. Schmolze\rsuper b}
\address{{\lsuper b}Department of Computer Science, Tufts University,  
  Medford, MA 02155  USA }

\keywords{Knowledge representation and reasoning, planning with
  incomplete information.}
\subjclass{I.2.4; I.2.8; F 4.1; F.2.2}

\begin{abstract}
  We consider the problem of reasoning and planning with incomplete
  knowledge and deterministic actions.  We introduce a knowledge
  representation scheme called \psiplan{} that can effectively
  represent incompleteness of an agent's knowledge while allowing for
  sound, complete and tractable entailment in domains where the set of
  all objects is either unknown or infinite.  We present a procedure
  for state update resulting from taking an action in \psiplan{} that
  is correct, complete and has only polynomial complexity. State
  update is performed without considering the set of all possible
  worlds corresponding to the knowledge state. As a result, planning
  with \psiplan{} is done without direct manipulation of possible
  worlds.  \psiplan{} representation underlies the \psipop{} planning
  algorithm that handles quantified goals with or without exceptions
  that no other domain independent planner has been shown to
  achieve. \psiplan{} has been implemented in Common Lisp and used in
  an application on planning in a collaborative interface.
\end{abstract}

\maketitle

\section{Introduction}

Much progress has been made in the area of planning with 
a correct but  incomplete description of the world,  i.e. an {\em open world}.
However, so far there have been relatively few formalisms proposed for efficient  open world representation and reasoning  in the
domains with a very large or unknown set of individual objects.   
This paper describes a representation, called \psiplan{},  for planning with incomplete information about the initial state, that handles open world planning problems that  have not been shown to be solved by any other implemented domain
independent planning system.   

For an example of an open world problem, consider a robot operating in a
warehouse that is given a goal of delivering box A to the front door. The
robot can pick up only those boxes that do not contain fragile items.  To
pick up box A the robot does not need to know the location of all fragile
items, nor the precise contents of box A. It is sufficient to know that A has
no fragile items. Thus, in such situations, if the robot knows that (a) {\em
  no containers have fragile goods except for B}, it can safely pick up box
A. When the list of all containers that the robot might eventually have to
reason about is unknown, it is impossible to represent (a) by enumerating all
containers that do not have fragile goods. Thus to include (a) in the robot's
description of the state, or to represent a goal such as {\em there is
  nothing at the front door except for box A}, a quantified statement is
necessary.

The robot must be able to update quickly and correctly its knowledge of the
state that results from performing an action.  For instance, after a new
container C, whose contents are unknown, is added to the warehouse, the robot's
updated state must reflect that (b) {\em there are no containers with fragile goods  except for  B and  possibly C.}

Effective and efficient operation in a partially known environment 
may require the ability to satisfy {\em knowledge goals}, such as for
example, (c) {\em identify all containers with fragile goods that will be shipped
  to Boston}, and {\em sensing} actions that return information about the
world, such as examining the contents of a box, or determining a box's
destination from its label.
\edit{}{A robot that is able to reason in a sound and complete manner about its
knowledge and lack of knowledge is better equipped to handle such knowledge goals.}
For instance, knowing only (b) plus that C's
destination is Chicago, when given a knowledge goal (c), the robot must be
able to conclude that the only necessary information that is missing is the
destination of B. Such precision of reasoning
ensures that
sensing is non-redundant and relies on sound and complete
reasoning in the underlying representation.



As this example demonstrates, a representation used in an
open-world application must distinguish between
propositions whose truth value the agent knows
and those whose truth value it does not know.  Merely listing all facts
marked with their truth value is inefficient when the set of all domain
objects is very large because, typically, the number of atoms whose value is
known to be {\em false} is large.  For example, the set of all objects that
are {\em not} in box A could be very large.
Furthermore, such enumeration of known atoms is impossible when the
set of all domain objects is infinite or unknown.
A key requirement of an open world planning representation with partially
unknown or infinite domains is to allow quantification for (at least) negative
information while retaining efficient reasoning.

In this paper we present a representation for open world reasoning that uses
a class of quantified sentences with exceptions that we call \psiforms{}.
\psiforms{} represent quantified negative information such as the following statement:  {\em there are no containers with fragile goods except for possibly B and  C.} 
We present an efficient calculus for {\em \psiforms{}}, that includes a
sound and complete entailment procedure that takes polynomial time
\edit{B0}{under certain reasonable assumptions on the structure of
\psiforms{}. }

Further, we present a formalism called \psiplan{}, based on
\psiforms{}, for reasoning and planning in partially known worlds. 
\psiplan{}  offers a unique combination of tractability, 
expressivity and completeness of reasoning.
This makes it uniquely suited for applications where the
set of all domain objects can never be acquired, and the 
ability to generate and explore alternative plans 
while avoiding redundant execution is critical. 
One of the advantages of  \psiplan{} compared to
other representation used in open world planners, is the fact that
all reasoning about actions and states in \psiplan{} is
carried out without an explicit manipulation of the set of possible
worlds. This property reduces a planner's sensitivity  to the amount of 
irrelevant information, which causes some planners that use 
explicit enumeration of the possible worlds to blow up.

A  partial order planning algorithm (\cite{mcallester-rosenblitt-91}) called \psipop{} based on \psiplan{}
is presented in \cite{babaian-schmolze-00}. \psipop{} is sound and
complete for the planning domains in which the set  of all domain
objects is infinite or can never be fully acquired. \corr{1.12}{As a partial order planner, \psipop{} produces a solution in a form of a partially ordered set of steps, which can be linearlized or used as a basis for a parallelized plan.}  The goal language
of \psipop{} includes quantified goals with exceptions such as (b)
that no other implemented domain independent planner has been shown to
handle.  

An extension of \psiplan{} that includes sensing actions and knowledge goals is presented in \cite{babaian-thesis}.
 \edit{A(03)}{A planner that uses the extended representation for 
planning with sensing and interleaved execution (\psipopse) was used in a
collaborative interface called Writer's Aid \cite{waid-iui}. 
Writer's Aid helps an author writing an academic manuscript by simultaneously
and autonomously identifying, locating  and  downloading relevant
bibliographic records and papers from the author's preferred local and
Internet sources.} The use of \psiplan{} at the core of Writer's Aid
ensures the expressiveness of its goal language, and its ability to precisely identify the missing information and never engage in redundant search, while exploring all possible courses of action. 

In this paper we make the following contributions:
\begin{itemize}
\item We present a complete description of the language of \psiforms{}
and \psiform{} calculus, and report on the complexity of the calculus
operations.  The \psiform{} calculus is an integral and critical part of  \psiplan{} reasoning, and, thus, its algorithms, complexity and completeness properties  bear direct effect on the properties of \psiplan{}-based planners (\cite{babaian-schmolze-00,babaian-thesis,waid-iui}, as well as a  Graphplan-style (\cite{blum-furst-95}) conformant planner  currently under development.)  

\item We introduce \psiplan{} representation of an agent's incomplete state of
knowledge and illustrate it with examples.
We further present \psiplan's  action language and a procedure for
state update after an action, and prove important properties of the
update procedure, including its completeness. 

\end{itemize}

\ignore{

This tool helps an author writing an academic manuscript by simultaneously
and autonomously  
identifying, locating  and  downloading relevant bibliographic records
and papers from the author's preferred local and Internet sources.

The richness of the \psiplan{} language permitted formulation and
achievement of expressive goals like: 
\begin{quotation}
\noindent For each paper containing keywords {\em incomplete} and
{\em knowledge} and listed in some bibliographic source 
preferred by the user, download a bibliographic record and a viewable
version of that paper. 
\end{quotation} 
Planning to satisfy such goals relies on \psiform{} reasoning techniques presented in this paper.
Completeness of the \psiplan{} reasoning and planning techniques were
critical for implementing a number of Writer's Aid's design features,
most notably  in ensuring Writer's Aid ability to generate and explore
alternative plans of achieving a goal while avoiding redundant execution of 
time consuming Internet searches in the setting where almost no prior
information is available about any papers, their locations or
authors.  
%
%
}

\subsection{Prior Approaches}

The lack of universally quantified reasoning in open-world planners
(e.g.\cite{brafman-hoffman-04}, \cite{baral-son-97}, \cite{eiter-etal-00},\cite{smith-weld-98}, \cite{weld-etal-98}, \cite{cimatti-roveri-99},
\cite{ambros-ingerson-steel-88}, \cite{krebsbach-etal-92},
\cite{peot-smith-92},) precludes their use  in domains in which the set of
all objects is not known, very large, or infinite (\cite{weld-99},
\cite{etzioni-etal-97}).

On the other hand, situation calculus-based approaches (e.g. \cite{reiter-00},
\cite{scherl-levesque-93}, \cite{levesque-96}) have the expressivity of full
first order logic (FOL), and thus admit planning problems with arbitrary
quantified formulas. However, a complete planner based on the unrestricted situation  calculus , i.e. that relies upon the full FOL, is impossible due to the 
undecidability of entailment in FOL. \corr{2}{Recently, Liu and Levesque have presented a subset of situation calculus, with a tractable, sound and complete action projection under certain restrictions (\cite{liu-levesque-05}). Other related approaches to reasoning about actions incorporating first-order features are presented in \cite{thielscher-05} and  \cite{shirazi-amir-05}.  These works are reviewed in Section \ref{sec:related}.}
\ignore{
 Although there exist decidable subsets of FOL, the situation calculus is not one of them, and to the best of our knowledge, no decidable subset of situation calculus suitable for sound and complete open world planning has so far been presented.}

The LCW (for Locally Closed Worlds) language of Etzioni et. al.
\cite{etzioni-etal-97} is designed to achieve expressivity {\em and}
tractability for open world reasoning and is most closely related to
\psiplan{}.  LCW sentences specify the parts of the world for which the agent
has complete information.  It does this by collecting formulas
\m{\Phi} where the agent knows the truth value of every ground instance of
\m{\Phi}.  For example, if the agent knows all fragile items that are in box \m{B},
it states \m{LCW(In(x,B)\land Fragile(x))}. Combined with a propositional database that
states that \m{In(\textit{Vase}, B), Fragile(\textit{Vase})}, the agent can conclude that nothing is fragile in \m{B} except for the \m{\textit{Vase}}.

LCW reasoning, although tractable, is incomplete
\cite{etzioni-etal-97,babaian-schmolze-00}.  There are two sources of
incompleteness in LCW: (1) incompleteness of inference and (2) inability of
LCW statements to represent {\em exceptions}, i.e., the inability to state
that the agent knows the value of all instantiations of formula \m{\Phi} {\em
  except} some. \edit{A}{These are the key difference between the LCW
and  \psiplan{} representations;  \psiplan{} reasoning is complete, and \psiplan{} can also express what exactly is {\em not known} via the \psiform{} exceptions mechanism. Adding a similar mechanism to the LCW framework would require the development of new entailment procedures, methods for state update and operations underlying the planning techniques akin to those developed in this paper.}   

\edit{A1-2}{As a result of the lack of exceptions in LCW
sentences,  when one or more instances of
\m{\Phi} is unknown, \m{LCW(\Phi)} cannot be stated. This limitation
on the expressive power will sometimes cause known information to be
discarded from the LCW knowledge base upon updating it after an
action, even when the effects of the action are completely specified,
 do not cause information loss, and create no new objects.
For example, consider
the result of moving an object  \m{\textit{Cup}} from some other box
to box \m{B} in the situation where \m{LCW(In(x,B)\land Fragile(x))} is asserted.  If it is not known 
whether the \m{\textit{Cup}} object is fragile, the LCW statement
above no longer holds and thus must be discarded, effectively discarding 
from the knowledge base all instances of \m{In(x,B)\land Fragile(x)}
that are known to be \m{false}.}  

\edit{A}{
\psiplan{} subsumes a large part of the LCW language. 
Every knowledge state represented by an LCW-based representation can also be
represented in \psiplan{}, except for those situations which require
an LCW statement \m{LCW(\Phi)}, where \m{\Phi}
contains atoms that unify when all variables are renamed to be
distinct\footnote{These limitations are related to the definition
of fixed length \psiforms{} presented in Section \ref{sec:rep} and
are critical to the tractability of \psiform{} reasoning.}. On the
other hand,
there are states
of knowledge that \psiplan{} can represent accurately, but LCW cannot. }

\ignore{  
XII \cite{golden-etal-94} and PUCCINI \cite{golden-thesis} are causal link
planners based on the LCW representation.  They incorporate sensing actions
and interleaved execution, and thus handle a larger class of problems than
those addressed here.  }
Both LCW and \psiplan{} representations can be used in
planning with sensing \cite{babaian-thesis,waid-iui}. Since LCW's reasoning is
incomplete, however, planners based on LCW are inherently incomplete.  To help remedy
this, LCW based planners use sensing actions to find out facts they
cannot infer, but this only works when an appropriate sensing action is both
available and not too costly.  In general, there is no effective replacement
for sound and complete reasoning.

The LCW \cite{etzioni-etal-97} representation is  extended in
\cite{levy-96,friedman-weld-97} to handling exceptions. However, 
both of these works only consider the setting in which there are no 
actions that can change the world, do not address a changing
world or planning, and do not present any methods that would 
make these extensions amenable to their use in reasoning about actions.

\subsection {A brief look at \psiplan{}} \label{sec:example}

\begin{ourexample}\label{ex:coml-example}
A robot operating in a warehouse  is told that there are
no fragile goods in any of the boxes except possibly for the
box marked {\it FragileStuff}, i.e. 
\begin{equation}\label{eq:fol-init}
\ourforall{g,c} \neg Box(c) \vee \neg In(g,c) \vee \neg
Fragile(g) \vee c = \textit{FragileStuff} 
\end{equation}
Here \m{Box(c)} states that \m{c} is a box, \m{In(g,c)} states that
item \m{g} is in container \m{c}, and \m{Fragile(g)} states that \m{g}
is fragile. Note that (\ref{eq:fol-init}) does not state whether or not \m{\textit{FragileStuff}}
actually contains any fragile items.

In \psiplan{},  statement (\ref{eq:fol-init}) is represented
by the following \psiform{} that represents a conjunction of all ground clauses
\footnote{A clause is a disjunction of literals. A clause is ground if it contains no variables.}   
that can be obtained by instantiating  the formula \m{\neg Box(c) \vee \neg
In(g,c) \vee \neg Fragile(g)}  in all
possible ways except for  instantiations in which
\m{c=\textit{FragileStuff}}. 
This is written as the \psiform{}   
\begin{equation}\label{eq:intro-psi}
\psi={\apsi{\neg Box(c) \vee \neg In(g,c) \vee \neg Fragile(g)}{\set{c=\textit{FragileStuff}}}}.
\end{equation}
whose interpretation is exactly the same as (\ref{eq:fol-init}).  Here and below, lowercase letters in \psiforms{} denote implicitly universally quantified variables, while symbols that start with a capital letter denote constants.

Suppose, it is also known that a bottle of wine  is the only fragile item. Consequently, it is in the {\it FragileStuff}
box,  i.e. 
\begin{equation}\label{eq:fol-init-atoms}
Box(\textit{FragileStuff}), In(Wine,\textit{FragileStuff}), Fragile(Wine) 
\end{equation}

With \psiplan{}, the original situation comprised by (\ref{eq:fol-init}) and
(\ref{eq:fol-init-atoms})
is represented as the following state of knowledge
\begin{equation}
s = \left\{
\begin{array}{l}
\psi=\apsi{\neg Box(c) \vee \neg In(g,c) \vee \neg Fragile(g)}{\set{c
=\textit{FragileStuff}}},\\ 
Box(\textit{FragileStuff}), In(Wine,\textit{FragileStuff}), Fragile(Wine)\\
  \end{array}
\right\}
\end{equation}

\ignore{
Recall
that LCW inference is incomplete.
If, in addition to (\ref{eq:lcw}), the robot also knew
that there is a book in \m{Box3}, i.e.\begin{equation}\label{eq:atoms}
				      {In (Book, Box3), Box(Box3)},
				      \end{equation}  it should be able to
conclude that 
the book is not fragile.  The LCW reasoner would return that the value of
\m{Fragile(Book)} is unknown. In contrast, \psiplan{} concludes from (\ref{eq:intro-psi}) and (\ref{eq:atoms}) that \m{\neg  Fragile(Book)}\footnote{ Note that in \psiplan{} and LCW representations it is assumed that different constants always denote different objects.}. 
}

Now suppose that a new container \m{Box10} is brought into the room whose
contents are completely unknown. In the resulting situation, the location of
all fragile goods is known {\em except} for those that might be in \m{Box10}.
The \psiplan{} state update would yield the following new state of knowledge \m{s'} by adding the atom \m{Box(Box10)} to \m{s} and adding an exception  to
\m{\psi} yielding \md{\psi'=\apsi{\neg  Box(c) \vee \neg In(g,c)
\vee \neg Fragile(g)}{\set{c = \textit{FragileStuff}},\set{c=Box10}}.} 
The updated state of knowledge \m{s'} represents the new situation precisely:
\begin{equation}\eqalign{
&s'= \cr&\left\{ \begin{array}{l}
\psi'=\apsi{\neg{} Box(c)\vee \neg In(g,c) \vee \neg Fragile(g)}{\set{c =
\textit{FragileStuff}}, \set{c=Box10}},\\
Box(\textit{FragileStuff}), Box(Box10), In(Wine,\textit{FragileStuff}), Fragile(Wine)\\
  \end{array}\right\}}
\end{equation}

\psiplan{}'s action language includes actions with \psiform{} preconditions. For example, the action {\em lift} of lifting object \m{B}, requires that there be no fragile items in it, i.e. 
\begin{equation}\label{eq:psi-precond}
\spsi{\neg In(g,B) \vee \neg Fragile(g)}
\end{equation}
\begin{sloppypar}
When an agent whose state of knowledge is described with \m{s'\cup{}Box(Box5)}
 is given a goal of lifting box \m{Box5} from its location, it will
establish that the precondition  \m{\spsi{\neg In(g,Box5) \vee \neg
Fragile(g)}} of the {\em lift} action is entailed by \m{\psi'} and
proposition \m{Box(Box5)}. Indeed, if no boxes contain fragile goods,
except for the box {\em FragileStuff} and possibly \m{\textit{Box10}},
then there are no fragile goods in \m{\textit{Box5}}.  The \psiplan{}
reasoning algorithms that are involved in this inference do not expand
the universal quantification in the universal base and do not require
the knowledge of all domain objects by the agent.   
\end{sloppypar}
\end{ourexample}

Another illustration of the  advantage of \psiplan's ability to 
reason with quantified sentences is an example from the blocks world domain.
\psiplan{} eliminates the need for predicate  \m{Clear(B)} as a way 
of stating that nothing is on block \m{B}. 
Instead, \psiplan's representation uses \m{\spsi{\neg{}On(b,B)}}.
The advantage of using the latter
representation is that the fact that block \m{A} is on block \m{B} by
itself implies that block \m{B} is not clear, eliminating the need to
state \m{\neg{}Clear(B)} as another effect of moving block \m{A} onto
\m{B}.

This use of quantified preconditions distinguishes \psiplan{} from
other representations  (\cite{brafman-hoffman-04},
\cite{baral-son-97}, \cite{eiter-etal-00},\cite{smith-weld-98},
\cite{weld-etal-98}, \cite{cimatti-roveri-99},
\cite{ambros-ingerson-steel-88}, \cite{krebsbach-etal-92},
\cite{peot-smith-92}), that only admit actions with preconditions
limited to atoms or literals.  On the other hand, many of these
conformant planners handle actions with conditional effects, which are
absent from \psiplan. While
\psiplan{} can be easily extended to represent actions with 
conditional effects, 
complete planning with conditional effects is in the complexity class
\m{\Sigma_2P} (e.g. \cite{baral-etal-00,{turner-02}}), while complete
planning with \psiplan{} appears to be an NP-complete problem. This
issue is further discussed in Section \ref{sec:rel-complexity}. 

\ignore{
Along with introducing \psiforms{}, the example demonstrated that
\psiplan{} correctly represents state resulting from taking the action
that introduces a new object (\m{\textit{Box10}}) to the domain. All of the algorithms involved in the state update after an action are presented in the paper. }
\psiplan{} admits procedures for entailment and state update after an
action (including actions that introduce a new object) that are sound, complete and take polynomial time. In 
particular, the complexity of entailment grows linearly with respect
to the number of \psiforms{} in the knowledge base when the number of
literals, variables and exceptions in each \psiform{} are bounded. The
complexity bound on  \psiform{} reasoning is polynomial in the number
of exceptions.

The rest of the paper is organized as follows: Section \ref{sec:rep}  formally defines \psiforms{} and presents a few simple properties. In Section \ref{sec:calculus}, we present the \psiform{} calculus and complexity results. 
Section \ref{sec:psiplan}  introduces 
the \psiplan{} representation of a state of knowledge, actions and state update after an action.  \ignore{ Section \ref{sec:eval} presents the
results of empirical evaluation of \psiplan{} and \psipop{}. }
 Section \ref{sec:related} contains an overview of the related work.  Finally,
Section \ref{sec:conclusions} summarizes and draws conclusions.

\section{The language of \psiforms}\label{sec:rep}

\subsection{Definitions and notation}\label{sec:defs}
\ignore{
In \psiplan{}, a {SOK}\label{def:SOK} is a finite set of \psiplan{} \mydefi{
domain propositions} or, simply, \mydefi{propositions}, which are defined to
be either ground atoms or \psiforms{}. 
}

We assume no function symbols except for constants in the language. 
The number of constants is infinite.

The general form of a \psiform{} is:
\begin{equation}\label{def:psiform}
{\psi = {\apsi{\neg{}Q_1(\vec{x_1}) \lor \ldots \lor \neg{}Q_k(\vec{x_k})} {\sigma_1, \ldots ,\sigma_n}}} 
\end{equation}

\noindent \m{k\geq 1} and \m{n\geq 0},  and each \m{Q_i(\vec{x_i})} is any atom whose only variables are  \m{\vec{x_i}}. The clause \m{\neg{}Q_1(\vec{x_1}) \lor \ldots \lor \neg{}Q_k(\vec{x_k})} is called the \mydefi{main clause} of \m{\psi} and is denoted by  \m{\main(\psi)}. The set of all variables of the main clause, i.e.  the set \m{\vec{x}  =\bigcup_{i=1}^{k}\vec{x_i}}, is denoted by \m{\vars(\psi)}.  
\ignore{In addition, we require, that for
every atom \m{Q_i(\vec{x_i})}, every variable in \m{\vec{x_i}} occurs no more
than once in its argument list \footnote{This requirement is essential for  keeping the \psiform{} calculus closed, i.e. ensuring the result of
the calculus operations is always a \psiform{}.}.
 Thus, \m{\apsi{\neg P(x,y)\vee
    \neg Q(y)}{\set{x = a}}} is a \psiform{}, while \m{\apsi{\neg P(x,x)\vee
    \neg Q(y)}{\set{x = a}} } is not because of multiple occurrences of \m{x}
in \m{P(x,x)}.}

\ignore{Each \m{\sigma_i} is just a ground substitution on a non-empty subset of variables in \m{\vec{x}}, where by ground we mean a substitution that
binds variables to constants. }
{Each \m{\sigma_i} is a substitution on a non-empty subset of variables in \m{\vars(\psi)}, that binds a variable to another variable from \m{\vars(\psi)}, or to a constant.}
The set of all substitutions appearing
in a \psiform{} is denoted \m{\Sigma(\psi)}. Each \m{\sigma_i} 
represents {\em exceptions} of \m{\psi}.
Thus, a \psiform{} can be abbreviated as
\m{[\main(\psi)\mbox{ except } \Sigma(\psi)]}.  

When \m{\Sigma(\psi)} is empty, we call such a \psiform{} \mydefi{simple}
and write \m{\spsi{\main(\psi)}} instead of \m{\apsi{\main(\psi)}{}}.
A simple \psiform{} with no variables is called a \mydefi{singleton} and
represents a single ground clause.

Given a \psiform{} \m{\psi =[\main(\psi)\mbox{ except } \Sigma(\psi)]} we will need to refer to the following:  
\begin{itemize}
\item  a simple \psiform{} \m{\spsi{\main(\psi)}} that is obtained
from the main clause of \m{\psi}, called \mydefi{main part}. 
\item simple \psiforms{} that are obtained by instantiating the main clause
with substitutions from \m{\Sigma(\psi)}, called \mydefi{exception forms}. For each \m{\sigma_i} from
\m{\Sigma(\psi)}, we call \m{\main(\psi)\sigma_i} the i-th \mydefi{exception
clause} denoted \m{\exci{i}(\psi)}.
\m{\exc(\psi)} is the set of all of \m{\psi}'s
  exception forms:
 \m{\set{ \spsi{\exci{i}(\psi)} \ourst 1\leq i \leq \norm{\Sigma(\psi)}}}.
\end{itemize}

%
\ignore{
So, for \m{\psi_0=\apsi{\neg P(x,y)\vee\neg  Q(y,A)}{\set{x=A},
\set{x=C, y=D}}} we have
\begin{itemize}
\item the main clause \m{\main(\psi_0)=\neg P(x,y)\vee\neg{}Q(y,A)},
\item the main part  \m{\spsi{\main(\psi_0)}=\spsi{\neg P(x,y)\vee\neg{}Q(y,A)}},
\item \m{\vars(\psi_0)=\{x,y\}}, 
\item  \m{\Sigma(\psi_0)=\set{\set{x=A},\set{x=C, y=D}}},
\item \m{\sigma_1=\set{x=A},\sigma_2 = \set{x=C, y=D}},
\item  \m{\exci{1}(\psi_0)=\neg P(A,y)\vee\neg Q(y,A)},
 \m{\exci{2}(\psi_0)=\neg P(C,D)\vee\neg Q(D,A)}, and 
\item \m{\exc(\psi_0)=\set{\spsi{\neg P(A,y)\vee\neg Q(y,A) },
 \spsi{\neg P(C,D)\vee\neg Q(D,A)}}}.
\end{itemize}
}

\begin{figure}
\begin{tabular}{lll}
\hline\\ 
Name & Notation & Example\quad$\psi=[\neg P(x,y,z)\vee\neg Q(y,A)$\\
&&\phantom{Example\quad$\psi=[$}$\mbox{ except }\set{x=A},\set{x=C,y=D}]$
\vspace{2.mm} \\
\cline{1-1}  \cline{2-2}  \cline{3-3}
\hline  \\
main clause & \m{\main(\psi)} & \m{\neg P(x,y,z)\vee\neg{}Q(y,A)}\\ 
main part & \m{\spsi{\main(\psi)}} & \m{\spsi{\neg P(x,y,z)\vee\neg{}Q(y,A)}}\\ 
variables & \m{\vars(\psi)}	& \m{\{x,y,z\}} \\ 
exceptions & \m{\Sigma(\psi)} &\m{\set{\set{x=A},\set{x=C, y=D}}} \\ 
exception clauses & \m{\exci{1}(\psi)} & \m{\neg P(A,y,z)\vee\neg Q(y,A)}\\ 
& \m{\exci{2}(\psi)} & \m{\neg P(C,D,z)\vee\neg Q(D,A)} \\
exception forms & \m{\exc(\psi)} & \m{\set{\spsi{\neg P(A,y,z)\vee\neg Q(y,A) }, \spsi{\neg P(C,D,z)\vee\neg Q(D,A)}}} \vspace{2.mm} \\
\hline
\end{tabular}
\caption{\nb{Summary of \psiform{} notation.}}
\end{figure}

\ignore{
\tamara{Jim - you added the following very useful paragraph, but at this point it wasn't defined that \psiforms{} denote clauses - it follows in the next subsection.}
Note that an exception may not bind all the variables in the main clause, as shown in \m{\sigma_1} of \m{\psi_0} above. In such cases, the exception is equivalent to a possibly infinite set of clauses, as shown in \m{\exci{1}(\psi_0)}, where \m{y} is universally quantified.}

We use \m{C} and \m{V}  to denote respectively the maximum number of
literals and number of variables in the main clause for a given set of \psiforms{}. We use  \m{E} to denote the maximum size of the set \m{\Sigma(\psi)}.
The cardinality of each predicate is assumed to be constant bounded, thus the time for unification of two literals is also constant bounded.

\ignore{
\m{C}, \m{V} and \m{E}  are assumed to be fixed for a given application, unless stated otherwise.  These assumptions are typical for planning problems that are
considered by the existing open world planners
(see \cite{golden-etal-94},\cite{golden-thesis} and experiments described in this paper). 
}

Unless noted otherwise, everywhere in this paper symbols \m{x,y,z, x_1, y_1,
  z_1, \ldots} denote variables, capital letters \m{A, B, \ldots} denote
 constants , and \m{\psi, \psi_1, \ldots} -- denote \psiforms{}.  Also,
 we assume that the variables in two distinct
\psiforms{} are always renamed to be distinct.

\subsection{\psiforms{} as Sets}\label{sec:psiforms-as-sets}

A \psiform{} is a representation of a possibly infinite set of ground
clauses. Here and throughout the paper, clauses that consist of the
same set of 
literals are considered {\bf equal}.  The logical equivalence of such clauses
allows us to disregard the order of their literals.

\ignore{
A set of ground clauses represented by a \psiform{} is called a
\mydefi{\m{\psi}-set}. The \m{\psi}-set corresponding to the \psiform{}
\m{\psi} is defined with operation \m{\setpsi}. 
}

We define the set of ground clauses represented by a \psiform{}
\m{\psi}, called a \mydefi{\m{\psi}-set} and denoted \m{\setpsi(\psi)},
as follows:

\begin{enumerate}

\item When \m{\psi} is {\em simple}, the set defined by  \m{\psi} 
consists of all ground instantiations of the main clause
\begin{equation}\label{def:psiformset-a}
\setpsi(\psi) = \set{\main(\psi)\sigma\ourst \main(\psi)\sigma \mbox{
is ground}} 
\end{equation}
\noindent This definition implies \m{\setpsi(\spsiset{c}) = \set{c}},
when \m{c} is a ground clause.  

\item When  \m{\psi} is {\em not simple}, the set defined by  \m{\psi}
 consists of all ground instantiations of the main clause
minus the set of  all ground instantiations of exception clauses.
\begin{equation}\label{def:psiformset-b}
\setpsi(\psi) = \setpsi(\spsiset{\main(\psi)}) -
\setpsi(\spsiset{\exci{1}(\psi)}) - \ldots -
\setpsi(\spsiset{\exci{n}(\psi)}),
\end{equation}
where \m{n = \norm{\Sigma(\psi)}}.
Any clause in the set 
\m{\setpsi(\spsiset{\exci{1}(\psi)}) \cup \ldots \cup
\setpsi(\spsiset{\exci{n}(\psi)})} is called an \mydefi{exception} of
\m{\psi}.

Figure \ref{fig:psifig} illustrates the  definition of a \m{\psi}-set.
\begin{figure}
  \centering
    \parbox[b]{\textwidth}{%
      \centering \psset{unit=.5}


\begin{pspicture}(0,0)(10,5)
  \newpsstyle{spsi}{fillcolor=lightgray,fillstyle=solid,
                    linestyle=solid,linewidth=.5pt}
  \newpsstyle{psiexc}{fillcolor=white,fillstyle=solid,
                    linestyle=solid,linewidth=.5pt}

  \psellipse[style=spsi](5,2.5)(3,2) 
     \rput(4,1.3){\rnode{main}{}}   \rput(0,0){\rnode{lmain}{\m{\setpsi(\spsi{\main(\psi)}})}}
	  \ncline{lmain}{main}

  \psellipse[style=psiexc](3.8,3)(.5,.6)
     \rput(3.8,3){\rnode{exc1}{}}   \rput(0,3.3){\rnode{lexc1}{\m{\setpsi(\spsi{\exc_1(\psi)}})}}
	  \ncline{lexc1}{exc1}

  \psellipse[style=psiexc](6,1.5)(.3,.3)
     \rput(6,1.5){\rnode{exc2}{}}   \rput(10,1.2){\rnode{lexc2}{\m{\setpsi(\spsi{\exc_2(\psi)}})}}
	  \ncline{lexc2}{exc2}

  \psellipse[style=psiexc](6, 3)(.6,.4)
   \rput(6,3){\rnode{exc3}{}}   \rput(10,3.3){\rnode{lexc3}{\m{\setpsi(\spsi{\exc_3(\psi)}})}}
	  \ncline{lexc3}{exc3}

\end{pspicture}}
\caption{The set of clauses defined by a \m{\psi}.  \m{\setpsi(\psi)}
  is depicted as the gray area and consists of all clauses of the main part
  \m{\spsi{\main(\psi)}} that are not exceptions, i.e. are not in any of
  \m{\spsiset{\exci{1}(\psi)},\ldots,\spsiset{\exci{3}(\psi)}} . The main
  part \m{\spsi{\main(\psi)}} contains the superset of all clauses of
  \m{\psi}.}
\label{fig:psifig}
\end{figure} 
\end{enumerate}

\edit{A2-2}{To combine and compare $\psi$-sets represented by  different \psiforms{},
we introduce set operations and define the resulting  $\psi$-set as follows.} 
\begin{enumerate}
\item For a set of \psiforms{} \m{\set{\psi_1,\ldots,\psi_k}} their
    \m{\psi}-set is the union of the \m{\psi}-sets of its elements.
\begin{equation}\label{def:setpsiformset}
\setpsi(\set{\psi_1,\ldots,\psi_k})  = \cup_{i=1}^k\setpsi(\psi_i),
\end{equation}

\item\edit{A2-2}{An expression \m{\Box_1 * \Box_2}, where  \m{\Box_1} and \m{\Box_2} denote either a single \psiform{} or a set of \psiforms{}, and \m{*}  denotes any of the set operations \m{\cap, \cup, -, {\image{}{}}} or \m{\diff} (last two operations are defined in the next section) represents a set of ground clauses obtained by applying the * operation to the corresponding $\psi$-sets.}

\begin{equation}\label{def:setoperations}
\setpsi(\Box_1 * \Box_2) = \setpsi(\Box_1) * \setpsi(\Box_2).
\end{equation}

\item \edit{}{Let \m{A} and \m{B} be \psiforms{} or \psiform{} expressions. 
We write \m{A=B} and call \m{A} and \m{B} \mydefi{equivalent} if  and only if  \m{\setpsi(A)=\setpsi(B)}, in other words, the sets of ground clauses represented by each \psiform{} or  expression are the same.} 
\end{enumerate}

For example, the statement  \m{\psi_1 = \psi_2\cap\psi_3}  represents the equivalence of two $\psi$-sets: 
\m{\setpsi(\psi_1)} and  \m{\setpsi(\psi_2 \cap \psi_3)}. The latter, in turn,  according to definition (\ref{def:setoperations}) denotes the intersection \m{\setpsi(\psi_2) \cap \setpsi(\psi_3)}
\ignore{
\begin{ourexample}
Consider the equality \m{\psi_1 = \psi_2\cap\psi_3} where 
\md{
\begin{array}{l}
\psi_1 = \apsi{\neg{}P(w,B)}{\set{w=A}}, \\
\psi_2 = \apsi{\neg{}P(x,y)}{\set{x=A}}, \\
\psi_3 = \spsi{\neg{}P(z,B)}
\end{array}
}
\m{\psi_1 = \psi_2\cap\psi_3}  represents the equivalence of two $\psi$-sets: 
\m{\setpsi(\psi_1)} and 
\m{\setpsi(\psi_2 \cap \psi_3)}. The latter, according to definition (\ref{def:setoperations}) denotes the intersection \m{\setpsi(\psi_2) \cap \setpsi(\psi_3)}. 
 \end{ourexample}
}

\subsubsection{\m{\psi}-set Membership}

We say that a {\em ground clause \m{c} is in \m{\Box}}, written \m{c
\in \Box} instead of  \m{c \in \setpsi(\Box)}.

Thus, according to  (\ref{def:psiformset-b}),
given a \psiform{} \m{\psi},  a ground clause \m{c} is in
\m{\psi} if and only if it can be obtained by instantiating the main
clause, \m{\main(\psi)}, with some ground substitution \m{\sigma},
and cannot be obtained by instantiating any of \m{\psi}'s exception
clauses \m{\exc_1(\psi), \ldots, \exc_{n}(\psi)}.

\begin{equation}\label{def:cinpsi}
c \in \psi \mbox { iff } \ourexists{\sigma} c =
    \main(\psi)\sigma \mbox{ and } \ourforall{\theta,i} 1\leq i \leq
    n \implies c \neq \exci{i}(\psi)\theta
\end{equation}

We also define membership of a clause in a set of \psiforms{} \m{\Psi}
in the obvious way:
\begin{equation}\label{def:cinPsi}
c \in \Psi \mbox { iff } \ourexists{\psi\in\Psi}  c\in\psi
\end{equation}

Deciding the membership of a ground clause in a simple \psiform{} amounts to
finding a 
substitution \m{\sigma} that {\em matches} the literals in the main
clause of the \psiform{} with the literals of the  clause. In a
\psiform{} with exceptions, after establishing membership in the main
part, \m{\spsi{\main(\psi)}}, it is necessary to verify non-membership
in the \psiform{}'s exception forms, all of which in turn are simple
\psiforms. 

We define an operation {\em set-match} that computes such matching
substitutions used to generate a clause from the main clause of a \psiform{}.
Given two clauses \m{a} and \m{b}, where variables in \m{a} and \m{b} are
distinct and denoted \m{V_a} and \m{V_b} respectively, we say that \m{a}
\mydefi{set-matches}\label{def:set-match} with \m{b} if and only if there
exists a substitution \m{\sigma} on variables in \m{V_a} such that
\m{a\sigma=b}. The set of all most general such matching \m{\sigma}'s is denoted
\m{\mgu(a,b,V_a)}.\footnote{Two substitutions \m{\sigma_1} and \m{\sigma_2} are equal if and only if the two can be made identical by consistently renaming variables in both. Thus \m{\mgu(a,b,V_a)} does not include two equivalent substitutions.}  \ignore{We use \m{\mgu(a,b,V_a)} as an instance of the
  more general operation \m{\mgu(a,b,V)} that restricts any substitution to
  binding only variables from \m{V}.}

Example \ref{ex:cinpsi} below demonstrates  that there can be  more than one way a \psiform{} clause can set-match with a ground clause. 

\begin{ourexample}\label{ex:cinpsi}
Let \m{\psi=\apsi{\neg{}P(x) \vee\neg{}P(y)}{\set{ x = A}}} and
let \m{c = \neg P(A)\vee \neg P(B)}. There are two substitutions
\m{\sigma_1 = \set{x=A, y=B}} and \m{\sigma_2 = \set{x=B, y=A}} such
that \m{\main(\psi)\sigma_1 = \main(\psi)\sigma_2 = c}.  
However, \m{c} can also be generated by instantiating the exception clause
\m{\exc_1(\psi) =\neg{}P(A) \vee\neg{}P(y)}  with substitution  
 \m{\set{y=B}}, therefore \m{c\not\in\psi}.
\end{ourexample}

\corr{2}{The possibility of multiple different instantiations producing the same ground clause complicates the reasoning with \psiforms{} of this type, as even deciding membership of a ground clause in a \psiform{} becomes somewhat more problematic compared to the case of \psiforms{} for which each ground clause is generated with a {\em unique} substitution. Essentially, when each ground clause is obtained with a unique instantiation of the main form, checking \m{c\in\psi} amounts to computing a set-match \m{\mgu(\main(\psi), c, \vars(\psi))} and, in case the set-match results in a substitution \m{\sigma}, checking that \m{\sigma} is not a superset of any substitution in \m{\psi}'s exceptions \m{\Sigma(\psi)}. When, as in the Example \ref{ex:cinpsi},  \m{\mgu(\main(\psi), c, \vars(\psi))} consists of more than a single substitution, we must consider {\em all} such substitutions in relation to the exceptions. 
} To avoid the increase in the complexity of reasoning we introduce a notion of a {\em fixed length} \psiform{}. 

A \psiform{} is called \mydefi{fixed length}\label{def:fixedLength} if
and only if no two  literals of the main clause unify when  the
variables in both literals are renamed to be distinct. Thus,
\m{\spsi{\neg{}P(x,A)\vee \neg{}P(B,x)}} is not a fixed length
\psiform{}, because \m{\neg{}P(x_1,A)} and  \m{\neg{}P(B, x_2)} are
unifiable. \corr{1.2}{On the other hand,  \m{\spsi{\neg{}P(x,A)\vee \neg{}P(x, B)}} and  \m{\spsi{\neg{}P(x,A)\vee \neg{}Q(y)}}  are examples  of fixed length \psiforms.}
\ignore{The notion of fixed length clause is stronger
than the notion of condensed clauses \cite{leitsch-97}, because the
variables in literals are renamed to be distinct. So,  
\m{\neg{}P(x,a)\vee \neg{}P(b,x)}  is a condensed, but not fixed
length. However, it is possible that the essential for us properties of fixed
length \psiforms{} are satisfied by requiring main clause be a condensed
clause}

\ignore{
\nb{move this definition where it is properly motivated?}
A \psiform{} is called \mydefi{limited form}\label{def:limitedForm} if
and only if all predicate symbols (\m{Q_i}-s) in its main clause are
distinct. Clearly, a limited form \psiform{} is also a fixed length
\psiform{}. 
}

\begin{ourobservation}\label{lm:unique}
\lemmaOne
\end{ourobservation}

\proof
  The proof is by contradiction. Assuming there is a ground clause that
  is generated by more than one substitution on \m{\main(\psi)}, it is
  possible to construct a unifier for two literals of the main clause.\qed

Other important properties of  fixed length \psiforms{} ensuring reduced complexity of reasoning are discussed in the end of Section \ref{subsec:complexity}.

Thus everywhere except for general \psiform{} entailment
theorems in Section \ref{sec:entailment} we limit our attention
to fixed length \psiforms{}.
 
\ignore{\begin{proof}
Suppose \m{c} is a clause in \m{\psi}. If there are more than one
substitution on \m{\main(\psi)} that generates \m{c}, then it is the
case that two different literals of   \m{\main(\psi)} match onto the
same literal of \m{c}. Suppose  \m{\sigma_1} and
  \m{\sigma_2} be substitutions corresponding to each set-match, and
that  they match two different
  literals \m{d_1} and \m{d_2} of \m{\main(\psi)} on the same
  literal \m{d} of \m{c}, i.e.  \m{d_1\sigma_1 = d_2\sigma_2 = d}.

Consider the following substitution \m{\sigma}.
\begin{itemize}
\item If \m{d_1} and \m{d_2} do not share variables, construct 
\m{\sigma} as follows: combine bindings on variables in \m{d_1} from
\m{\sigma_1} and  bindings on variables in \m{d_2} from \m{\sigma_2}.
Then, obviously, \m{d_1\sigma = d_2\sigma = d}.
  
\item Otherwise, if \m{d_1} and \m{d_2} do share variables, rename
variables in \m{d_2} so that there is no overlap with variables in
 \m{d_1}.  Modify \m{\sigma_2} by renaming the variables in the same
 way we did with  \m{d_2}, and construct \m{\sigma} as in the previous
 case. Again, \m{d_1\sigma = d_2\sigma}.
\end{itemize}

Thus, both cases produced a contradiction to the fact that no two
literals of a fixed length \m{\psi_1} unify.

\end{proof}
}
\ignore{
The proof immediately follows from the fact that given two fixed
length \psiforms{} such that the main clause of one subset-matches  onto
another,  no two literals of the first match the same literal of the
second. This property is proven next by Lemma \ref{lm:fixednooverlap}. 
}

A \psiform{} is called \mydefi{well-formed} if and only if it has no redundant
exception forms, i.e. there is no subsumption between any two exception
clauses.  Any \psiform{} can be reduced to a well-formed equivalent;
henceforth, we only consider well-formed \psiforms.  The reduction procedure
is simple and consists of examining pairs of different substitutions
\m{\sigma_i, \sigma_j} of \m{\Sigma(\psi)}. If \m{\sigma_i\subseteq
  \sigma_j}, then (and only then)
\m{\spsi{\exci{j}(\psi)}\subseteq\spsi{\exci{i}(\psi)}}, and so we remove
\m{\sigma_j} from \m{\Sigma(\psi)} and vice versa.  In the worst case we will
need to examine \m{E(E-1)/2} pairs.

\subsection{\psiform{} Logic}  \label{ssec:logic}

\ignore{\edit{B(5)}{
Interpretation of a set of \psiplan{} propositions is defined
similarly to the classical first order notion of interpretation but is
restricted to what we call {\em sufficiently large domains}. A domain
\m{D} is \mydefi{sufficiently large} with respect to a set \m{S} of
\psiplan{} propositions (atoms and \psiforms) if its size is greater
than the number of distinct constants 
in \m{S}. Certainly, an infinite domain is sufficiently large with
respect to any finite set of \psiforms{}.  
}
}

An \mydefi{interpretation} or \mydefi{world} is a triple (D, M, A), where D is a domain, M is a mapping between the constants of the language and the domain objects, and \m{A} is a truth assignment on all ground atoms of the language. We limit worlds to those with infinite domains. \corr{} {We further assume each constant denotes a distinct domain object.} A \mydefi{model} of a proposition is a world that assigns \m{\true} to that formula. 

When \m{s} is a proposition or a set of propositions and \m{w} is a world, we write \m{w(s)} if and only if \m{w(s)} is {\em true} in \m{w}. For a set of propositions to be true in \m{w}, each element must be true in \m{w}. 
We write \m{\Ms(s)} to denote the set of all models of \m{s}, i.e. \m{\Ms(s) = \set{w|w(s)}}.

\edit{B(5)}{
We use the standard rules regarding the interpretation of atoms, negation
and logical connectives. A \psiform{} or a set of \psiforms{}, denoted
below by \m{\Box}, is interpreted as (a possibly infinite) conjunction of all
ground clauses it represents, and therefore 
\begin{equation}
\label{def:psiforminterpretation}
{\Ms}(\Box) = \bigcap_{c\in\setpsi(\Box)}{\Ms}(c).
\end{equation}
}

\edit{B(5)}{
Now that we have defined an interpretation for \psiforms{} we can
define {\em entailment} in a logical language containing \psiforms{}
in the usual way.  A formula \m{a} \mydefi{entails} a formula \m{b}, denoted \m{a\entails{}b}, if and only if every model of \m{a} is also a model of \m{b}, i.e. \m{\Ms(a) \subseteq \Ms(b)}.} 

We first examine the entailment between two ground clauses of negated
literals.  We write \m{c_1\subseteq c_2} when a set of literals of the clause
\m{c_1} is a subset of the set of literals of the clause \m{c_2}.
It is easy to see, that given two non-empty grownd clauses of negated literals 
\m{c_1} and \m{c_2}, \m{c_1\entails c_2} if and only if \m{c_1\subseteq c_2}.
Further, observe, that when \m{\Box_1} and \m{\Box_2} are two \psiforms{} or sets of
\psiforms{} \m{\Box_1 = \Box_2} if and only if \m{\Box_1\entails\Box_2} and
\m{\Box_2\entails\Box_1}.

\ignore{
\begin{ourobservation}\label{lm:LitEntailment}
\lemmaTwo
\end{ourobservation}
}

\ignore{
\begin{ourlemma}\label{lm:equiv} When  \m{\psi_1} and \m{\psi_2}
are  fixed length \psiforms{},   \m{\psi_1 = \psi_2} if and
only if \m{\psi_1\entails\psi_2} and   \m{\psi_2\entails\psi_1}.
When  \m{\Box_1} and \m{\Box_2}
consist of only  fixed length \psiforms{},   \m{\Box_1 = \Box_2} implies
that \m{\Box_1\entails\Box_2} and \m{\Box_2\entails\Box_1}.
\end{ourlemma}

\begin{proof}
  Trivially follows from the definitions of entailment and \psiform{}
  equivalence.
\end{proof}
}

Note as well,  that \m{\psi =\apsi{Q(\vec{x})}{\sigma_1, \ldots,\sigma_n}} can
be equivalently written as a first order formula that universally
quantifies the variables of \m{\psi}: 
\md{
\ourforall{{\vec x}}Q({\vec x})\vee c_1 \vee \ldots \vee c_n,}
where each \m{c_i} is an equality constraint obtained from
\m{\sigma_i}, for instance if \m{\sigma_i = \set{x=A,y=B}} then
\m{c_i = (x=A \land y=B)}.
We therefore call non-singleton \psiforms{} \mydefi{quantified}.

\section{Calculus of \psiforms{}}\label{sec:calculus}

  In this section we present the calculus of \psiforms{}.  We first
  demonstrate how subset, intersection and set-difference between
  \psiforms{} are computed in simple cases. {\corr{1.1}{These
  operations lay the foundation for algorithms that compute
  entailment, as well as the computation of {\em image} and {\em
  e-difference} operations, which we define here. These operations are
  essential parts of reasoning and planning with \psiforms{}. For
  example, they are used in \psipop{} (\cite{babaian-schmolze-00}) and
  \psigraph{} (\cite{carlin-etal}) planners to determine if an effect
  of an action can bring about or undo a goal. E-difference is also
  used in the \psiplan's{} state update computation (\ref{update}),
  presented later in this paper.}}  Sound and complete methods of
  computing entailment, image and e-difference of fixed length
  \psiforms{} are presented in the form of theorems that are
  \corr{1.3}{easily convertible to algorithms}.  We summarize the
  complexity of computing entailment, image and e-difference between
  \psiforms{} in \psiplan{}.

\subsection{Operations \m{\subseteq, \cap} and \m{-}}

  The operations subset, intersection and set-difference between
  \psiforms{} are defined in the obvious way: \edit{A(3)}{for any
  ground clause \m{c} \md{c\in \Box_1*\Box_2\textrm{~if and only
  if~}c\in{}\phi(\Box_1)*\phi(\Box_2)} } where \m{\Box} represents
  either a single \psiform{} or a set of \psiforms, and \m{*}
  represents any of the operations \m{\subseteq, \cap} or \m{-}.
  Calculation of \m{\subseteq, \cap} and \m{-} is straightforward for
  simple fixed length \psiforms{}.

\edit{A(4)}{
\m{\spsi{\main(\psi_1)}\subseteq\spsi{\main(\psi_2)}} if and only if all of
\m{\spsi{\main(\psi_1)}}'s clauses are also clauses of
\m{\spsi{\main(\psi_2)}}. This requires that the main clause
\m{\main(\psi_2)} set-matches onto \m{\main(\psi_1)} with some substitution
\m{\sigma}.  When this is true, for every ground substitution \m{\sigma_1} on
the variables of \m{\psi_1}, there is a ground substitution \m{\sigma_2 =
  \sigma\sigma_1} on \m{\main(\psi_2)} such that
\m{\main(\psi_1)\sigma_1=\main(\psi_2)\sigma_2}.
}Thus, for any \m{\psi_1}
and \m{\psi_2}, checking whether or not
\m{\spsi{\main(\psi_1)}\subseteq\spsi{\main(\psi_2)}} amounts to computing
\m{\mgu(\main(\psi_2),\main(\psi_1), \vars(\psi_2))}.

\m{\spsi{\main(\psi_1)}\cap\spsi{\main(\psi_2)}} is defined by all ground
substitutions \m{\sigma_g} such that \m{\main(\psi_1)\sigma_g =
  \main(\psi_2)\sigma_g}.  The set of most general \m{\sigma}'s for which
\m{\main(\psi_1)\sigma = \main(\psi_2)\sigma} defines a set of main clauses
that generate the \psiforms{} denoting the intersection
\m{\spsi{\main(\psi_1)}\cap\spsi{\main(\psi_2)}}.

We say clause \m{a} \mydefi{set-unifies}\label{def:set-unify} with 
clause \m{b} if and only if there exists a substitution \m{\sigma} 
such that \m{a\sigma = b\sigma}, denoting the set of  all most
general  such \m{\sigma}'s by \m{\mgu(a,b)}. Thus, \m{\spsi{\main(\psi_1)}\cap\spsi{\main(\psi_2)} = \set{\spsi{\main(\psi_1)\sigma} \ourst \sigma\in
\mgu(\main(\psi_1),\main(\psi_2))}}.

\ignore{
\begin{ourexample}\label{ex:inter}
Let \md{\begin{array}{l}
\psi_1 = \spsi{\neg P(x,y) \vee \neg Q(x,A)} \\  
\psi_2 = \spsi{\neg P(w,B) \vee \neg Q(w,z)}
        \end{array}
}
First, observe that the intersection \m{\psi_1\cap\psi_2} consists of
only those clauses that are obtained by instantiating the main clause
\m{\neg P(w,B) \vee \neg Q(w,A)}.
\m{\mgu(\main(\psi_1),\main(\psi_2))} consists of a single
\footnote{up to variable renaming}
substitution \m{\sigma = \set{x=w,y=B,z=A}}, which, when applied to 
either  \m{\main(\psi_1)} or \m{\main(\psi_2)} produces exactly 
\m{\neg P(w,B) \vee \neg Q(w,A)}, i.e.,  \m{\psi_1\cap\psi_2 =
\set{\spsi{\main(\psi_1)\sigma}}= \set{\spsi{\main(\psi_2)\sigma}}}.
\end{ourexample}
}
\ignore{
As Example \ref{ex:inter}  demonstrates, 
}
The  intersection of the main parts of two \psiforms{} \m{\psi_1} and
\m{\psi_2} consists of the clauses denoted equivalently by each of the
two sets of \psiforms{}
\md{\begin{array}{l}
\set{\spsi{\main(\psi_1)\sigma} \ourst \sigma\in
\mgu(\main(\psi_1),\main(\psi_2))}, \mbox{ and } \\
\set{\spsi{\main(\psi_2)\sigma} \ourst \sigma\in
\mgu(\main(\psi_1),\main(\psi_2))}.
\end{array}
} 
For example, when \m{\psi_1 = \spsi{\neg P(x,A)}} and  \m{\psi_2 = \spsi{\neg P(B,y)}}, \m{\mgu(\main(\psi_1),\main(\psi_2)) = \set{\set{x=B,y=A}}} and \m{\psi_1\cap\psi_2} equals  \m{\set{\neg P(B,A)}}.

When \m{\psi_1} is simple and is a subset of \m{\spsi{\main(\psi_2)}} 
and both \psiforms{} are fixed length, {\em
subtracting} \m{\psi_1} from \m{\psi_2} \label{op:minus} 
is a matter of adding a substitution \m{\sigma = \mgu(\main(\psi_2), \main(\psi_1), \vars(\psi_2))},  which generates \m{\psi_1} from \m{\main(\psi_2)}, to \m{\Sigma(\psi_2)}.
Indeed, \m{\spsi{\main(\psi_2)\sigma}} equals \m{\psi_1}, thus, by
adding \m{\sigma} to  \m{\Sigma(\psi_2)} we are
 subtracting from \m{\psi_2}  the clauses of
 \m{\psi_1}.
\ignore{
The added substitution 
\nb{is obtained by transforming \m{\sigma = \mgu(\main(\psi_2), \main(\psi_1), \vars(\psi_2))} so that the resulting substitution  is equivalent to \m{\sigma} up to variable renaming, and it conforms to the format of exception-generating substitutions. 
}}
As a matching substitution,  \m{\sigma} may contain bindings on variables of \m{\psi_2} either to variables of \m{\psi_1} or to constants. Procedure \m{\trans(\sigma, \psi_2)}, presented in Figure \ref{fig:trans}, generates an equivalent substitution, which uses only variables from \m{\vars(\psi_2)}, as only those variables can appear in the set of \m{\psi_2}'s exceptions. 
\begin{figure}[tb!]
{
\small
\begin{tabular}{rl}
\hline \\
{\bf Trans\m{(\sigma, \psi)}}&
\\
&
-- Returns part of \m{\sigma} that binds variables of \m{\psi} in exception-conformant format \\

1. &
 In \m{\sigma}, replace all groups of bindings of the form \m{v_1=v,
\ldots, v_n=v}, \\
& where \m{n>1}, \m{v_1, \ldots v_n \in\vars(\psi)},
 and  \m{v\not\in\vars(\psi)} with a set of bindings:\\
& \m{v_1=v_n, \ldots, v_{n-1}=v_n.}\\
2.&
Further, remove from  \m{\sigma} all bindings involving
variables that are not in \m{\vars(\psi)}.\\
3.& Return \m{\sigma}
 \vspace{2.mm} 
\\\hline
\end{tabular}
}
\caption{ Procedure \m{Trans(\sigma, \psi)} transforms a substitution \m{\sigma} into format suitable for the exceptions of \m{\psi}.} \label{fig:trans}
\end{figure}

Figure \ref{fig:opexamples}  contains examples of \psiform{} calculations described to this point.

When \m{\psi_1} is simple, and both \m{\psi_1} and \m{\psi_2} are fixed length, but \m{\psi_1} is not necessarily a subset of \m{\spsi{\main(\psi_2)}}, the computation of \m{\psi_2 - \psi_1} is reduced to the previous case by observing that \m{\psi_2 - \psi_1 = \psi_2 - (\spsi{\main(\psi_2)}\cap\spsi{\main(\psi_1)})}, since 
\m{\spsi{\main(\psi_2)}\cap\spsi{\main(\psi_1)}} is a subset of \m{\spsi{\main(\psi_2)}}. When \m{\spsi{\main(\psi_1)}\cap\spsi{\main(\psi_2)}} is empty, then \m{\psi_2 - \psi_1} is just \m{\psi_2}.
\ignore{
If \m{\spsi{\main(\psi_1)}\cap\spsi{\main(\psi_2)}} is empty, then \m{\psi_2 - \psi_1} is just \m{\psi_2}.  Otherwise, \m{\psi_2 - \psi_1} is \m{\psi_2} with additional exceptions accounting for \m{\spsi{\main(\psi_1)}\cap\spsi{\main(\psi_2)}}.}

\begin{figure}[tb!]
{
\centering \begin{tabular}{lp{1.4in}p{3in}l}
\hline\\
Calculation&
a,b/Operator&
Result&
 \vspace{2.mm}  \\
\hline	\\
&
\multicolumn{2}{l}{\m{a=\neg{}P(x,y)}, \m{b=\neg{}P(v,A)}}&
\\
&
\multicolumn{2}{l}{\m{\mgu(a,b,V_a) = {\set{\set{x=v,y = A}}}}}&
\\
&
&
yes  \vspace{2.mm} 
\\
&
\multicolumn{2}{l}{\m{a=\neg{}P(B,y)}, \m{b=\neg{}P(A,x)}}&
\\
&
\multicolumn{2}{l}{\m{\mgu(a,b,V_a) = \emptyset}}&
\\
\m{\spsi{b}\subseteq \spsi{a}} ?&
&
no  \vspace{2.mm} \\
\hline\\
&\multicolumn{2}{l}{
\m{a=\neg R(x,y,z,A) \vee \neg Q(t)},
\m{b=\neg R(w,C,v,A) \vee \neg Q(w)}
}
\\
&
\multicolumn{2}{l}{\m{\mgu(a,b) = \set{\set{x=w, y=C, z=v, t=w}}}} &
\\
&
\multicolumn{2}{l}{\m{\trans(\mgu(a,b), \spsi{a}) = \set{\set{x=t, y=C}}}} &
 \vspace{2.mm}  \\
\m{\spsi{a}\cap \spsi{b}}&
 &
\m{\set{\spsi{\neg{}R(t,C,z,A)\vee \neg Q(t)}}} 
 \vspace{2.mm}  \\
  \m{\spsi{a} - \spsi{b}}&	
&
\m{\apsi{\neg R(x,y,z,A) \vee \neg Q(t)}{\set{x=t, y=C}}}
\\\\
\hline
\end{tabular}\par}

\caption{Examples of the calculus computations involving set-match
(p.~\pageref{def:set-match}) and set-unification
(p.~\pageref{def:set-unify}) operators on simple \psiforms{} \spsi{a}
and \spsi{b}.}\label{fig:opexamples}

\end{figure}

Operations of \m{\cap, -} and \m{\subseteq} for two simple fixed
length \psiforms{} and membership of a clause in any \psiform{} each
take constant bounded time when the  number of clauses, variables and exceptions in a \psiform{} are constant bounded. 

\subsection{E-Difference and Image}

To capture the relations between parts of \psiforms{} necessary to formulate the \psiform{} entailment theorem, we introduce two new operations:
 the {\em image}, denoted   \m{\image{\psi_1}{\psi_2}},   is the
subset of \m{\psi_2} that {\em is\/} entailed by \m{\psi_1}, and the
{\em e-difference}, denoted   \m{\psi_2\diff \psi_1}, is the subset of
\m{\psi_2} that {\em is not\/} entailed by   \m{\psi_1}.  Thus,
\m{(\psi_2\diff \psi_1)} and \m{(\image{\psi_1}{\psi_2})} always
partition \m{\psi_2}. 

Formally, for any two sets of ground propositions $A$ and $B$,  \mydefi{
  e-difference} and \mydefi{ image} are defined respectively as follows.
\md{\begin{array}{c}
B\diff A = \{b\:|\:b\in B \widewedge
  A\not\entails b\}, \\ 
\image{A}{B} = \{ b\:|\:b\in B \widewedge A\entails b
  \}.
       \end{array}
}

\begin{ourexample}\corr{1.5}{
Let \m{A=\set{p, \neg{}q}} and \m{B=\set{p, r, \neg{}q\vee\neg{}v}}. Then, 
\m{\image{A}{B} = \set{p, \neg{}q\vee\neg{}v}} and \m{B\diff A = \set{r}}.}
\end{ourexample}

The following equivalences trivially follow from the definitions.
\begin{equation}\label{eq:DiffImg}
B\diff A  =  B - (\image{A}{B}) 
\end{equation}
\begin{equation}\label{eq:ImgDiff}
\image{A}{B}  =  B - (B \diff A)
\end{equation}

As we show later in this section, the operations \m{\image{}{}} and \m{ \diff}
applied to fixed length \psiforms{} produce sets of clauses that can 
always be represented by a {\em finite} set of fixed length \psiforms{}.

\subsection{Entailment}\label{sec:entailment}

   The next Theorem establishes the fact that a set of \psiforms{}
   \m{\Psi} entails another \psiform{} \m{\psi} if and only if for
   each clause of \m{\psi} there exists a clause in \m{\Psi} entailing it.
   The intuition behind this observation is: any two negated ground 
   clauses  \m{c_1} and  \m{c_2} {\em conjoined} entail the same set of
   negated ground    clauses as the union of  clauses entailed by
   each of \m{c_1} and \m{c_2} separately.

\providecommand{\thrmOne}{Given a set of \psiforms{} \m{\Psi = \set{\psi_1,\ldots,\psi_n}} and
  a \psiform{} \m{\psi}, \m{\Psi\entails\psi} if and only if for every
  ground clause \edit{A(5)}{\m{c \in \psi} there exists a ground clause \m{c'\in \Psi} such that \m{c'\entails c}.}}

\begin{ourtheorem}\label{th:clause-to-clause}
\thrmOne
\end{ourtheorem}

\proof (\m{\Rightarrow})\ A proof by contradiction is straightforward
  and thus omitted.  

\noindent(\m{\Leftarrow})\ Trivially follows from the definition of
  entailment.\qed

\edit{}{
The property  critical for the efficiency of \psiform{} reasoning 
is formulated in Theorem \ref{th:PreEntailmentthrm} and depicted in
Figure \ref{fig:psientfig1}:   given a set of
\psiforms{} \m{\Psi =\set{\psi_1,\ldots,\psi_n}}
\m{\Psi\entails\psi} only if there is a \psiform{} \m{\psi_i\in \Psi}
that \mydefi{nearly entails} \m{\psi}, i.e.
\m{\spsi{\main(\psi_i)}\entails\spsi{\main(\psi)}}.}

\begin{figure}[tb!]
  \centering
    \parbox[b]{\textwidth}{%
      \centering \psset{unit=.45}


\begin{pspicture}(0,0)(15,6.5)
  \newpsstyle{spsi}{fillcolor=lightgray,fillstyle=solid,
                    linestyle=solid,linewidth=.5pt}
  \newpsstyle{psiexc}{fillcolor=white,fillstyle=solid,
                    linestyle=solid,linewidth=.5pt}

  \rput [0,0](2,1) 
	{
	 \psellipse[style=spsi](0,0)(1.5,1) 
          \rput(-1.5,-.3){\rnode{psi1}{}} \rput(-2.5,-.5){\rnode{lpsi1}{\m{\psi_1}}}  \ncline[linewidth=.2pt]{lpsi1}{psi1}
	}

\rput [0,0](2.5,4.5) 
	{
	 \psellipse[style=spsi](0,0)(2, 1.7) 
          \rput(-2,0){\rnode{psi2}{}} \rput(-3, .3){\rnode{lpsi2}{\m{\psi_2}}}  \ncline[linewidth=.2pt]{lpsi2}{psi2}
	 \psellipse[style=psiexc](.1,.3)(.4,.5) 		
	 \psellipse[style=psiexc](.8,-.8)(-.3,-.2) 			
	 \rput(1,1.5){\rnode{psi2ref}{}}	
	 \rput(-1,-1.5){\rnode{psi2refb}{}}	
	}

\rput [0,0](10, 5) 
	{
	 \psellipse[style=spsi](0,0)(1.7, 1.3) 
          \rput(1.7,0){\rnode{psi3}{}} \rput(2.5, .3){\rnode{lpsi3}{\m{\psi_3}}}  \ncline[linewidth=.2pt]{lpsi3}{psi3}
	 \psellipse[style=psiexc](-.1,.3)(-.3,-.2)
	 \psellipse[style=psiexc](.8,-.8)(.4,.2)
	 \psellipse[style=psiexc](-.4,-.4)(.3,.4)
	}
  \rput [0,0](10,1) 
	{\psset{unit=.7}
         \newgray{imggray}{.55}	
	 \psellipse[style=spsi,fillcolor=imggray](0,0)(3,2) 
	 \psellipse[style=psiexc](1,0.5)(.4,.5) 		
	 \psellipse[style=psiexc](-.8,-.8)(.3,.2) 			
	\psellipse[style=psiexc](-1.2,.8)(.6,.4)
	 \rput(1.3,1.8){\rnode{psiref}{}}
         \rput(-2.1,-1.5){\rnode{psirefb}{}}	
	 \rput(2.7,0){\rnode{psi}{}} \rput(4.5,	-.3){\rnode{lpsi}{\m{\psi}}}  
	\ncline[linewidth=.2pt]{lpsi}{psi}
} 	
 \ncline{psiref}{psi2ref}
 \ncline{psirefb}{psi2refb}

\end{pspicture}
\hspace{.3in}}
\caption{For a set of \psiforms{} \m{\set{\psi_1, \psi_2, \psi_3}} to
    entail another \psiform{} \m{\psi} there must be a \psiform{} in
    the set (\m{\psi_2} in the figure) whose main part entails the
    main part of \m{\psi}. } \label{fig:psientfig1}
\end{figure}

Nearly entailment is a necessary condition for \psiform{}
entailment. As follows from Theorem
\ref{th:SubsetMatch}, entailment between two simple
\psiforms{} is  a matter of finding what we call a {\em
subset-match}, or {\em subsumption} of their main clauses.

Given two clauses \m{a} and \m{b}, where variables in \m{a}
and \m{b} are distinct and denoted \m{V_a} and \m{V_b} respectively, 
we say that \m{a} \mydefi{subsumes} \m{b} (or \m{a}
\mydefi{subset-matches} \m{b} ) if and only if there exists a
substitution \m{\sigma} on variables in \m{V_a} such that
\m{a\sigma\subseteq b}.  We denote by \m{\mgusubset(a,b,V_a)} \label{def:subset-match}the set
of all  such \m{\sigma}'s.  

Note that there can be
  more than one way a clause can subsume another clause.
  For example, matching \m{a = P(x,y)} onto \m{b =P(z,D)\vee Q(D, E)
    \vee P(A,B)}, produces two different substitutions: \m{\set{x = z,
      y = D }} and \m{\set {x = A, y = B }}, and hence
\m{\mgusubset(a,b,\set{x,y}) = \set{\set{x = z, y
          = D }, \set {x = A, y = B }}}.

\ignore{
The effect of limiting the interpretations to domains with finite number of objects is that there are other ways of showing entailment, and thus, \psiform{} reasoning remains sound but is no longer complete. }

\providecommand{\thrmTwo}{
 Given two  simple \psiforms{}, \m{\psi_1} and \m{\psi_2},
  \m{\psi_1\entails\psi_2} if and only if the main clause of \m{\psi_1}
subsumes the main clause of \m{\psi_2}, i.e. there exists a
  substitution \m{\sigma} such that
  \m{\main(\psi_1)\sigma\subseteq\main(\psi_2)} and consequently
  \m{\mgusubset(\main(\psi_1), \main(\psi_2), \vars(\psi_1))\neq
    \emptyset}.}
\begin{ourtheorem}\label{th:SubsetMatch}
\thrmTwo
\end{ourtheorem}

\proof
\noindent(\m{\Rightarrow})\ Given
    \m{\psi_1\entails\psi_2}, suppose that \md{\main(\psi_1) = \set{\neg Q_1({\vec
          x_1}), \ldots, \neg Q_n({\vec x_n})}\quad\hbox{and}\quad\main(\psi_2)
      = \set{\neg P_1({\vec y_1}), \ldots, \neg P_k({\vec y_k})}\,.}
    Assume \m{\mgusubset(\main(\psi_1), \main(\psi_2),
      \vars(\psi_1))=\emptyset}.
                  
                  This means that for any substitution \m{\sigma} on
                  variables of \m{\psi_1}, there exists a literal in
                  \m{\main(\psi_1)\sigma}, which does not match any of
                  \m{\set{\neg P_1({\vec y_1}), \ldots, \neg P_k({\vec
                        y_k})}}. We can assume without any loss of
                  generality that the mismatched literal is \m{\neg
                    Q_1({\vec x_1})}.  Note that the mismatch can
                  occur due to one of the following reasons (we also
                  call them {\em mismatch types}):
\begin{enumerate}
\item \label{case1} the predicate denoted by \m{Q_1} is not the same
  as denoted by \m{P_i},
\item \label{case2} the argument list of \m{Q_1({\vec x_1})} has a
  constant, call it a {\em matching value}, at the position where
  \m{P_i({\vec y_i})}'s argument list has a variable, call it, a {\em
    mismatched variable}.
\item \label{case3} the argument list of \m{Q_1({\vec x_1})} has a
  constant, call it \m{B_i}, at the position where \m{P_i({\vec y_i})}
  has a different constant, \m{C_i}.
\end{enumerate} 
We now construct a clause from \m{\psi_2} and show that it cannot
contain as a subclause any clause of \m{\psi_1}. According to Theorem
\ref{th:clause-to-clause} we would then contradict the fact that
\m{\psi_1 \entails \psi_2}.

Let \m{P'} be an instance of \m{\main(\psi_2)}, which is obtained by
assigning to each variable in \m{\vars(\psi_2)} a constant value which
does not occur anywhere in \m{\main(\psi_1)}. \edit{}{Since the number of constants in the language is infinite}, this can always be done. Since \m{\main(\psi_1)} does not
subset match onto \m{\main(\psi_2)}, it surely does not subset match
onto \m{P'}, because for any substitution \m{\sigma} on
\m{\vars(\psi_1)} the number of mismatches between any \m{Q} and any
literal of \m{P'} is at least the same as the number of mismatches
between a \m{Q} and a \m{P} in \m{\main(\psi_2)}, or higher, because
of the new mismatches of type \ref{case3}. Therefore, there is no
instance of \m{\main(\psi_1)} that is a subclause of \m{P'}.

\noindent(\m{\Leftarrow})\ The existence of \m{\sigma} such that
\m{\main(\psi_1)\sigma\subseteq\main(\psi_2)} means that for every
ground clause \m{c_2} of \m{\psi_2}, assuming \m{c_2 =
  \main(\psi_2)\sigma'}, there is a clause \m{c_1} in \m{\psi_1},
\m{c_1 = \main(\psi_1)\sigma\sigma'}, which is a subset of \m{c_2},
and therefore \m{\psi_1\entails\psi_2}.\qed

\ignore{
Assuming  that unification of two literals takes
constant time, determining  entailment
between two simple \psiforms{} has the  time  complexity
of computing the \m{\mgusubset(\main(\psi_1), \main(\psi_2),
\vars(\psi_1))}, which is constant when the maximum length of a \psiform{}
clause is bounded by a constant. 
}

We proceed to the necessary condition for \psiform{} entailment.
Theorem \ref{th:PreEntailmentthrm} states that in order for a set of
\psiforms{} \m{\Psi} to entail another \psiform{} \m{\psi}, there must
exist a \psiform{} in \m{\Psi} that nearly entails \m{\psi},
i.e. whose main part entails the main part of \m{\psi}.

\providecommand{\thrmThree}{
  Given a set of \psiforms{} \m{\Psi = \set{\psi_1,\ldots,\psi_n}} and
  a \psiform{} \m{\psi}, 
  \m{\Psi\entails\psi} only if there is a \psiform{} \m{\psi_i} in
  \m{\Psi} such that \m{(\spsi{\main(\psi_i)}\entails\spsi{\main(\psi)}).}
}
\begin{ourtheorem}\label{th:PreEntailmentthrm}
\thrmThree
\end{ourtheorem}

\proof
  We construct a clause of \m{\spsi{\main(\psi)}} and show that if
  none of \m{\spsi{\main(\psi_1)}, \ldots, \spsi{\main(\psi_n)}}
  entail it, then \m{\set{\psi_1, \ldots, \psi_n}} does not entail
  \m{\psi}.
  
  Suppose none of \m{\spsi{\main(\psi_1)}, \ldots,
    \spsi{\main(\psi_n)}} entail \m{\spsi{\main(\psi)}}.
  Therefore according to Theorem \ref{th:SubsetMatch} none of the main
  parts of these \psiforms{} subset match onto \m{\main(\psi)}.  Let
  \m{\sigma} be a substitution on \m{\vars(\psi)} that assigns to each
  variable a constant value that does not occur in any of \m{\psi_1,
    \ldots, \psi_n}, nor in the exceptions of \m{\psi}. This is always
  possible due to infinite number of constants in the language.  None of the
  clauses in \m{\main(\psi_1), \ldots, \main(\psi_n)} subset match
  onto \m{c=\main(\psi)\sigma}, because none of the main clauses of
  these \psiforms{} subsume \m{\main(\psi)} and the
  constants of \m{\sigma} do not appear in any of
  \m{\spsi{\main(\psi_1)}, \ldots, \spsi{\main(\psi_n)}}.  Thus,
  the clause \m{c=\main(\psi)\sigma} of \m{\psi} is not entailed by
  any clause in \m{\set{\spsi{\main(\psi_1)}, \ldots,
      \spsi{\main(\psi_n)}}}.  Since  \m{\Phi\subseteq
    \set{\spsi{\main(\psi_1)}, \ldots, \spsi{\main(\psi_n)}}},
  according to Theorem \ref{th:clause-to-clause} we conclude that
  \m{\Psi\not\entails \psi}. We arrive at a contradiction.\qed

The next example demonstrates that Theorem \ref{th:PreEntailmentthrm}
contains a necessary but not sufficient condition for the \psiform{}
entailment, and further motivates the operations of image and
e-difference. 
{
\begin{ourexample}\label{ex:nearEntailment} \label{ex:imgdiff}

Consider two \psiforms{} \m{\psi_1} and \m{\psi_2}
  below.  \ourmd{\begin{array}{l}
      \psi_1=\apsi{\neg In(x,Box1)\lor \neg Fragile(x)}{\set{x=Wine}} \\
      \psi_2=\spsi{\neg In(y,Box1) \lor \neg Fragile(y) \lor
        \neg Owner(y,Joe) } \\
    \end{array}
    } Here, \m{In(x,y)} states that  \m{x} is in \m{y},
  \m{Fragile(x)} denotes that \m{x} is a fragile item, and
 \m{Owner(x,y)}   denotes that \m{x}'s owner is \m{y}.
Thus, \m{\psi_1} states that  there are no fragile items in Box1 except for possibly a bottle of wine.
\m{\psi_2} states
  that there are  no fragile items in Box1 that are owned by \m{Joe}.
  Notice that \m{\psi_2} is simple and thus \m{\psi_2 =
    \spsi{\main(\psi_2)}}.
  
  The main clause of \m{\psi_1} subsumes the main clause of
  \m{\psi_2}, so \m{\psi_1} {\em nearly entails} \m{\psi_2}.  Therefore,
  the main part of \m{\psi_1}, \m{\spsi{\main(\psi_1)}} entails
  \m{\psi_2}, but because the exception of \m{\psi_1} weakens it,
  \m{\psi_1} does not entail \m{\psi_2}.  In fact, \m{\psi_1} entails
  {\em all} clauses of \m{\psi_2} {\em except} for the clause \m{\neg
    In(Wine,Box1) \lor \neg Fragile(Wine) \lor \neg Owner(Wine,Joe)}. 

 The only clause of \m{\psi_2} that is not entailed by
  \m{\psi_1} is \m{\neg
    In(Wine,Box1) \lor \neg Fragile(Wine) \lor \neg Owner(Wine,Joe)}, which is exactly the clause entailed by \m{\psi_1}'s
  single exception, i.e.  \md{\psi_2\diff\psi_1 =
    {\image {\spsi{\exc_1(\psi_1)}}{\psi_2}}= \spsi{\neg
    In(Wine,Box1) \lor \neg Fragile(Wine) \lor \neg Owner(Wine,Joe)}.}  The image
  \m{\image{\psi_1}{\psi_2}} is simply \m{\psi_2} with a single
  exception added: \md{{\image{\psi_1}{\psi_2}} = \apsi{\neg In(y,Box1) \lor \neg Fragile(y) \lor \neg Owner(y,Joe) }  {\set{y=Wine}}. }

So, while \m{\psi_1} {\em nearly entails} \m{\psi_2}, the e-difference
\m{\psi_2\diff\psi_1} is not empty, i.e.  \m{\psi_1} does not {\em entail}
\m{\psi_2}. This  is illustrated in Figure \ref{fig:imgdiff}.
\end{ourexample}
}
\begin{figure}[tb!] 
    \parbox[b]{\textwidth}{%
       \centering\psset{unit=0.5}


\begin{pspicture}(0,0)(15,5)

  \newpsstyle{spsi}{fillcolor=lightgray,fillstyle=solid,
                    linestyle=solid,linewidth=.5pt}
  \newpsstyle{psiexc}{fillcolor=white,fillstyle=solid,
                    linestyle=solid,linewidth=.5pt}

\rput [0,0](2.5,2) 
	{
	 \psellipse[style=spsi](0,0)(2, 1.7) 
          \rput(-2,0){\rnode{psi1}{}} 
	  \rput(-3,-.3){\rnode{lpsi1}{\m{\psi_1}}}
	  \ncline[linewidth=.2pt]{lpsi1}{psi1}
	 \psellipse[style=psiexc](.1,.3)(.2,.2) 	
  
	\rput(0,1.7){\rnode{top1}{}}
	\rput(0,-1.7){\rnode{bottom1}{}}

	\rput(.1,.5){\rnode{exctop1}{}}
	\rput(.1,.1){\rnode{excbottom1}{}}

 	\rput(.1,.3){\rnode{excctr1}{}}	
        \rput(-1.5, 3){\rnode{excclause1}
   {\m{\spsi{\exc_1(\psi_1)}=\spsi{\neg In(wine,Box1)\lor \neg Fragile(wine) }}}}	
	\ncline{excctr1}{excclause1}
 
	}

 \rput [0,0](10,2) 
	{
	 \psellipse[style=spsi](0,0)(1.5,1)
 	 \psellipse[style=spsi](.1,.3)(.2,.2) 	
	
	\rput(0,1){\rnode{top2}{}}
	\rput(0,-1){\rnode{bottom2}{}}

	\rput(.1,.5){\rnode{exctop2}{}}	
	\rput(.1,.1){\rnode{excbottom2}{}}	

 	\rput(.1,.3){\rnode{excctr2}{}}	
        \rput(1.0, -2.5){\rnode{excclause2}
			{\m{\psi_2\diff\psi_1 = 
	 \spsi{\neg In(wine,Box1)\lor \neg Fragile(wine)
				\lor \neg Owner(wine,Joe)}}}}	
	\ncline{excctr2}{excclause2}

        \rput(1.5,0){\rnode{psi2}{}}	
	\rput(2.5,-.25){\rnode{lpsi2}	{\m{\psi_2}}}  
	\ncline[linewidth=.2pt]{psi2}{lpsi2}

	}

\ncline{exctop1}{exctop2}
\ncline{excbottom1}{excbottom2}

\ncline{top1}{top2}
\ncline{bottom1}{bottom2}

\end{pspicture}
\vspace{5mm}}

\caption{Illustrates Example \ref{ex:imgdiff}. The small ellipse inside
       \m{\psi_2} represents the only clause of \m{\psi_2} not
       entailed by \m{\psi_1}, i.e. \m{\psi_2\diff\psi_1}.
         The area between the outer  and  the inner ellipsis is the
       image \m{\image{\psi_1}{\psi_2}}.}  \label{fig:imgdiff}
\end{figure}

\begin{ourtheorem} \label{th:Entailmentthrm}
\thrmFour
\end{ourtheorem}

\proof The first requirement of this Theorem follows from Theorem
  \ref{th:PreEntailmentthrm}.  While the main part of \m{\psi_k} entails \m{\psi},
  the exceptions of \m{\psi_k} weaken \m{\psi_k}.  Thus, each clause
  in \m{\psi \diff \psi_k} must be entailed by some other \psiform{}
  in \m{\set{\psi_1,\ldots,\psi_{k-1},\psi_{k+1},\ldots,\psi_n}}.\qed

Thus, in order for a set of 
\psiforms{} \m{\Psi} to entail another \psiform{} \m{\psi}, there must
exist a \psiform{} \m{\psi_k} in \m{\Psi} that  entails {\em most} of 
\m{\psi}, and the rest of  \m{\psi}, i.e. \m{\psi\diff\psi_k} must be
entailed by \m{\Psi} without \m{\psi_k}.


We have formulated the necessary and sufficient conditions for
\psiform{} entailment using e-difference.
We next present the methods of computing image and
e-difference via simple operations of subset matching and unification,
first for simple fixed length \psiforms{} and then for fixed length
\psiforms{} with exceptions. Complexity bounds  for the  computation of  entailment, image and e-difference appear in Section \ref{subsec:complexity}.

\subsection{Simple Fixed Length \psiforms{}}\label{sec:SimplePsiforms}

\begin{figure}[tb!]
{\centering \begin{tabular}{lp{1in}p{3.2in}l}
\hline\\
Calculation&
a,b/Operator&
Result&  \vspace{2.mm} 
\\
\hline	\\
&
\multicolumn{2}{l}{
\m{a=\neg{}P(x,A)},
\m{b=\neg{}P(B,y)\vee\neg P(C,z)  \vee\neg Q(y)}}
\\
&
\multicolumn{2}{l}{\m{\mgusubset(a,b,V_a) = \emptyset}}
 \vspace{2.mm} 
\\
\m{\spsi{a} \entails \spsi{b}} ?&
&
no
\\
&
\multicolumn{2}{l}{\m{\mgusubset(a,b) = \set{\set{x=B,y=A},\set{x=C,z=A}}}}&
 \vspace{2.mm} \\
\m{\image {\spsi{a}}{ \spsi{b}}}&
&
\multicolumn{2}{l}
{\m{\{\spsi{\neg P(B,A)\vee\neg P(C,z) \vee\neg Q(A)},}
}\\
&
&
\multicolumn{2}{l}
{\m{\spsi{\neg P(B,y)\vee\neg P(C,A) \vee\neg Q(y)}\}}}
 \vspace{2.mm}  \\
\m{\spsi{b}\diff\spsi{a}}&
&
\multicolumn{2}{l}{
\m{\apsi{\neg P(B,y)\vee\neg P(C,z) \vee\neg Q(y)}{\set{y = A},\set{z=A}}}}
 \vspace{2.mm}  \\
\hline
\end{tabular}\par}

\caption{Examples of the entailment, image and e-difference computations on simple \psiforms These computations utilize subset-match and subset-unification operators (defined on pages \pageref{def:subset-match} and \pageref{def:subset-unify} respectively).}\label{fig:opexamples2}

\end{figure}

In this section, we present methods of computing the operations of
\psiform{} image and e-difference for two simple fixed length
\psiforms{}.

The image
\m{\image {\psi_1} {\psi_2}} denotes a set of all clauses of \m{\psi_2}
that are entailed by \m{\psi_1}, i.e. all clauses of \m{\psi_2}  that
have a subclause in \m{\psi_1}. Thus, when \m{\psi_1} and \m{\psi_2}
are simple fixed length \psiforms{},  computing \m{\image {\psi_1} 
{\psi_2}} reduces to instantiating \m{\main(\psi_2)} with
{\em subset-unifying} substitutions, i.e. substitutions
\m{\sigma} for which
\m{\main(\psi_1)\sigma\subseteq\main(\psi_2)\sigma}.
Formally, 
we say that \m{a} \mydefi{subset-unifies}\label{def:subset-unify} with 
\m{b} if and only if there exists a substitution \m{\sigma} 
such that \m{a\sigma \subseteq b\sigma}, denoting the set of  all most
general  such \m{\sigma}'s by \m{\mgusubset(a,b)}. 

For example, consider \m{a = P(x,y)}, \m{b =P(z,D)\vee Q(D, E) \vee
    P(A,B)}, and \m{c =
    P(A,x)}.  We have \m{
 \mgusubset(a,b) = \set{\set{x = z, y = D }, \set {x = A, y = B }}}
    and \m{    \mgusubset(c,b) = \set{\set{z= A, x= D}, \set{x= B}}}.

\begin{ourtheorem}\label{th:SimpleImage-1}
\thrmFive
\end{ourtheorem}

\proof It is easy to verify equality of the two sets by showing
  inclusion both ways.\qed

Computing the e-difference, \m{\psi_2\diff\psi_1}, similar to the
regular difference  \m{\psi_2-\psi_1} (page \pageref{op:minus}),
amounts to adding exceptions to \m{\psi_2}. These exceptions represent
the set of all clauses of \m{\spsi{\main(\psi_2)}} entailed by
\m{\spsi{\main(\psi_1)}}, i.e. the image
\m{\image{\spsi{\main(\psi_1)}}{\spsi{\main(\psi_2)}}}, and are
obtained by computing the \m{\mgusubset(\main(\psi_1),\main(\psi_2))}.

\ignore{
\begin{ourexample}
\begin{sloppypar}
Let \m{\psi_1 = \spsi{\neg P(x,A)}, \psi_2 = \spsi{\neg
        P(B,y)\vee\neg P(C,z) \vee\neg Q(y,z)}}.  The main clause of
    \m{\psi_1} unifies with two disjuncts of \m{\main(\psi_2)} and
    \m{\mgusubset(\main(\psi_1),\main(\psi_2)) =
      \set{\set{x=B,y=A},\set{x=C,z=A}}}.
The image \m{\image{\psi_1}{\psi_2}}
consists of two \psiforms{}
\m{\spsi{\neg P(B,A)\vee\neg P(C,z) \vee\neg Q(A,z)}} and     
\m{\spsi{\neg P(B,y)\vee\neg P(C,A) \vee\neg Q(y,A)}}.
Thus, \md{\psi_2\diff\psi_1= \apsi{\neg
        P(B,y)\vee\neg P(C,z) \vee\neg Q(y,z)}{\set{y = A},\set{z=A}}.}
\end{sloppypar}
\end{ourexample}
}
\begin{ourtheorem}\label{th:SimpleDiff}
\thrmSix
\end{ourtheorem}

\proof
  To prove this theorem we use Theorem \ref{th:SimpleImage-1} and the
  equality \m{\psi_2 \diff \psi_1 = \psi_2 - ({\image
      {\psi_1}{\psi_2}})}.
  
\noindent According to definition (\ref{def:psiformset-b})
 \md{[\main(\psi_2)\mbox{ except }\subst(\psi_2)\cup\Sigma'] =
    \spsi{\main(\psi_2)} - \exc(\psi_2) -
    \spsi{\main(\psi_2)\sigma_1'} - \ldots -
    \spsi{\main(\psi_2)\sigma_n'}, } where \m{\Sigma' =
    \set{\sigma_1',\ldots,\sigma_n'}}. Note that
  \m{\spsi{\main(\psi_2)} - \exc(\psi_2) = \psi_2}, and that
  \m{\spsi{\main(\psi_2)\sigma'_1}\cup\ldots\cup
    \spsi{\main(\psi_2)\sigma'_n}= {\image
      {\psi_1}{\spsi{\main(\psi_2)}}}}, and thus \md{
    [\main(\psi_2)\mbox{ except }\subst(\psi_2)\cup\Sigma'] = \psi_2 -
    ({\image {\psi_1}{\spsi{\main(\psi_2)}}}). }  It remains to
  show that
\begin{equation}\label{eq:loc}
\psi_2 -  ({\image {\psi_1}{\spsi{\main(\psi_2)}}}) = 
\psi_2 -  ({\image {\psi_1}{\psi_2}}).
\end{equation}
 Indeed
\m{({\image {\psi_1}{\psi_2}}) = ({\image
    {\psi_1}{\spsi{\main(\psi_2)}}}) - ({\image
    {\psi_1}{\exc(\psi_2)}}).}  Substituting the right hand side
instead of \m{({\image {\psi_1}{\psi_2}})} in (\ref{eq:loc}), we get
\md{ \psi_2 - ({\image {\psi_1}{\spsi{\main(\psi_2)}}}) = \psi_2 -
 [ ({\image {\psi_1}{\spsi{\main(\psi_2)}}}) - ({\image
    {\psi_1}\exc(\psi_2)}) ] } Since \m{({\image
    {\psi_1}{\exc(\psi_2)}})} is in \m{\exc(\psi_2)} and therefore
definitely not in \m{\psi_2}, \md{
\psi_2 - ({\image {\psi_1}{\spsi{\main(\psi_2)}}}) =
  \psi_2 - [({\image
    {\psi_1}{\spsi{\main(\psi_2)}}}) - ({\image
    {\psi_1}{\exc(\psi_2)}})] = \psi_2 - ({\image
    {\psi_1}{\spsi{\main(\psi_2)}}}) } We have arrived at a
tautology, which proves (\ref{eq:loc}).\qed

Note that the result of the e-difference may not be a well-formed
\psiform{}.
\begin{figure}[ht!] 
\small
\begin{tabular}{c}
\hline \\
\parbox[t]{3in}{
 {\bf ComputeSimpleImg\m{(\psi_1, \psi_2)}}

-- Computes  \m{\image{\spsi{\main(\psi_1)}}  {\spsi{\main(\psi_2)}}} 

\begin{tabbing}
111\=22\=33\=44\=\kill
\>Compute \m{\Theta = \mgusubset(\main(\psi_1), \main(\psi_2))} \\
\> Set \m{\Psi = \emptyset} \\
\>For each \m{\theta_i \in \Theta} do \\
\>\> Set \m{\Psi = \Psi \cup \set{\spsi{\main(\psi_2)\theta_i}}}\\
\>Return \m{\Psi}
\end{tabbing}
}
\        \ 
\parbox[t]{2.7in}{
 {\bf ComputeSimpleEDiff\m{(\psi_2,\psi_1)}}

-- Computes  \m{\psi_2\diff  {\spsi{\main(\psi_1)}}} 
\begin{tabbing}
111\=22\=33\=44\=\kill
\> If \m{\mgusubset(\main(\psi_1), \main(\psi_2),\vars(\psi_1))\neq\emptyset} \\
\>\>  Return \m{\emptyset}. \\
\> Compute \m{\Theta = \mgusubset(\main(\psi_1), \main(\psi_2))} \\	
\> \m{\Sigma' = \emptyset}\\
\>For each \m{\theta_i \in \Theta} do \\	
\>\> Set \m{\Sigma' = \Sigma' \cup Trans(\theta_i, \psi_2)}\\
\>Return \m{\set{[\main(\psi_2)\mbox{ except }\Sigma(\psi_2)\cup\Sigma']}}.
\vspace{2mm}
\end{tabbing}
}
\\\hline
\end{tabular}

\caption{ Image and e-difference operations for simple \psiforms{}. \m{\textit{ComputeSimpleImg}(\psi_1, \psi_2)} and
\m{\textit{ComputeSimpleEDiff}(\psi_1,\psi_2)} return
\m{\image{\spsi{\main(\psi_1)}}  {\spsi{\main(\psi_2)}}}  and 
\m{\psi_2\diff  {\spsi{\main(\psi_1)}}}  respectively.
}\label{fig:computesimpleimgdiff}
\end{figure}

\begin{sloppypar}
Figure \ref{fig:opexamples2} presents examples of computing image and e-difference between simple \psiforms, and 
Figure  \ref{fig:computesimpleimgdiff} presents algorithms for these computations, based  on Theorems \ref{th:SimpleImage-1} and
\ref{th:SimpleDiff}. 
\end{sloppypar}

\subsection{Arbitrary Fixed Length \psiforms{}}
{
In this section, we present methods of computing the operations of
\psiform{} image and e-difference for two arbitrary fixed length
\psiforms{}. 

Let \m{\psi_1} and  \m{\psi_2} be arbitrary fixed length \psiforms{}.
To find either the image or the e-difference we first find
the image of the main part of \m{\psi_1} onto the main part of \m{\psi_2}. 
Since the exceptions of \m{\psi_1} weaken it, we must then calculate the
part of \m{\spsi{\main(\psi_2)}} that is not entailed by \m{\psi_1}
due to the exceptions. We'll call this a set of ``holes'' (
denoted by \m{H(\psi_1, \psi_2)}). 

Formally, we define \mydefi{set of holes \m{H(\psi_1, \psi_2)}} as follows
\begin{equation}\label{def:holes}
{H(\psi_1,\psi_2)} =
({\image{\spsi{\main(\psi_1)}}{\spsi{\main(\psi_2)}} } ) -
({\image{\psi_1}{\spsi{\main(\psi_2)}} }) ,
\end{equation}  
i.e. holes are parts of \m{\spsi{\main(\psi_2)}} that are entailed by \m{\spsi{\main(\psi_1)}}, but not by  \m{\psi_1}.

Image and e-difference operations are easily formulated using 
\m{H(\psi_1, \psi_2)}.
The image of \m{\psi_1} onto \m{\psi_2} consists of clauses of the main
part of \m{\psi_2} entailed by  the main
part of \m{\psi_1}, i.e. \m{\image{\spsi{\main(\psi_1)}}{\spsi{\main(\psi_2)}}}, minus the set of holes
\m{H(\psi_1,\psi_2)} and minus \m{\psi_2}'s own exceptions.
Similarly, the e-difference  \m{ \psi_2 \diff \psi_1} consists of the
part of the main  part of \m{ \psi_2 }, \m{\spsi{\main(\psi_2)}},
not entailed by \m{\spsi{\main(\psi_1)}}},
i.e. \m{\spsi{\main(\psi_2)} \diff\spsi{\main(\psi_1)}} plus the
set of holes 
\m{H(\psi_1,\psi_2)}, minus the set of \m{\psi_2}'s  exceptions.
These two facts are presented in the following two Lemmas.
\begin{ourlemma}\label{lm:Image}
\lemmaThree
\end{ourlemma}

\proof
  Let \m{A={\image {\spsi{\main(\psi_1)}}
      {\spsi{\main(\psi_2)}}}}.  We substitute the definition of \m{
    H(\psi_1,\psi_2)} from (\ref{def:holes}) on the right hand side.
\md{{\image{\psi_1}{\psi_2}} =  A-(A-({\image{\psi_1}{\spsi{\main(\psi_2)}} }))-
    \exc(\psi_2)}
  When \m{X,Y} and Z denote arbitrary sets, we have
\begin{equation}\label{eq:sets1}
X-(Y-Z) = (X-Y)\cup(X\cap Y\cap Z),
\end{equation} so
\md{\eqalign{
     \image{\psi_1}{\psi_2} 
={}& {(A-A)\cup(A\cap A \cap ({\image{\psi_1}{\spsi{\main(\psi_2)}} }))} -
     \exc(\psi_2) \cr
={}& (A\cap ({\image{\psi_1}{\spsi{\main(\psi_2)}} })) - \exc(\psi_2)
     \cr
={}& ({\image{\psi_1}{\spsi{\main(\psi_2)}} }) - \exc(\psi_2) \cr
={}&  {\image{\psi_1}{(\spsi{\main(\psi_2)} - \exc(\psi_2) )} } \cr
={}&  {\image{\psi_1}{\psi_2}} 
}} 
\qed

\begin{ourlemma}\label{lm:Diff}
\lemmaFour
\end{ourlemma}

\proof
  From Lemma \ref{lm:Image} and equivalence (\ref{eq:DiffImg}) we have
  \m{\psi_2\diff\psi_1 = \psi_2-(\image{\psi_1}{\psi_2})}, or
  \md{\begin{array}{c}

\psi_2 \diff \psi_1 = 
\underbrace{\psi_2}_A - 
[(\underbrace{({\image {\spsi{\main(\psi_1)}} {\spsi{\main(\psi_2)}}})}_B - \underbrace{H(\psi_1,\psi_2)}_C) - \underbrace{\exc(\psi_2)}_D] 
    \end{array}}
Using (\ref{eq:sets1}), we rewrite the right hand side equivalently \md{
  \psi_2\diff\psi_1= A-((B-C)-D) = (A-(B-C))\cup(A\cap (B-C) \cap D) }
Since in our case \m{A\cap D =\emptyset}, therefore \m{(A\cap (B-C)
  \cap D)= \emptyset} and we get \md{ \psi_2 \diff \psi_1 = (A-(B-C))
  = (A-B)\cup(A\cap B \cap C)}
We now evaluate \m{A\cap B\cap C}.  We note that \m{A\cap B=
  ([\main(\psi_2)] - D)\cap B } and since \m{(X-Y)\cap Z = X\cap Z -
  Y\cap Z}, we have \md{A\cap B= [\main(\psi_2)]\cap B - D\cap B =
  B-B\cap D = B-D.}
Next, \m{(B-D)\cap C = B\cap C -D\cap C} and since \m{C\subseteq B},
\m{(B-D)\cap C = C -D\cap C = C-D,} so we get
\md{\eqalign{
     \psi_2\diff\psi_1
={}& (A-B)\cup(C-D) \cr
={}& (\psi_2 - ({\image {\spsi{\main(\psi_1)}}
      {\spsi{\main(\psi_2)}}})) \cup ( H(\psi_1,\psi_2) -
     \exc(\psi_2)) \cr
={}& (\spsi{\main(\psi_2)} - \exc(\psi_2) - ({\image
      {\spsi{\main(\psi_1)}}
      {\spsi{\main(\psi_2)}}})) \cup ( H(\psi_1,\psi_2) -
     \exc(\psi_2)) \cr
={}& ({\spsi{\main(\psi_2)}} \diff {\spsi{\main(\psi_1)}} - \exc(\psi_2))
    \cup ( H(\psi_1,\psi_2) - \exc(\psi_2)) \cr
={}& (({\spsi{\main(\psi_2)}} \diff {\spsi{\main(\psi_1)}}) \cup
    H(\psi_1,\psi_2)) - \exc(\psi_2).
}}
\qed

  We calculate \m{\image {\psi_1} {\psi_2}} and
\m{{\psi_1}\diff {\psi_1}} separately in each of the following three
cases 

\noindent{\bf{}Case  1:}
\m{\mgusubset(\main(\psi_1), \main(\psi_2)) = \emptyset},
i.e. the image
  \m{\image{\spsi{\main(\psi_1)}}{\spsi{\main(\psi_2)}}} is
  empty.

\noindent{\bf{}Case  2:}
\m{\mgusubset(\main(\psi_1), \main(\psi_2),\vars(\psi_1))\neq
    \emptyset}, i.e. by Theorem
  \ref{th:SubsetMatch}, \m{\spsi{\main(\psi_1)}\entails\spsi{\main(\psi_2)}}
and hence the image
  \m{\image{\spsi{\main(\psi_1)}}{\spsi{\main(\psi_2)}}} equals 
the entire \m{\spsi{\main(\psi_2)}}.
  
\noindent{\bf{}Case  3:}
\m{\mgusubset(\main(\psi_1), \main(\psi_2)) \neq \emptyset}, i.e. the image  \m{\image{\spsi{\main(\psi_1)}}{
    \spsi{\main(\psi_2)}}} is non-empty. 

Cases  1 and 3 are complementary. However we have separated 
case 2, which  is a specific subcase of 3, because it comes up while
deciding \psiform{} entailment (see Theorem
\ref{th:Entailmentthrm}). Moreover, case 3    
is reduced to case 2, as we will demonstrate.

Case 1 is the simplest and is covered by Theorem \ref{th:emptyImage}.

\begin{ourtheorem}\label{th:emptyImage}
\thrmSeven
\end{ourtheorem} 

\proof  As follows from Theorem \ref{th:SimpleImage-1},   
\m{\mgusubset(\main(\psi_1), \main(\psi_2)) = \emptyset} implies that
the image  \m{\image {\spsi{\main(\psi_1)}} {\spsi{\main(\psi_2)}}}
is empty. Since \m{(\image{\psi_1}{\psi_2} )\subseteq ({\image
{\spsi{\main(\psi_1)}} {\spsi{\main(\psi_2)}}})}, we have that 
\m{\image{\psi_1}{\psi_2} = \emptyset}  and therefore  
by equivalence (\ref{eq:DiffImg}) \m{\psi_2 \diff \psi_1 =  \psi_2}.\qed

Case 2 amounts to \m{\psi_1} nearly entailing \m{\psi_2}. 

\begin{ourtheorem}\label{th:all1}
\thrmEight
\end{ourtheorem}

\proof
  (\ref{th:case2Img}) and (\ref{th:case2Diff}) trivially follow from
  Lemma \ref{lm:Image} and definition (\ref{def:holes}) by substituting
  \m{\spsi{\main(\psi_2)}} in place of
  \m{\image{\spsi{\main(\psi_1)}}{\spsi{\main(\psi_2)}}} and
  substituting \m{\emptyset} in place of
  \m{{\spsi{\main(\psi_1)}}\diff{\spsi{\main(\psi_2)}}}.\qed

The expression for the set of holes \m{ H(\psi_1, \psi_2)} in this
case is derived in Lemma \ref{lm:aux}.
We first demonstrate the computation of the set of holes, image and e-difference in the following example.  

The algorithm is straightforward when there is only one
subset-unifier of \m{\main(\psi_1)} with 
\m{\main(\psi_2)}, i.e. each clause of \m{\image
{\spsi{\main(\psi_1)}} {\spsi{\main(\psi_2)}}}
is entailed by {\em exactly one} clause of \m{ \spsi{\main(\psi_1)}}.
The set of holes in this case is simply the union of images from each
exception of \m{\psi_1} onto \m{\spsi{\main(\psi_2)}}.
\begin{ourexample}\label{ex:holesSimple}
Consider 
\md{\begin{array}{c}
      \psi_1 = \apsi{\neg P(x, y, z)}{\set{x=B},\set{x=C,y=D},\set{x=A}}\\
      \psi_2 = \apsi{\neg P(w, E, A)}{\set{w=G}}
\end{array}
}
Since there is only one
subset-unifier of \m{\main(\psi_1)} with 
\m{\main(\psi_2)},
the image of  \m{\psi_1}  onto \m{\psi_2} is simply  the image
\m{\spsi{\main(\psi_1)}} onto \m{\spsi{\main(\psi_2)}} minus 
exceptions of \m{\psi_2} and the image of exceptions of  \m{\psi_1}
on \m{\spsi{\main(\psi_2)}}, i.e.
\md{
\image { \psi_1 } {\psi_2} = {\image
{\spsi{\main(\psi_1)}} {\spsi{\main(\psi_2)}}} - \exc(\psi_2) -
\bigcup_{i=1}^{3}(\image {\spsi{\exc_i(\psi_1)}}{\spsi{\main(\psi_2)}} )
}

In this case 
\m{ \image {\spsi{\main(\psi_1)}} {\spsi{\main(\psi_2)}} =
{\spsi{\main(\psi_2)}} } and, since (by definition
(\ref{def:psiformset-b}))
\m{\psi_2 ={\spsi{\main(\psi_2)}} - \exc(\psi_2)},
\md{\eqalign{
  \image { \psi_1 } {\psi_2} 
={}&  \apsi{\neg P(w, E, A)}{\set{w=G}}
      - \image {\spsi{\neg P(B, y, z)}}{\spsi {\neg P(w, E, A)}} \cr 
   &  - \image {\spsi{\neg P(C, D, z)}}{\spsi {\neg P(w, E, A)}}  
        \image {\spsi{\neg P(A, y, z)}}{\spsi {\neg P(w, E, A)}} \cr
={}&  \apsi{\neg P(w, E, A)}{\set{w=G}} - {\spsi {\neg P(B, E,A)}} 
      - {\spsi {\neg P(A, E, A)}} \cr
={}&  \apsi{\neg P(w, E, A)}{\set{w=G},\set{w = B},\set{w = A}}\cr
}}
\end{ourexample}

Computing the holes is more complex  when there is more than  one
subset-unifier  of \m{\main(\psi_1)} with 
\m{\main(\psi_2)}, because in this case some clauses of \m{\image
{\spsi{\main(\psi_1)}} {\spsi{\main(\psi_2)}}} are entailed by
{\em more than one} clause of \m{ \spsi{\main(\psi_1)}}. Then, even
though an exception removes from \m{\spsi{\main(\psi_1)}} an
entailing clause for some clause \m{c} of \m{\psi_2}, \m{c} may be
entailed by another clause in \m{\psi_1}, and consequently the set of
holes is not simply a set of images from \m{\psi_1}'s exceptions, but
rather an intersection of such images.   Example \ref{ex:holesComplex} illustrates  this computation.

\begin{figure}[ht!]
\begin{tabular}{c}
\hline \\
\parbox[t]{3.2in}{\small {\bf Compute \m{H(\psi_1, \psi_2)}}

-- Requires that  \m{\psiset{\main(\psi_1)} \entails
  \psiset{\main(\psi_2)}} 
\begin{tabbing}
111\=22\=33\=\kill
\>Compute \m{\Theta = \mgusubset(\main(\psi_1),
    \main(\psi_2), \vars(\psi_1))} \\
\>Set \m{\Psi = \emptyset} \\
\>For each \m{i} from 1 to \m{\norm{\Theta}} do \\

\>\> Set \m{\psi_1^i = \psiset{\main(\psi_1)\theta_i}} \\
\>\> Set \m{\Psi_2 = \emptyset}\\

\>\> For each \m{j} from 1 to \m{\norm{\Sigma(\psi_1)}} do\\
\>\>\> Set \m{I_{ij} =  ({\image {(\psiset{\exci{j}(\psi_1)}\cap \psi_1^i
            )} {\psiset{\main(\psi_2)}}})}\\	
\>\>\> Set \m{\Psi_2 = \Psi_2 \cup I_{ij}} \\

\>\> If \m{\Psi_2 = \emptyset} Then Return \m{\emptyset} \\
\>\> If \m{i=1} Then  Set \m{\Psi = \Psi_2}\\
\>\>\>Else Set \m{\Psi = \Psi \cap \Psi_2}. \\

\>Return \m{\Psi}

\end{tabbing}
} \  \ 
\parbox[t]{2.3in}{\small {\bf ComputeImg2\m{(\psi_1,\psi_2)}}

-- Requires that  \m{\psiset{\main(\psi_1)} \entails \psiset{\main(\psi_2)}}
\begin{tabbing}
111\=22\=33\=\kill
\>Set \m{\Psi_H =Compute H(\psi_1, \psi_2)}, \m{\psi = \psi_2}\\
\>For each simple \psiform{} \m{\psi_h\in\Psi_H}\\
\>\> Set \m{\psi = \psi - \psi_h}\\
\> Return   \m{\set{\psi}}\\
\end{tabbing}
{\bf ComputeEDiff2\m{(\psi_2,\psi_1)}}

-- Requires that  \m{\psiset{\main(\psi_1)} \entails
  \psiset{\main(\psi_2)}}
\begin{tabbing}
111\=22\=33\=\kill
\>Set \m{\Psi_H =Compute H(\psi_1, \psi_2)}, \m{\Psi = \emptyset} \\
\>For each simple \psiform{} \m{\psi_h\in\Psi_H}\\
\>\> Set \m{\psi = \psi_h} \\
\>\> For each simple \psiform{} \m{\psi_e\in\exc(\psi)}\\
\>\>\>   Set \m{\psi= \psi - \psi_e }\\
\>\>   Set \m{\Psi = \Psi \cup \set{\psi} }\\
\> Return   \m{\Psi}\\
\end{tabbing}

}

\\\hline
\end{tabular}
\caption{ Computing the set of holes, image and e-difference
operations in case \m{\psi_1} nearly entails  \m{\psi_2}.
}\label{fig:computeHoles}

\end{figure}


We derive an expression for calculating  \m{H(\psi_1,\psi_2)} in
the next Lemma
by first  introducing the set of \psiforms{} 
within \m{\spsi{\main(\psi_1)}} (denoted \m{\Psi_1}), all of which entail
some part of \m{\spsi{\main(\psi_2)}}, and showing how to combine them
to calculate \m{H(\psi_1,\psi_2)}.

\begin{ourlemma}\label{lm:aux}
\lemmaFive
\end{ourlemma}

\proof
 Each \m{\psi_1^i} entails
  \m{\spsi{\main(\psi_2)}}, i.e.  \m{\image {\psi_1^i}{
      \spsi{\main(\psi_2)}} = {\spsi{\main(\psi_2)}}} for each \m{1\leq
    i\leq{\norm{\Theta}}}, because the main clause of each \m{\psi_1^i}
equals some subset of literals of \m{\main(\psi_2)}. However,
 each \m{\psi_1^i} may contain clauses that are exceptions of
  \m{\psi_1}, namely \m{\bigcup_{j=0}^{\norm\Sigma(\psi_1)}
  {(\spsi{\exci{j}(\psi_1)}\cap \psi_1^i)}}. We call these clauses 
{\em \m{\psi_1}'s exceptions in  \m{\psi_1^i} }.

  Thus, \m{\setpsi(\Psi_1)} contains all clauses of \m{\psi_1} that entail
  something in \m{\spsi{\main(\psi_2)}} and more, namely
  \m{\psi_1}'s exceptions in  \m{\psi_1^i}.  Therefore  
  \md{{\image {\psi_1}{\spsi{\main(\psi_2)}}} =   
  \bigcup_{\psi_1^i\in\Psi_1}({\image {({\psi_1^i} - \cup_{j=0}^{\norm{\Sigma(\psi_1)}}(\spsi{\exci{j}(\psi_1)}\cap
  \psi_1^i))} {\spsi{\main(\psi_2)}}}).}  
 Note that for a given \m{\psi_1^i\in\Psi_1}, each clause of
  \m{{ \spsi{\main(\psi_2)}}} is entailed by exactly
  one clause of \m{\psi_1^i}, i.e. for every \m{\psi_1^i\in\Psi_1}, and
  every \m{c\in  \spsi{\main(\psi_2)}, (\psi_1^i
    \entails c) \Leftrightarrow \ourexists{!c_1 \in \psi_1^i}
    {c_1\entails c}}
  (existence of a clause 
   \m{ \main(\psi_2)\sigma} such that there are 2 different clauses
  \m{c',c''\in\psi_1^i} that entail it, leads to a contradiction to the
  fact that \m{\psi_2} and \m{\psi_1} are fixed length \psiforms).
 The last observation allows to distribute the image operator and rewrite the last   expression as follows
\md{\eqalign{
     \image {\psi_1}{\spsi{\main(\psi_2)}}
={}& \bigcup_{\psi_1^i\in\Psi_1}\Bigl(({\image {\psi_1^i}{\spsi{\main(\psi_2)}}}) -
 {\image {\bigcup_{j=0}^{\norm{\Sigma(\psi_1)}}(\spsi{\exci{j}(\psi_1)}\cap
  \psi_1^i)} {\spsi{\main(\psi_2)}} }\Bigr) \cr
={}& {\spsi{\main(\psi_2)}}  - \bigcap_{\psi_1^i\in\Psi_1}\bigcup_{j=0}^{\norm{\Sigma(\psi_1)}} \bigl({\image
{(\spsi{\exci{j}(\psi_1)}\cap \psi_1^i )}
{\spsi{\main(\psi_2)}}}\bigr).
}
} i.e. the set of clauses of \m{\spsi{\main(\psi_2)}} not entailed by
 \m{\psi_1}  is  the {\em intersection} of images
 of \m{\psi_1}'s exceptions  in {\em all} of \m{\psi_1^i\in\Psi_1},

The formula for \m{H(\psi_1, \psi_2)} follows from substituting the
derived expression for \m{\image {\psi_1}{\spsi{\main(\psi_2)}}} in
the definition (\ref{def:holes}) and noticing that since \m{\psi_1}
nearly entails \m{\psi_2},
\m{\image{\spsi{\main(\psi_1)}}{\spsi{\main(\psi_2)}} =
{\spsi{\main(\psi_2)}}}.\qed

The procedure for computing the set of holes  \m{H(\psi_1,\psi_2)} in case \m{\psi_1} nearly entails \m{\psi_2} presented  in Figure \ref{fig:computeHoles}
is based on Lemma \ref{lm:aux}. 
Procedures for computing the image and e-difference in case \m{\psi_1} nearly entails \m{\psi_2} are also presented in Figure \ref{fig:computeHoles}.

\begin{ourexample}\label{ex:holesComplex}
Consider 
\md{\begin{array}{c}
      \psi_1 = \apsi{\neg P(x, y, z)}{\set{x=B},\set{x=C, y=D},\set {x=A}}\\
      \psi_2 = \apsi{\neg P(w, E, A)\vee\neg P(C, D, w )\vee \neg
        Q(w)}{\set{w=G}}
\end{array}
}

Here, the clause \m{c = \neg P(K, E, A)\vee\neg P(C, D, K )\vee \neg
Q(K)}, for example, 
 is entailed by two clauses of \m{\spsi{\main(\psi_1)}}, namely, by  \m{\neg
P(K, E, A)} and \m{\neg{}P(C, D, K )}. Although the second of these clauses
is not in \m{\psi_1} due to the second exception, the first one, \m{\neg P(K, E, A)}
is in \m{\psi_1}  and therefore 
\m{\psi_1 \entails \neg P(K, E, A)\vee\neg P(C, D, K )\vee \neg Q(K)}. 
Thus, even though  \m{c\in \image{\spsi{\exc_2(\psi_1)}}{\psi_2}}, \m{c \in
\image{\psi_1}{\psi_2}}.

Computing the set of holes according to Lemma \ref{lm:aux} yields 
\md{\begin{array}{c}H(\psi_1,\psi_2) = 
({\image {(\exc(\psi_1)\cap\psi_1^1)} {\spsi{\main(\psi_2)}}})\cap
({\image {(\exc(\psi_1)\cap\psi_1^2)} {\spsi{\main(\psi_2)}}}) = \\
\left\{\begin{array}{c}
 {\spsi{\neg P(B, E, A)\vee\neg P(C, D, B )\vee \neg  Q(B) }}, \\
{\spsi{\neg P(A, E, A)\vee\neg P(C, D, A )\vee \neg  Q(A)}} 
    \end{array}
\right\} .
    \end{array}
}

Furthermore, according to Theorem \ref{th:all1},
 e-difference \m{\psi_2\diff\psi_1} equals the set of holes
\m{H(\psi_1,\psi_2)} minus exceptions of \m{\psi_2}
\md{\begin{array}{c}
\psi_2\diff\psi_1 = H(\psi_1,\psi_2) 
 - 
\set{{\spsi{\neg P(G, E, A)\vee\neg P(C, D, G )\vee \neg  Q(G)}}} =\\ 
\left\{\begin{array}{c}
{\spsi{\neg P(B, E, A)\vee\neg P(C, D, B )\vee \neg  Q(B) }}, \\
{\spsi{\neg P(A, E, A)\vee\neg P(C, D, A )\vee \neg  Q(A)}} 
    \end{array}
\right\} .
    \end{array}
}

\noindent The image  \m{\image {\psi_1} {\psi_2}} equals
\m{{\spsi{\main(\psi_2)}}}  minus the set of holes, and minus
exceptions of \m{\psi_2}, i.e. 
%
%
\md{\eqalign{
     \image {\psi_1} {\psi_2} 
={}& {\psi_2} - {H(\psi_1,\psi_2)} \cr
={}& \apsi{\neg P(w, E, A)\vee\neg P(C, D, w )\vee \neg
        Q(w)}{\set{w=G}} - {H(\psi_1,\psi_2)} \cr
={}& \apsi{\neg P(w, E, A)\vee\neg P(C, D, w )\vee \neg
        Q(w)}{\set{w=G},\set{w=A}, \set{w=B}}
}}
\end{ourexample}


Recall that the computation of e-difference in case
\m{\spsi{\main(\psi_1)}\entails\spsi{\main(\psi_2)}} comes up in 
verifying entailment (Theorem \ref{th:Entailmentthrm}).  The following
Observation guarantees that each \psiform{} in the e-difference
\m{\psi_2\diff\psi_1} is strictly ``smaller'' than \m{\psi_2}. This
observation plays a critical role in establishing the complexity
bounds on \psiform{} reasoning.

\begin{ourobservation}\label{lm:varnum}
\lemmaSix
\end{ourobservation}

\proof As follows from Theorem \ref{th:all1} and Lemma \ref{lm:aux}
  the e-difference is a subset of a union of images of exceptions of
  \m{\psi_1} on \m{\spsi{\main(\psi_2)}}. Each such image is obtained
  by instantiating the main clause \m{\main(\psi_2)} with a subset
  unifying substitution, call it \m{\sigma}. When \m{\sigma} does not
  bind any variables of \m{\psi_2} to constants, the image is equal to
  \m{\spsi{\main(\psi_2)}}, which contradicts the conditions of the
  Observation.  Thus, \m{\sigma} must bind some variables of
  \m{\main(\psi_2)} to constants, and therefore
  \m{\spsi{\main(\psi_2)\sigma}}, and in turn, every subset of this
  \psiform{} is expressed with a \psiform{} that contains strictly
  fewer variables than \m{\spsi{\main(\psi_2)}}.\qed

  We now consider case 3.  There is a non-empty image of the main part
  of \m{\psi_1} onto the main part of \m{\psi_2} which occurs when
  \m{\mgusubset(\main(\psi_1),\main(\psi_2)) \neq\emptyset}.  In this
  case we first compute the image \m{\image
  {\spsi{\main(\psi_1)}}{\spsi{\main(\psi_2)}}}, denoted below by
  \m{\Psi}. Every \psiform{} in \m{\Psi} is nearly entailed by
  \m{\psi_1}, and thus we can compute the image of \m{\psi_1} on each
  of \psiforms{} in \m{\Psi} using the methods of Case 2. The image
  \m{\image{\psi_1} {\psi_2}} equals the union of images of \m{\psi_1}
  onto each \psiform{} in \m{\Psi}, minus exceptions of \m{\psi_2}.

\begin{ourtheorem}\label{th:FLImageDiff}
\thrmNine
\end{ourtheorem}

\proof
  By Theorem \ref{th:SimpleImage-1} \m{\mgusubset(\main(\psi_1),\main(\psi_2))\neq\emptyset} implies that 
the image \m{\Psi} 
is non-empty and thus consists of a set of simple \psiforms{}. Each 
\psiform{} in \m{\Psi} is nearly entailed by \m{\psi_1}. 
The image \m{\image {\psi_1} {\psi_2}} is a subset of \m{\Psi}, and
equals exactly the set of all clauses in \m{\Psi} that are not exceptions of
\m{\psi_2} and that are entailed by \m{\psi_1}, i.e.
\md{{\image {\psi_1} {\psi_2} = ({\image {\psi_1} {\Psi}}) - \exc(\psi_2)
}.} Since each of \psiforms{} in \m{\Psi} is nearly entailed by
\m{\psi_1} the calculation of the image \m{\image {\psi_1} {\Psi}}  in the above expression can be
carried out according to Theorem \ref{th:all1}.

The proof of (\ref{th:FLDiff}) is similar. The part of \m{\psi_2} that
is not entailed by \m{\psi_1} includes \m{\psi_2 - \Psi} plus parts of 
\m{\Psi} that are not exceptions of \m{\psi_2} and are not entailed by \m{\psi_1}, i.e. \m{(\Psi  - \exc(\psi_2)) \diff {\psi_1}}. 
\qed

The procedures for computing image and e-difference in case 3 are 
given in Figure \ref{fig:alg3}.

\begin{ourtheorem}\label{th:thelast}
\thrmTen
\end{ourtheorem}

\proof
  The fact that all operations produce sets of \psiforms{} follows
  from the fact that all of them produce subsets of operand \psiforms.
  The fact that indeed this set is finite follows from the Theorems
  \ref{th:emptyImage}, \ref{th:FLImageDiff}.
  
  The resulting \psiforms{} are fixed length, because they contain
  clauses from argument \psiforms{}, and each subset \psiform{} of a
  fixed length \psiform{} is obviously a fixed length \psiform{}.\qed

\begin{figure}[ht!]
\small
\begin{tabular}{c}
\hline \\
\parbox[t]{3.3 in}{
{\bf{ComputeImg3\m{(\psi_1, \psi_2)}}}

-- Requires  \m{\mgusubset(\main(\psi_1), \main(\psi_2)) \neq \emptyset} 
\begin{tabbing}
111\=22\=33\=44\=\kill
\> Set \m{\Psi = ComputeSimpleImg(\psiset{\main(\psi_1)},
 					\psiset{\main(\psi_2)})}\\
\> Set \m{\Psi_r = \emptyset}\\
\> For each \m{\psi \in \Psi}\\
\>\>  Set \m{\psi_r = ComputeImg2(\psi_1, \psi)}\\
\>\>  Set \m{\Psi_r = \Psi_r \cup \psi_r}\\
\>Set  \m{\Psi_r = \Psi_r  - \exc(\psi_2)}\\
\>Return \m{\Psi_r}
\end{tabbing}
}
 \   \ 
\parbox[t]{2.4in}{
 {\bf{ComputeEDiff3\m{(\psi_2,\psi_1)}}}

-- Requires  \m{\mgusubset(\main(\psi_1), \main(\psi_2)) \neq \emptyset}
\begin{tabbing}
11\=22\=33\=44\=\kill
\> Set \m{\Psi =} {ComputeSimpleImg}\\ 
\>\>\>	\m{(\psiset{\main(\psi_1)},
 					\psiset{\main(\psi_2)})}\\
\> Set \m{\Psi_r = {\psi_2 - \Psi}}\\
\> For each \m{\psi \in \Psi}\\
\>\>  Set \m{\psi = \psi - \exc(\psi_2) }\\
\>\>  Set \m{\psi_r =\textit{ComputeEDiff2}(\psi, \psi_1) }\\
\>\>  Set \m{\Psi_r = \Psi_r \cup \psi_r}\\
\>Return \m{\Psi_r}
\vspace{2mm}
\end{tabbing}
}
\\\hline
\end{tabular}
\caption{Procedures \m{\textit{ComputeImg3}(\psi_1, \psi_2)} and
\m{\textit{ComputeEDiff3}(\psi_1,\psi_2)} compute \m{\image
{\psi_1}{\psi_2}} and  \m{\psi_2\diff\psi_1} in case there is a
non-empty image of the main part of \m{\psi_1} onto the main part of
\m{\psi_2}, i.e. \m{\image {\psiset{\main(\psi_1)}}
{\psiset{\main(\psi_2)}}\neq \emptyset}
} 
\label{fig:alg3} 
\end{figure}

\subsection{Complexity of \psiform{} operations}\label{subsec:complexity}

The  recursive procedure  for determining entailment \m{\Psi\entails\psi} based on  Theorem \ref{th:Entailmentthrm} takes time \m{\bigo(n)}, where \m{n}
is the number of \psiforms{} in \m{\Psi}, when the maximum number of
exceptions, and variables and literals in the main clause of a
\psiform{} are fixed.     We  assume unification takes constant
bounded time, which is guaranteed when the cardinality of predicate
symbols is bounded by a constant.  
These assumptions are common in open world applications: 
\begin{itemize}
\item the number of variables and literals in the main clause
and cardinality of predicate symbols are always finite and bounded by the specification of the initial and goal states and the action descriptions.
Moreover, they are typically small. 
\item the number of exceptions is limited by a function of the number of objects known in the initial state and those objects created by the actions in a constructed plan. When the length of the plan is constant bounded, the number of exceptions is therefore also constant bounded. In general, the complexity of entailment is polynomially bounded in the maximum number of exceptions, as presented in Figure \ref{fig:complexities} and discussed briefly at the end of this section.
\end{itemize}

To obtain the linear bound on the complexity of entailment, notice that finding a \psiform{} that nearly entails \m{\psi} requires a pass through at most  \m{n} \psiforms{} of \m{\Psi} spending constant time at each, since checking
nearly entailment takes constant time. Once a nearly entailing
\psiform{} \m{\psi_k} is found, we calculate the difference
\m{\psi\diff\psi_k}  and apply  Theorem (\ref{th:Entailmentthrm})  to each
\psiform{} in the 
e-difference.  This is a recursive procedure, which can be represented by
a recursion tree. In the tree, each node represents the
non-recursive computation, i.e.  finding a nearly
entailing \psiform{}  \m{\psi_k}, and computing the e-difference
\m{\psi\diff \psi_k}; and each branch represents a
recursive call to the same procedure for checking entailment of each
\psiform{} in the e-difference \m{\psi\diff\psi_k}. 
The complexity of the entire procedure equals the sum of the complexities at
the nodes of the recursion tree.
The time spent at each node is  proportional to the number of
\psiforms{} in \m{\Psi}, which is \m{n} at the root of the tree, and
decreases by one at each subsequent level.
The branching factor \m{\beta} at each node equals the number of
\psiforms{} in the e-difference \m{\psi\diff\psi_k}. \m{\beta} 
is constant bounded when we bound by constants the maximum number of
exceptions, variables and literals in the \psiforms{}. 
Therefore, at each
level \m{i} of the tree we have at most  \m{\beta^i} nodes, and
computation at each node has time  complexity proportional to \m{(n-i)}. 

The depth of the recursion tree is bounded by
\m{n}. However, it is also bounded by \m{min(n, V+1)}, where \m{V} is
the maximum number of variables in a \psiform{}.
As follows from Observation \ref{lm:varnum} (page \pageref{lm:varnum})
 unless \m{\psi\diff\psi_k = \set{\psi}}, each \psiform{} in
the difference \m{\psi\diff\psi_k} uses strictly fewer variables than
the original \m{\psi}, because when \m{\psi_k} nearly entails
\m{\psi}, all \psiforms{} in the e-difference 
\m{\psi\diff\psi_k} are  subsets of images from \m{\psi_k}'s exceptions, and unless  an exception entails the whole \m{\psi}, this image is obtained by instantiating some variables of \m{\psi}.
In the case where  \m{\psi\diff\psi_k = \set{\psi}},  
the branching factor out of the node equals one, and we can collapse the
parent  and the child  into one node. 
Thus, assuming \m{V} is less than \m{n}, the depth of recursion in
checking \m{\Psi\entails\psi} is bounded by the maximum number of
variables in a \psiform{},  \m{V}. The  overall time
complexity  bound is \m{\bigo(\beta^Vn)=\bigo(n)}, since \m{\beta} and \m{V}  are constants.


Figure \ref{fig:complexities} shows time complexity bounds of the
 \psiform{} calculus operations as functions of the number of participating \psiforms{} $n$, maximum number of exceptions $E$, maximum number of variables \m{V}, and maximum number of literals in a \psiform{} clause \m{C}.
The complete treatment of the time complexity
issues of  the calculus of \psiforms{}
can be found in \cite{babaian-thesis}. 

\begin{figure}[h!]

{\centering \begin{tabular}{ccccc}
\hline \\
&
\m{\set{\psi_1, \ldots, \psi_n}\entails\psi}&
\m{\image{\psi_1}{\psi_2}}&
\m{{\psi_2}\diff{\psi_1}}
\\\\
\hline\\
Simple fixed length \psiforms{}& 
\m{\bigo(n)}&
\m{\bigo(1)}&
\m{\bigo(1)}
\\
Non-simple fixed length \psiforms{}&
\m{\bigo(nE^{V(t+1)+1})}&
\m{\bigo(E^t)}&
\m{\bigo(E^{t+1})}
\\
Non-simple limited form \psiforms{} &
\m{\bigo(nE^{V+1})}&
\m{\bigo(E)}&
\m{\bigo(E^2)}
\\
Singleton \m{\psi_1}&
\m{\bigo(nE)}&
\m{\bigo(E)}&
\m{\bigo(E)}
\\\\
\hline
\end{tabular}\par}
\caption{Time complexity of computing \psiform{} operations. Assumes unification takes constant time and the 
maximum number of literals (\m{C}) and variables (\m{V}) in the main form of a
\psiform{} are constant. 
\m{E} denotes the maximum number of exceptions, \m{t}
 denotes   the maximum number of possible subset matches between main forms of
two \psiforms{}.\m{ t =\bigo(e^{C/e})},  where \m{e} is the Euler's
number, for fixed length \psiforms{}. }\label{fig:complexities}

\end{figure}

To summarize:
\psiform{} entailment takes
linear time in the number of participating \psiforms{} $n$, when the
maximum length of clause, maximum number of variables in a \psiform{}
and maximum number of exceptions are all fixed. When the number of
exceptions is proportional to \m{n}, computing entailment remains bounded by 
a polynomial of the order proportional to the
maximum number of variables in a \psiform{} times the number of
possible subset-matches between the main clauses.
The complexity  of \psiform{} operations depends  on 
the number of  subset-matches between the main clauses of two
\psiforms{}, 
\corr{2}{which in case of unrestricted \psiforms{}  is \m{C^C}.}
In fixed length \psiforms{} the number of  subset-matches between the
main forms of two \psiforms{}  is
bounded by \m{e^{C/e}}, as stated by the next Observation.
To limit the number of subset-matches to at most one, \psiforms{} can
be restricted to contain no duplicate occurrences of a predicate
symbol in the main clause.  We call such \psiforms{}  \mydefi {limited
form \psiforms{}}.

\begin{ourlemma}\label{lm:fixednooverlap}
  Let \m{\psi_1} and \m{\psi_2} be fixed length \psiforms{}. In two
  different subset-matches of \m{\main(\psi_1)} onto
  \m{\main(\psi_2)}, no two different literals of \m{\main(\psi_1)}
  match the same literal of \m{\main(\psi_2)}.
\end{ourlemma}

\proof
  As always, we assume that there is no overlap between the variables
  in two different \psiforms{}.  Suppose that the above statement is
  not true, i.e. there exist two different subset-matches of
  \m{\main(\psi_1)} onto \m{\main(\psi_2)} that match two different
  literals \m{d_1} and \m{d_2} of \m{\main(\psi_1)} on the same
  literal \m{d} of \m{\main(\psi_2)}.  Let \m{\sigma_1} and
  \m{\sigma_2} be substitutions corresponding to each subset-match,
  i.e.  \m{\sigma_1} and \m{\sigma_2} are in
  \m{\mgusubset(\main(\psi_1), \main(\psi_2), \vars(\psi_1))}.
  
  Consider the following substitution \m{\sigma}.
\begin{itemize}
\item If \m{d_1} and \m{d_2} do not share variables, construct 
\m{\sigma} as follows: combine bindings on variables in \m{d_1} from
\m{\sigma_1} and  bindings on variables in \m{d_2} from \m{\sigma_2}.
Then, obviously, \m{d_1\sigma = d_2\sigma = d}.
  
\item Otherwise, if \m{d_1} and \m{d_2} do share variables, rename
variables in \m{d_2} so that there is no overlap with variables in
 \m{d_1}.  Modify \m{\sigma_2} by renaming the variables in the same
 way we did with  \m{d_2}, and construct \m{\sigma} as in the previous
 case. Again, \m{d_1\sigma = d_2\sigma}.
\end{itemize}

Thus, both cases produced a contradiction to the fact that no two
literals of a fixed length \m{\psi_1} unify.\qed

\begin{ourobservation}\label{lm:fixednummatches}
\lemmaEight
\end{ourobservation}

\proof Suppose there are \m{i_k} literals in \m{\main(\psi_2)} that
matched the k-th literal of \m{\main(\psi_1)}.  Since according to
Lemma \ref{lm:fixednooverlap} (page \pageref{lm:fixednooverlap}) no
two different matches can match two different literals of
\m{\main(\psi_1)} to the same literal of \m{\main(\psi_2)},
\m{i_1+\ldots+i_k \leq C}, otherwise two literals of \m{\main(\psi_1)}
would match the same literal of \m{\main(\psi_2)}.

The maximum size of \m{{\mgusubset(\main(\psi_1), \main(\psi_2), \vars(\psi_1))}} is bound by the product \m{i_1\times\ldots\times i_k }. 
By the Cauchy's inequality \md{ \frac{i_1+\ldots+i_k}{k}\geq
(i_1\times\ldots\times i_k)^{1/k}.} 
  The product \m{(i_1\times\ldots\times i_k)} is limited by   
\m{(\frac{C}{k})^k} with the equality reachable only when
\m{i_1=i_2=\ldots=i_k = \frac{C}{k}}. Then the product \m{i_1\times\ldots\times i_k }.   The  maximum of this product is reached when  \m{\log(\frac{C}{k}) -1 = 0}, i.e. when  \m{\frac{C}{k} = e}, and equals \m{e^{C/e}}.
  
Thus, when the maximum number of disjuncts in a \psiform{} is \m{C}, number of possible subset-matches is bounded by \m{e^{C/e}}.\qed


\section{
\psiplan{} Representation}\label{sec:psiplan}

As in previous work on open world planning, we assume that the world evolves
as a sequence of states, where the transitions occur only as the result of
deliberate action taken by the single agent.  Since the agent's model of the
world is incomplete, the actual state of the  world differs from the state of the agent's knowledge of the world, which we call \mydefi{ SOK}.  The agent's 
knowledge of the world is assumed to be {\em correct}.

\subsection{ States of Knowledge.  \psiplan{} propositions. }

A SOK is a set of propositions that represents what the agent knows is true
about the world. In \psiplan{}, a {SOK}\label{def:SOK} is a finite set of \psiplan{} \mydefi{domain propositions} or, simply, \mydefi{propositions}, which are defined to be either ground atoms or \psiforms{}. 

Since the agent's theory of the world is assumed to be correct, every
proposition that a SOK entails is true in the actual world.  Moreover, we make the
following \emph{closed knowledge assumption} (CKA): \edit{B-2}{A literal  \m{L} is known to be true in \m{s} if \m{s\models{}L}, known to be false if \m{s\models{}\neg{}L} and unknown if both \m{s\not\models{}L} and \m{s\not\models{}\neg{}L}}. This assumption is {\em closed} because entailment is decidable in \psiplan{}.
 
A model of a \psiplan{} proposition is a world in which it is true.
We refer to the set of all models of a SOK \m{s}, denoted \m{\Ms(s)}, as the set of \mydefi{possible worlds} of \m{s}. It is the set of all worlds in which everything known by the agent is true. Correctness of the agents SOK implies that  the set of models of a SOK always contains the actual world. 

We  use symbols \m{w, w', w_1 \ldots w_n} to refer to  worlds, \m{W,W'} to refer to the sets of worlds, and \m{s, s', s_1
\ldots s_n } to refer to the agent's SOK. \m{\cal W} denotes a set of all worlds.

\ignore{
The set of world states represented by a SOK \m{s} is called a set of
\mydefi{possible worlds} of \m{s} and denoted \m{\Ms(s)}. Since we assume that
all of the agent's knowledge is correct, the set of
possible worlds is a set of all worlds  in which everything that the
agent knows is true:
\md{\Ms(s) = \set{w\ourst w(s)=\true}.}
In other words,  the set of \mydefi{possible worlds} of \m{s} equals the  set of all models of \m{s}.
}

\subsection{Entailment in \psiplan{}}


First observe that any consistent set \m{s} of atoms plus \psiforms{}
does not entail any atoms but those in \m{s}. 
However, it may entail more \psiforms{} than are entailed by
\psiforms{} of \m{s} alone, because of the possibility of resolution
between a ground clause \m{c}, represented by a \psiform{}, and some
atom \m{a}, such that \m{\neg a} is a literal of \m{c}. 
However, once all resolutions are performed and the resolvents added
to \m{s}, each \psiform{} entailed by  \m{s} is entailed by the set of  only 
\psiforms{} of  \m{s}, and each atom entailed by \m{s} is in \m{s}. 
In other words, \m{s} becomes {\em saturated}, i.e. 
for any ground proposition \m{q} entailed by \m{s} there is a single
proposition \m{p\in s} that entails \m{q} (see also Theorem
\ref{th:clause-to-clause}). 
A set of propositions \m{s} is \mydefi{saturated} if and only if for any ground proposition \m{q},
 \begin{equation}\label{def:saturated}
(s\entails q) \implies \ourexists{p}(p\in s) \land
(p \entails q).
\end{equation}
Thus, when \m{s} is saturated, one need not combine elements of \m{s}
(through resolution) in order to show entailment. To determine
entailment of an atom, \m{s\entails a}, \m{a} must be found in \m{s}.
Entailment of any ground negated clause, or, in other words, singleton  \m{\psi},  \m{s\entails \psi} is completely determined by entailment from a single \psiform{} in \m{s}.

A {\em saturated} equivalent of \m{  s} is obtained by computing
all possible resolutions from the unit clause resolution rule
\m{\infruletop{a,\neg 
    a \vee \neg a_1\vee\ldots \vee\neg a_n}{\neg a_1\vee\ldots
    \vee\neg a_n}} between domain atoms in \m{  s} and clauses
represented by \psiforms{}.

For example, suppose \md{{  s} = \set{In(paper, /tex), \psi =
    \apsi{\neg In(x,y) \vee \neg T(x,PS)}{\set{y=/img}}}.
  } Here, \m{In(x,y)} states that file \m{x} is in directory \m{y}, and 
\m{T(x,PS)} states that file \m{x} has type Postscript.

\m{  s} is not saturated because, even though it entails that file
\m{paper} is not a Postscript file, no single proposition of \m{ 
  s} alone entails \m{\neg T(paper, PS)}.

However, \m{\psi} in \m{  s} contains a clause \m{c= \neg In(paper,
  /tex) \vee \neg T(paper, PS)}, and we can perform a resolution
between $c$ and the atom \m{In (paper, /tex)}, resulting in \m{\neg
  T(paper, PS)}. We add the \psiform{} \m{\spsi{\neg T(paper,
    PS)}} to the initial SOK, \m{s_0}.

\corr{2}{
The procedure \m{ Saturate({s})}, depicted in Figure
\ref{fig:saturate}, returns a saturated equivalent of \m{s} and
consists of the following steps.  Initially we set \m{s_0 = {  s}}.
For every \psiform{} \m{\psi} in \m{s_0} and for every atom \m{a}, we
compute \m{\image {(\neg{a})}{\psi}}, as those are all and only
clauses for which resolution is possible. If this image is empty, we
go to the next \psiform{} in \m{s_0}. Otherwise, suppose \m{\image
  {(\neg{a})}{\psi} = \psi'}.  From the properties of the image, it
follows that \m{\neg{a}} is a subclause of \m{\main(\psi')}. Let
\m{\psi_{new}} denote the \psiform{} that is obtained from \m{\psi'}
by removing \m{\neg{a}} from its main  clause, i.e. \m{\psi_{new} =
  [\main(\psi') - {(\neg{a})} \mbox{ except } \subst(\psi')]}.
We add \m{\psi_{new}} to \m{s_0}, and continue until all \psiforms{} in \m{s_0}, including the newly added, are processed in this way.
}

\begin{figure}[h!] {
\small
\begin{tabular}{rp{.25in}p{.25in}p{.25in}p{.25in}p{.25in}}
\hline \\
\multicolumn{2}{l}{{\bf Saturate($s$)}}&\\
1.& \multicolumn{1}{l}{Set \m{s_0 =  s}}\\
2.& \multicolumn{3}{l}{For each \psiform{} \m{\psi\in s_0}}\\ 
3.&& \multicolumn{3}{l}{For each atom \m{a \in s_0}}\\
4.&&&  \multicolumn{2}{l}{If \m{\image {[\neg{a}]}{\psi} \neq \emptyset} Then}\\
5.&&&&  \multicolumn{1}{l}{Set \m{\psi' =\image {[\neg{a}]}{\psi}}}\\
6.&&&&  \multicolumn{2}{l}{Let \m{D} denote the main clause of \m{\psi'} without the literal \m{\neg{a}}} \\ 
7.&&&&\multicolumn{2}{l}{ If \m{D=\emptyset}, Then  return {\bf fail}.}\\
8.&&&& \multicolumn{2}{l}{Else Set \m{\psi_{new} = [D \mbox{ except } \Sigma(\psi)]},  Add \m{\psi_{new}} to \m{s_0}. }\\
9. && \multicolumn{1}{l}{End For}\\
10. & \multicolumn{1}{l}{End For}\\
11.& \multicolumn{2}{l}{Return \m{s_0}.}\vspace{2mm}\\
\hline
\end{tabular}
}
\caption{ Procedure Saturate(\m{s}). Preprocessing the Initial SOK. 
  If set \m{s} is unsatisfiable, returns {\bf fail}. }\label{fig:saturate}
\end{figure}

Notice that, as a side effect, procedure \m{Saturate()} determines if
\m{  s} is consistent. If at any moment we obtain an empty clause
as a main part of some \m{\psi_{new}}, that indicates that
\m{\main(\psi') - ({\neg{a}}) = \emptyset}, i.e. \m{\main(\psi') =
  {\neg{a}}}, which means that we can derive both \m{a} and \m{\neg a}
from \m{  s}, and hence \m{  s} is inconsistent.

\begin{ourlemma}\label{lm:saturate}
\lemmaSeven
\end{ourlemma}

\proof The SOK returned by {\bf Saturate} (Figure \ref{fig:saturate})
  contains the input set of propositions \m{s} and some additional
  propositions that are derived from \m{ s} using unit clause
  resolution, i.e. those that follow from \m{s}. Thus the returned SOK
  \m{s_0} is equivalent to the input.
  
  It is saturated because we compute and add the results of all
  possible resolutions to \m{s_0}. Thus, for every ground proposition \m{p}
  such that \m{s\entails p}, there is a  proposition
  \m{c\in s_0} such that \m{c\entails p}.
  
  {\bf Saturate} computes all possible resolutions in \m{s} and
  returns {\bf fail} whenever an empty disjunct is derived, as follows
  from the the known property of resolution deduction (see
  \cite{genesereth-nilsson} page 87): If a set \m{\Delta} of ground
  clauses is unsatisfiable, then there is a resolution deduction of
  the empty clause from \m{\Delta}.

To estimate the time complexity bound, we consider the following
stages of the algorithm.  During the first stage, for each of n
\psiforms{} in \m{s} the procedure will first compute the resolution
with every atom in \m{s}, when the resolution rule is
applicable. Determining if resolution between an atom and a single
fixed length \psiform{} is applicable consists of computing an image
of a single literal onto a \psiform{} (step 4 in Figure
\ref{fig:saturate}), which takes constant time, assuming the number of
exceptions, clauses and variables in a \psiform{} are constant
bounded.  Thus, the first stage takes time \m{O(mn)}.

Note that the result of each resolution is another \psiform{} (denoted
\m{\psi_{new}} in the algorithm), which is added to the saturation
\m{s_0}.  For each \psiform{} at most \m{m} new \psiforms{} can be
added to \m{s_0} as a result of resolution with the atoms in
\m{s}. Furthermore, each of the added \psiforms{} will have 1 fewer
literals in the main clause than the original. Overall, \m{nm} new
\psiforms{} with the maximum clause length \m{C-1} could be added to
\m{s_0} during the first stage.

During the second stage, resolutions are computed between the
\psiforms{} added in the previous stage and \m{m} atoms of \m{s}. This
process will take time \m{O(m^2n)} and add no more than \m{m^2n} new
\psiforms{} with the maximum clause length of \m{C-2}.

The third stage will take \m{O(m^3n)} time and add no more than
\m{m^3n} new \psiforms{} with the clause length \m{C-3}, and so on.
Since the maximum possible length of the main clause in the
\psiforms{} added at each stage is decreasing by one at each stage of
this process, it is evident that the number of stages is bounded by
the size of the longest \psiform{} clause \m{C}. Overall computation
will thus take time \m{\bigo(mn+m^2n+\ldots+m^Cn) = \bigo(m^Cn)}.\qed

When  a set of \psiplan{} propositions \m{s} consists
of \m{m} atoms and \m{n} \psiforms{},  
checking \m{s\entails a}, where \m{a} is an atom, takes time 
 \m{\bigo(m)} because a set of \psiplan{} propositions only entails those atoms that it contains. Checking \m{s\entails \psi} takes time
\m{\bigo(nm^C + n)}, where  \m{\bigo(nm^C)} is the time to  saturate \m{s}.
When \m{s} is saturated,  checking \m{s\entails\psi} is \m{\bigo(n)}.

\subsection {\psiplan{} Actions.}
\corr{1.6,1.8}{
Actions are deterministic and are represented 
via preconditions and effects. Actions are described using parameterized schemas, however, for the simplicity of presentation, the examples in this section present instantiated, i.e. fully grounded, versions of actions.  }

Each action \m{a} has a {\em name\/}, \m{\aname(a)}, and  a set of {\em
  preconditions\/}, \m{\precond(a)}, which identify the domain
  propositions necessary for executing the action. The propositions in  
\m{\precond(a)} can include literals and quantified \psiforms{}
\footnote{Ruling out other forms of non-quantified disjunction is not
  a limitation, since any action schema with a non-quantified
  disjunction as its precondition can be equivalently split into
  several actions.}.

Each domain action has a set of domain literals called the {\em assert
  list\/}, \m{\assertlist(a)}.  The assert list, also called the {\em
  effects\/} of the domain action, identifies the complete set of
domain propositions whose value may change as a result of the action.
We assume that an action is deterministic and can change the truth value 
({\em true} or {\em false}) of
only a  finite number of atoms, and thus any \psiform{} in the assert
list defines a single negated literal.

Consider, for example, \psiplan{} encoding of the action of moving a file
   from one directory to another, as  depicted in Figure \ref{fig:actions}.
   $mv(F,S,D)$ moves file $F$ from directory $S$ into the directory
  $D$. The single precondition requires that file $F$ be in directory
  $S$. The effects are given by a \psiform{} that denotes that file
  $F$ is not in directory $S$, and an atom that denotes that file $F$
  is in directory $D$. 
 Action  $lift(B, L)$, from our warehouse domain, 
lifts object \m{B} from location \m{L}.
One of its preconditions is a quantified \psiform{} and   requires
that \m{B} contains no fragile goods.

\begin{figure}[tbh!]
{\small
\begin{tabular}{lllllll}
\hline\\
\multicolumn{3}{l}{\psiplan{}  action \m{a=mv(F,S,D)}} &
\multicolumn{3}{l}{\psiplan{}  action \m{a=lift(B,L)}}
\vspace{0.1in}
\\

&\multicolumn{2}{l}{\m{\precond(a): In(F,S), File(F), Dir(S), Dir(D)}}
&&\multicolumn{2}{l}{\m{\precond(a): \spsi{\neg In(g, B) \vee \neg Fragile(g)}}, \m{ At(B,L)}} \\
&\multicolumn{2}{l}{\m{\assertlist(a): \psiset{\neg In(F, S)}, In(F,D)}}
&
&\multicolumn{2}{l}{\m{\assertlist(a): \spsi{\neg At(B,L) }, Lifted(B)}} \\\\
\hline \\
\end{tabular}
}
\caption{\psiplan{} domain actions}\label{fig:actions}
 \end{figure}

\subsection{Planning Problem}
  A \mydefi{planning problem} is a three tuple \m{\langle\Lambda, {\cal I},
  {\cal G}\rangle} where \m{\Lambda} is the set of available \psiplan{}
  actions, \m{\cal I} is the set of initial conditions -- i.e., a set
  of \psiplan{} propositions -- and \m{{\cal G}} is the goal, which is
  also a set of \psiplan{} propositions.

A {\em solution plan} is a  sequence of actions, that is executable and
transforms any world state satisfying the initial conditions into a world
state satisfying the goal.

Given a sequence of ground actions \m{a_1, \ldots, a_n}, 
let \m{W_i} denote the set of possible worlds obtained by executing the sequence up to the i-th action from any of the possible initial worlds. 
Let \m{W_0} denote the set of possible worlds corresponding to the initial
conditions, i.e. \m{\Ms({\cal I})}.
Then, a sequence of actions \m{a_1, \ldots, a_n} is called a
\mydefi{solution plan} to a planning problem  \m{\langle\Lambda, {\cal I},
{\cal G}\rangle}, if and only if:
\begin{enumerate}
\smitem \label{cond1} The goal \m{\cal G} holds in all final worlds, i.e. for all \m{w} in \m{W_n}, \m{w({\cal G})}, and
\smitem \label{cond2} Each action \m{a_i} of the plan is executable in every
possible world \m{w} in \m{W_i}, for all values of \m{i}, \m{0\leq i < n}, i.e. for all \m{w} in \m{W_i}, \m{w (\precond(a_i))}.
\end{enumerate}

Since our agent uses the SOK \m{s} to represent the {\em set} of possible
worlds \m{\Ms(s)}, in order to plan, it must be able to
{\em progress} the SOK in order to predict the {\em set} of worlds
resulting from executing a sequence of actions. The function
\m{update()} does exactly that.

If \m{\vec{a}} is a sequence of actions executable
from the SOK \m{s_0}, \m{update(\vec{a},s_0)} denotes the SOK our
agent uses to predict the set of possible worlds resulting from
executing \m{\vec{a}} in any of the worlds \m{\Ms(s_0)}.

Ideally, the SOK obtained by progression must include {\em all and
  only } worlds that are the result of executing the sequence
\m{\vec{a}} in some world described by the initial SOK. Then, every plan
that is executable and achieves the goal in the {\em agent's knowledge} of the
world is indeed a solution plan for the real world.
This requirement is satisfied when the \m{update()} function is {\em
  correct and complete}. 

The next section formally defines the correctness and completeness
properties, presents \psiplan's update procedure and proves that it is
correct and complete. Thus, a sequence of ground actions \m{a_1,
\ldots, a_n}, is a solution to the planning problem \m{\langle\Lambda,
{\cal I}, {\cal G}\rangle} if and only if

\begin{enumerate}
\smitem  The goal \m{\cal G} is entailed by the final SOK, i.e. \m{update(a_1 \ldots a_n, {\cal I})\entails {\cal G}}, and
 \smitem \label{cond3} Each  \m{a_i} is executable, i.e. 
\m{update(a_1\ldots a_{i-1}, {\cal I})\entails \precond(a_i)}.
\end{enumerate}
Thus, goal achievement can be established by checking entailment from the updated SOK without considering the set of all possible worlds.

\subsection{SOK Update}
\ignore{
To this end we informally introduce the problem of planning. Formal
definition is given later. 
In a planning problem we are given a set of initial conditions
\m{\cal I}, a set of goals \m{\cal G} and a set of available actions.
\m{\cal I} and \m{\cal G} are sets of propositions.
A {\em solution plan} is a  sequence of actions, that is executable and
transforms any world state satisfying the initial conditions into a world
state satisfying the goal. }

Actions cause transitions between worlds.
The agent's SOK must evolve in parallel with the world, and must
adequately reflect the changes in the world that occur due to an action.
{\em Correctness} of a SOK update guarantees that the SOK is always
consistent with 
the world model, given a consistent initial SOK. The other desirable property 
of the SOK update is {\em completeness}: we would like the agent to take
advantage of all information that becomes available and not to discard what
was previously known and has not changed.
The correctness and completeness properties of the SOK update, as
well as soundness and completeness of entailment within the state language, are
prerequisites for a sound and complete planning algorithm.   
\ignore{
Since our agent uses the SOK \m{s} to represent the {\em set} of possible
worlds \m{\Ms(s)}, in order to plan, it must be able to
{\em progress} the SOK in order to predict the {\em set} of worlds
resulting from executing a sequence of actions. The function
\m{update()} does exactly that.

If \m{\vec{a}} is a sequence of actions executable
from the SOK \m{s_0}, \m{update(\vec{a},s_0)} denotes the SOK our
agent uses to predict the set of possible worlds resulting from
executing \m{\vec{a}} in any of the worlds \m{\Ms(s_0)}.

Ideally, the SOK obtained by progression must include {\em all and
  only } worlds that are the result of executing the sequence
\m{\vec{a}} in some world described by the initial SOK. Then, every plan
that is executable and achieves the goal in the {\em agent's knowledge} of the
world is indeed a solution plan for the real world.
This requirement is satisfied when the \m{update()} function is {\em
  correct and complete}.
}
The correctness and completeness criteria are best formulated in the context of possible worlds.
Let \m{do(a, W)} denote the set of worlds obtained from performing action \m{a}  
in any of the worlds in \m{W,} and  \m{update(s,a)} denote the SOK 
that results if the agent performs  action \m{a} from SOK \m{s}.
We say that the update procedure is \mydefi{correct} if and only if  
\begin{equation}
\label{eq:correct}
 \Ms(update(s,a))\subseteq do(a, \Ms(s)),
\end{equation}
i.e. every possible world after performing the action \m{a} has to have a
possible predecessor. 

The update procedure is \mydefi{complete} if and only if   
\begin{equation}\label{eq:complete}
 do(a, \Ms(s)) \subseteq  \Ms(update(s,a)),
\end{equation}
i.e. every world obtained from a previously possible world is a model of
the new SOK. This implies that all changes to the world must be reflected in
the new SOK.

To achieve correctness of SOK updates, the agent
must remove from the SOK all propositions whose truth value might have 
changed as the result of the performed action. 
	To achieve completeness, the
agent must  add to the SOK  all facts that become known.
The complexity of the SOK update, therefore, depends critically on the process of 
identifying the propositions that must be retracted to preserve
correctness. In our language, this computation is reduced to computing 
e-difference, which has polynomial complexity.

To obtain the agent's SOK \m{s} after performing an action \m{a}, we
first remove all propositions implied by the negation of the assert
list, as only those propositions of \m{s} might change their values
after \m{a}. \edit{A-7}{Symbol \m{\assertlist^-(a)} is used to denote the set of negations of propositions in the assert list of \m{a}. For example, 
 the action \(a=mv(fig, /img,  /tex)\) of moving file \m{fig} from
directory \(/img\) into \(/tex\) has  assert list 
\md{\assertlist(a)=\set{\psiset{\neg{}In(fig,/img)}, In(fig, /tex)},}
and thus
\md{\assertlist^-(a)=\set{In(fig,/img),\psiset{ \neg In(fig, /tex)}}.}}

During the update we also remove from the SOK all redundant propositions,
i.e. those that follow from the effects of the action, and then add
these effects to the new state.  The agent's SOK after executing
action \m{a} in the SOK \m{s} is described below. 

\begin{equation} \label{eq:update}
 update(s, a) = ((s \diff {\assertlist}^-(a) )\diff {\assertlist}(a))
 \cup \assertlist(a) 
\end{equation}

Though it is not necessary, we include \m{\diff {\assertlist}(a)} in (\ref{eq:update}) to keep the SOK simple. \m{\diff {\assertlist}(a)} removes {\em all} propositions entailed by \m{{\assertlist}(a)}, not just those in \m{{\assertlist}(a)}.

\begin{ourexample}
\edit{A-7,8}{  For an example, consider action \(a=mv(fig, /img,
  /tex)\) introduced above.
  Let \m{\precond(a)=\set{In(fig,/img)}}, which states that \(fig\)
  must be in \(/img\).  Recall that 
  \m{\assertlist(a)=\set{\psiset{\neg{}In(fig,/img)}, In(fig, /tex)}}
and \m{\assertlist^-(a)=\set{In(fig,/img),\psiset{ \neg In(fig,
/tex)}}}.}

 We  begin with an SOK \m{s}, which states that \m{fig} and \m{a.bmp} are
  the only files in \m{/img}, and that \m{a.ps} is the only Postscript
  file in the system, except for possibly files in directory \m{/img}.
  \md{ s=\left\{\begin{array}{l}
        In(fig,/img), In(a.bmp,/img),T(a.ps, PS),\\
        \apsi{\neg In(x,/img)}{\set{x=fig},\set{x=a.bmp}},\\
        \apsi{\neg{}In(x,d) \vee\neg{}T(x,PS)}{\set{x=a.ps},\set{d = /img}}
	\end{array}\right\}}

\noindent\(a=mv(fig, /img, /tex)\) is executable in s, and as a result
of computing \m{s\diff{}\assertlist^-(a)}, the atom \m{In(fig,/img)} is
removed from \m{s} and the exception \m{\set{x = fig,d = /tex}} is added to the second \psiform{}.    
The first \psiform{} is left intact, producing
\ignore{
\md{ s \diff \assertlist^-(a) =\left\{\begin{array}{l}
     In(a.bmp,/img),  T(a.ps, PS),\\
      \apsi{\neg{}In(x,/img)}{\set{x=fig},\set{x=a.bmp}},\\
      \apsi{\neg{}In(x,d) \lor \neg{}T(x,PS)}{\set{x = a.ps},\set{d =
/img},\set{x = fig,d = /tex}}
        \end{array}\right\}
      .} 
}
{
\md{ s \diff \assertlist^-(a) =\left\{\begin{array}{l}
     In(a.bmp,/img),  T(a.ps, PS),\\
      \apsi{\neg{}In(x,/img)}{\set{x=fig},\set{x=a.bmp}},\\
     {[\neg{}In(x,d) \lor \neg{}T(x,PS) \mbox { except} } \\
 	 \mbox{ } {\set{\set{x = a.ps},\set{d = /img},\set{x = fig,d = /tex}} ]}
        \end{array}\right\}
      .} 
}

\edit{A-8}{
Further e-difference with \m{\assertlist(a)} and union with
\m{\assertlist(a)} yields the following SOK
\md{ s'=\left\{\begin{array}{l} In(fig, /tex), In(a.bmp,/img),
      T(a.ps, PS),\\
      \apsi{\neg{}In(x,/img)}{\set{x=fig},\set{x=a.bmp}}, \\
      \apsi{\neg{}In(x,d) \lor \neg{}T(x,PS)}{\set{x = a.ps},\set{d =
/img},\set{x = fig,d = /tex}}\\
      \spsi{\neg{}In(fig,/img)}
        \end{array}\right\}
      .}
}
 Note that \(s\) contained \(\neg{}In(fig,
    /tex)\lor\neg{}T(fig, PS)\) and that we added \(In(fig, /tex)\)
    when determining \(s'\).  If our update rule retained
    \(\neg{}In(fig, /tex)\lor\neg{}T(fig, PS)\) in \(s'\), then in
    \(s'\) we could perform resolution and conclude that
    \(\neg{}T(fig, PS)\).  However, this would be wrong because we
    have no information on whether or not \(fig\) is a Postscript file.
    Instead, our update rule removes any clause that is entailed by
    \(\neg{}In(fig, /tex)\), and so \(s'\) does not contain
    \(\neg{}In(fig, /tex)\lor\neg{}T(fig, PS)\).

\end{ourexample}

This  update rule (\ref{eq:update}) produces the same result as
 Winslett's update  operator \cite{winslett-88} in the special case
 where actions are  deterministic.  Moreover, our rule accomplishes
 this without considering all  possible worlds corresponding to SOK
 \m{s} explicitly, and thus is more efficient.

\edit{}
{We show next how the same rule is used for updating the state of knowledge after the actions that create new objects\footnote{Note that {\em discovering} a new object is a different issue (since it assumes that the object always existed in the domain and is therefore included in the universal quantification even prior to being discovered, which is not true of a newly created object) that is also handled in \psiplan{}, however it is beyond the scope of this paper.}.}

\begin{ourexample}
Consider an action of creating a new file named \m{\textit{afile}} in directory \m{\textit{/code}} with effect \m{In(\textit{afile,/code})} and no preconditions. When this action is executed in \m{s'}, the state update rule yields the new SOK \m{s''} that differs from \m{s'}
in the following way:
\begin{enumerate}
\item \m{s''} contains a new atom \m{In(\textit{afile},/code)}, reflecting the effect of the action added by the update rule, 				
\item the first \psiform{} of \m{s'} now has a new exception, reflecting the fact that it is unknown whether \m{afile} has Postscript format. This is the result of the e-difference \m{s'\diff\set{\spsi{\neg{}In(\textit{afile},/code)} }}.
\end{enumerate}
 \md{ s''=\left\{\begin{array}{l}
 In(\textit{afile},/code), In(fig, /tex), In(a.bmp,/img),
      T(a.ps, PS),\\
\mbox{[}{{\neg{}In(x,d) \lor \neg{}T(x,PS)}}\mbox{ except }\{\set{x = a.ps},\set{d = /img},\set{x = fig,d = /tex}, \\
 \set{x = \textit{afile}, d = /code}\}\mbox{]}\\
\apsi{\neg{}In(x,/img)}{\set{x=fig},\set{x=a.bmp}}, \\
\spsi{\neg{}In(fig,/img)}
\end{array}\right \}
      }
\end{ourexample}

A factor that turns out to be  critical for the use of \psiplan{} in planning
is that the SOK resulting 
from updating a saturated SOK is also saturated. After the initial
 SOK of the agent is saturated, there is no need to
consider resolution of the initial conditions and action effects in satisfying a goal. 

We call an SOK \m{s} \mydefi{minimal} if it is saturated and it does not
contain any ground clause entailed by some other clause in \m{s}, i.e.
for any two ground clauses \m{p}, \m{q} from \m{s}
\begin{equation}\label{def:minimal}
(q\entails p) \implies (q=p)
\end{equation}

\begin{ourtheorem}\label{th:update}
\thrmEleven\qed
\end{ourtheorem}

\proof
We start by proving that the result of updating a saturated SOK is a
saturated SOK.   
We first show that if \m{s} is saturated  then \m{s_1= ((s \diff
{\assertlist^-}(a)) \diff \assertlist(a))} is also saturated (
recall that \m{\mathbf{{\assertlist}^-(a)}} is used to denote the set of negations of propositions in the assert list of \m{a}).
Suppose this is not true, i.e. there's a ground \psiplan{} proposition \m{p} such that, while
\m{s_1\entails p },  \m{\ourforall{p'\in s_1}p' \not\entails{}p
}. Let \m{q} be a smallest (non-empty) subclause of \m{p} such that \m{s_1\entails
q} (there might be several such \m{q}). Since \m{s_1\subseteq s},
 \m{q}  must also be entailed by \m{s}, but \m{s} is saturated, so
 there exists a \m{q'} in \m{s} such that  \m{q'\entails q}, i.e.  
\m{q'\subseteq q}.

 \m{q'\not\in s_1} (otherwise \m{q'} as a subclause of \m{p} would
entail \m{p}) so \m{q'} must be entailed by either  \m{\assertlist(a)}
or  \m{\assertlist^-(a)}.  We abbreviate the union
\m{\assertlist(a)\cup{}{\assertlist^-(a)}} by
\m{\closure\assertlist(a)}.    
Since \m{q'\in ({\image {\assertlist(a)}{s}}) \cup( {\image{
\assertlist^-(a)} {s}})}, there must be a literal \m{e} in
\m{\closure\assertlist(a)} such that \m{e} is a subclause of  \m{q'}, and
therefore, \m{e} is a subclause of  \m{q}.
Note, that 
 since \m{s_1\entails q}, in case \m{q\not\in s_1} there must be a way of
deriving \m{q} by resolution from some propositions of \m{s_1}, which is only possible when there exists a proposition \m{r} in \m{s_1} such that \m{r}
 contains \m{q} as a subclause. \ignore{(To see this note that if \m{q\in{}s_1} then \m{r=q}. Otherwise, since for any subclause \m{q'} of \m{q}, \m{s_1\not\entails{}q'}, {q} is derived from \m{s} by resolution. Thus, there is a clause \m{r\in{}s} such that \m{q\subseteq{}r}.)} This means that \m{r} has \m{e} as a
subclause, and consequently \m{r} is entailed by \m{e}. But
according to  the definition of \m{s_1}, it does not contain any
clauses entailed 
by any clause in \m{\closure\assertlist(a)}. We arrive at a
contradiction, so \m{s_1} is saturated. 
  
Adding \m{\assertlist(a)}, which consists of literals and  is itself saturated, to \m{s_1} also produces a saturated state.
 Assume there is a proposition \m{q}
that is not entailed by either \m{s_1} or some element of
\m{\assertlist(a)}, but is entailed by
\m{s_1\cup{\assertlist(a)}}. \m{q} cannot be an atom, therefore it is
 a negated clause. The only clauses that are not entailed by
 \m{\assertlist(a)} and  \m{s_1} separately, but are entailed by their
 union are those obtained via resolution of some atom \m{e\in\assertlist(a)}
with a ground clause of the form \m{\neg{}e\vee x} in  \m{s_1}. But
 this is impossible, because \m{ \assertlist^-(a)\entails\neg{}e\vee
 x}, so such  clause would not be in \m{s_1}. We arrive at a contradiction. 

In case \m{s} was minimal, \m{s'} is also minimal, because
\begin{itemize}
\item removing clauses from a minimal set preserves minimality, so \m{s_1}
is minimal, and 
\item \m{s' = s_1\cup{\assertlist(a)} } is minimal because
\begin{enumerate}
\item \m{s_1} does 
not contain any propositions entailed by any literal in \m{\closure\assertlist(a)}, 
\item \m{\assertlist(a)} is minimal, and 
\item \m{s_1} does not entail anything in
\m{\assertlist(a)}.
\end{enumerate}
\end{itemize}

To prove that the \m{update} function is correct and complete, we need
to show that  for \m{s'= update(s,a)}: 
\begin{equation}
\Ms(s') = do (a, \Ms(s)).
\end{equation}

\ignore{\em For the purpose of this proof, instead of representing a
world with a set of atoms and making the closed world assumption, 
we will assume  that each world \m{w} is represented by a complete set of
literals that are true in the world. These two representations are
equivalent.}

{\em Completeness proof.} We first show that \m{ do (a,
\Ms(s)) \subseteq \Ms(s') }, i.e. for every world \m{w} in
\m{\Ms(s)}, its successor,  \m{w' = do (a,w)}, is  in \m{\Ms(s')}.    
The set of possible worlds consists of all and only worlds that
model the agent's knowledge of the world, i.e. for any \m{s} \m{\Ms(s) = \set{w\ourst
(p \in s) \implies w(p) = true}.}
Thus, to show that \m{w'\in\Ms(s')} we need to prove, that for every
proposition \m{p} in \m{s'},  \m{w'(p)=\true}.

Note that according to the world transition model, \m{w'= (w -
\closure{\assertlist}(a)) \cup {\assertlist(a)}}.

We partition \m{s'} into \m{s'_1 =((s \diff{\assertlist^-}(a)) \diff
 \assertlist(a)) } and \m{s'_2 = \assertlist(a)}. We partition
\m{w'} into \m{w'_1 = w - {\assertlist^-}(a)- {\assertlist}(a)} and
\m{w'_2= {\assertlist}(a)}.
Since \m{w\in\Ms(s)}, for every \m{p_1} such that \m{p_1\in s'_1},
 \m{w'_1(p_1)}. Also, for each \m{ p_2 \in s'_2}, we have
 \m{w'_2(p_2)}. Therefore for every \m{p} such that \m{p\in s},
\m{w(p)}. 
 
{\em Correctness proof.}  Now we need to show that 
 \m{\Ms(s')\subseteq do(a, \Ms(s))}, i.e. every possible world  \m{w'}
of \m{s'} has a predecessor \m{w} that is a possible world of \m{s}.
We need to show that for every \m{w' \in \Ms(s')} there is a world 
\m{w} such that \m{w' = (w - \closure{\assertlist}(a)) \cup
\assertlist(a) } where \m{w} is in \m{\Ms(s)} and 
\m{\closure\assertlist(a)} denotes the union \m{\assertlist(a)\cup{}{\assertlist^-(a)}}. A possible world
is a model of all propositions  \m{p} such that \m{p\in s},
i.e. \m{w} is in \m{\Ms(s)} if and only if \m{w} models every such
domain proposition \m{p}.

The proof is by construction. Since \m{s} is saturated,  for every
proposition \m{p} that is implied by \m{s}, there's a single
proposition \m{q\in s} such that \m{q\entails p}.

STEP 1. Since \m{w' = do (a,w)}
we need to include in \m{w} all literals of \m{w'} that are not in
\m{{\assertlist (a)}}, because those would not have changed as a
result of the action. Let \m{w_0 = w' - {\assertlist}(a)} and \m{w} will include \m{w_0}.

STEP 2. We also include in  \m{w} those literals from
\m{\closure{\assertlist (a)}} that are {\em known} in \m{s},
i.e. the literals \m{l\in \closure{\assertlist} (a)} such that \m{l
\in s}.  

STEP 3. At this point every literal or its complement is included in
\m{w} 
except for \m{l\in \closure{\assertlist}(a)} where neither \m{l} 
nor \m{\neg l} belongs to \m{s}. We now describe  a procedure for choosing 
either the literal or its complement for inclusion in \m{w} from these
``leftover'' literals. Suppose  \m{\neg l} is an arbitrary negated
literal from this set. Further, let \m{C = {\image {\neg l}
s}}. If there is 
a proposition in \m{C} that is not already implied by some \m{p},
where \m{p\in w}, then we must
include \m{\neg l} in \m{w} in order to keep it accessible from \m{s}.
Otherwise we may include in \m{w} either \m{l} or \m{\neg l}.  

The world \m{w} is now completely specified and is easy to verify that 
\m{w\in\Ms(s)}, as well as \m{w' = do(a,w)}.\qed

We should note that although the state update procedure above is
defined for fully grounded actions, it does not mean that all
\psiplan{}-based planners must work with fully grounded
representations. For example, the partial order planner \psipop{}
operates using action schemas and grounds action parameters only as
needed.

\section {Related Work}\label{sec:related}

\subsection{Representations for conformant planning}
Representations used for reasoning and planning in an open world  can be broadly categorized as those that operate using the set of all possible worlds, and those that rely instead on reasoning using a specification of incomplete  state of knowledge. Presented here \psiplan{} belongs to the second of these categories.

\ignore{
Among planners that operate in presence of incompleteness in the 
knowledge of the initial state of the world are those that perform
{\em contingency} (or {\em  conditional}) planning
(\cite{peot-smith-92}, \cite{pryor-collins-96}, 
\cite{rintanen-99}), i.e. build branching
plans for all possible values of unknown propositions, planners that
 find a solution for any state consistent with 
the initial information, also called {\em conformant} planning, 
(\cite{cimatti-roveri-99},
\cite{reiter-00},\cite{eiter-etal-00}, \cite{smith-weld-98}, \cite{eiter-etal-00}) or use sensing actions interleaved with execution of incomplete plans
(\cite{ambros-ingerson-steel-88}, \cite{etzioni-etal-92},
\cite{golden-etal-94}, \cite{golden-thesis},  \cite{babaian-thesis}) 
The planner \psipop{} presented here is a conformant algorithm,
and we have presented a brief comparison with other systems in the
introduction. 
}


Among the planning systems in the second category is a situation-calculus based planner by 
Finzi et al. \cite{reiter-00}, implemented in GOLOG \cite{golog}.
\ignore{
that builds plans  by 
regressing \cite{pirri-reiter,waldinger-77}  the formula that captures achievement of the goal   
after a sequence of actions executed from the initial situation 
to a logically equivalent formula about the initial situation only, and then 
checking whether the regressed formula can be proven from the specification 
of the initial situation. Along with regression this planner
eliminates futile partial plans by checking them against the domain specific 
description of a ``bad situation'' in the style of Bacchus and Kabanza
\cite{bacchus-kabanza-00}. 
}
The planning task is  reduced to theorem
proving in situtation calculus, and the authors present two approaches to theorem proving
from the initial state. One approach invokes a Davis-Putnam based theorem
prover every time entailment from initial situation is checked.
The other approach intends to minimize the time spent checking
entailment by precompiling the
specification of the initial state into its equivalent form containing
all prime implicates of the original specification. (The reduction to prime
implicates is akin to \psiplan's saturation of the initial state.)
From the prime-implicate form further theorem proving is
done by subsumption of clauses.  \ignore{However, in the worst case
the number of prime implicates generated for a clause is exponential
in the number of distinct atoms in those  clauses. } 

The foregoing is the only conformant planner that 
subsumes the state and goal language of \psiplan{}. However, the examples presented in the paper do not contain  universally quantified disjunctive goals with exceptions that are handled by \psiplan{}.   
The generality of the situation-calculus permits any
first-order specification of the initial and goal
situations, and actions, however, at a price of the complexity  of
planning. In \psiplan{}, we have deliberately and significantly
restricted the language for the sake of  reduced complexity of reasoning. 

\corr{2}{
A  subset of situation calculus with equality, which has tractable, sound
and under certain conditions complete action progression procedure from an incompletely specified state is presented by Liu and Levesque in
\cite{liu-levesque-05}. There are similarities 
between \psiplan{} and the language of Liu and Levesque, in particular
in the use of universally quantified statements in the knowledge
base. However, neither language subsumes the other one in the
expressive power. 
}

\corr{2}{
Shirazi and Amir (\cite{shirazi-amir-05}) also address the problem of progressing a belief state encoded in first-order logical sentences over a sequence of actions. They present special purpose algorithms for computing the progression, which they call {\em logical filtering}, in polynomial time. The polynomial time complexity of belief update is achieved for STRIPS and also {\em permuting} actions. An action is called permuting, if for each world  \m{w'} there is at most one \m{w} such that \m{do(a,w)=w'}, i.e. for every world potentially resulting from execution of action \m{a}, there is a unique "original" world state. \psiplan's actions are not permuting, however they are similar to STRIPS actions in the sense that the assert list of an action includes only those literals that change as the result of an action and there are no conditional effects. Thus, the polynomial time complexity of \psiplan's update procedure is consistent with the findings of Shirazi and Amir.
}

\ignore{
Levesque in \cite{levesque-96} points to the weaknesses of 
early works on open-world planning in specifying what the solution
,plan is in presence of incomplete information, sensing and conditional
execution, and 
introduces a robot program language for encoding plans with sensing
actions, conditional execution and iteration. He presents a
specification of a solution plan in an algorithm-independent manner
using the logical account of actions and the solution to the frame
problem in the presence of sensing actions in situation-calculus
presented in  \cite{scherl-levesque-93}. 
}

\ignore{
  }

Eiter et al. \cite{eiter-etal-00,eiter-etal-04} propose a (propositional) logic
based planning language \m{\cal K} for planning with incomplete information as answer set programming. In this framework,  proposed originally by Lifschitz \cite{lifschitz-99}, a plan is the answer set of a logical program formulated using a specialized logical language. \m{\cal K} represents lack of knowledge using  negation as failure semantics. It supports both knowledge state and possible world planning. The authors further distinguish between optimistic and secure (i.e. conformant) planning. Optimistic plans may not be executable, due to their assumptions on the missing information. \m{\cal K} supports conditional effects, but does not allow any universal quantification on goals or state description.

\corr{2}{
Thielscher \cite{thielscher-05} presents FLUX - a logical programming 
framework for agent program design in the presence of incomplete information and sensing.
FLUX is based on fluent calculus and is implemented as a set of
constraints, defining the domain, action update, agent's knowledge and
action execution. The syntax of the language is carefully restricted
to provide linear time evaluation of the constraints.  The constraint
language includes universally quantified negated clauses, similar to
the simple \psiforms{} of \psiplan. However, unlike \psiplan{},
the constraint solver assumes a finite domain, and does not represent
exceptions to the universally quantified clauses.  FLUX has nice computational
properties, but it is not complete.  Also differently from \psiplan{}, the
 FLUX framework is designed for programming the intended
behavior of the agent via a designer-specified strategy, which defines
the set of agent control rules, rather than  the problem of
automatically constructing a sequence of actions that will result in
the achievement of the goal. }

Conformant Graphplan \cite{smith-weld-98}  and its extension to
planning with sensing, SGP \cite{weld-etal-98} are propositional
Graphplan  \cite{blum-furst-95} based open world planners that
consider every possible world and thus rely on the domain of objects being
sufficiently small. However, in small domains these planners are able of
generating remarkably long plans. Graphplan based planners perform a
search in a  
space of graphs generated by forward-chaining in the state space,
and their performance degrades when the initial state contains large
number of irrelevant atoms.  \ignore{ Nguyen and Kambhampati \cite{nguyen-kambhampati-01} use search control heuristics similar to those used in  
Graphplan to produce a partial order planning RePOP algorithm that 
is at least competitive with Graphplan and outperforms it in some domains.
}

CMBP planner \cite{cimatti-roveri-99} is a conformant planner based
on model checking. Like Conformant Graphplan it performs a
forward-chaining analysis, but relies on an effective way of
encoding sets of possible worlds and its performance is less
dependent on the amount of irrelevant information in the initial state.
CMBP uses action representation in the form of non-deterministic state
transition relations.

An approach to conformant planning as a heuristic search in the space of belief states that are sets of world states is presented  by Bonet and Geffner (\cite{bonet-geffner-00}). An admissible heuristic function is computed based on the distance to the goal state under the assumption of complete information. The search produces an optimal plan, however the algorithm relies on the finiteness of the state space, which is not achievable when the domain of objects is infinite.  The action language used is an extension  of STRIPS that includes function symbols, negation, disjunction, non-deterministic actions and conditional effects.

IPE \cite{ambros-ingerson-steel-88}, SENSE-P \cite{etzioni-etal-92},
XII\cite{golden-etal-94}, PUCCINI \cite{golden-thesis} are  causal
link planners that interleave planning with  execution of incomplete plans.
The action description language of PUCCINI, SADL \cite{golden-weld-96}
includes actions with conditional and informational effects. However,
to the best of our knowledge there are no completeness results for
conformant planning with SADL. 

PKS \cite{petrick-bacchus} is a forward chaining planner based on a
representation of the agent's knowledge that captures a set of
possible worlds via a set of knowledge formulas similarly to
\psiplan's SOK. The representation of 
Petrick and Baccus includes functional symbols, conditional plans and
actions with conditional effects, all of which are not represented in
\psiplan{} for the reason of tractability. However, the PKS planner only
admits ground literals in its goal language and does not handle
universally quantified negated goals. PKS is also incomplete, but
the authors report that it is able to generate plans in many
domains.

\ignore{
Petrick and Bacchus have also used the LCW mechanism in their representation  
(\cite{bacchus-petrick,petrick-bacchus}) for open world planning with sensing and execution that distingushes between planning and execution-time effects. The inference procedure IA presented in \cite{bachus-petrick} is designed for only atomic queries and is incomplete, however the authors claim that this does not preclude their PKS planner from generating solution plans. }

\edit{A(01),B(15)}{
The LCW \cite{etzioni-etal-97}  (see introduction) language is  extended in
\cite{levy-96,friedman-weld-97} to handling exceptions. Levy in  
\cite{levy-96}
uses extended LCW sentences, called Local Completeness (LC) sentences,  to
represent database 
completeness information and  derive answer completeness
property of a conjunctive query. This is analogous to computing
whether a SOK entails a universally quantified knowledge goal, where the SOK is
given by the combination of relational tables and the knowledge goal 
is to know all individual objects that satisfy a given query.
Friedman and Weld \cite{friedman-weld-97} extend on Levy's work 
for the purpose of eliminating redundant information gathering from
databases by an Internet agent. They present a method of determining
subsumption from LC sentences: whether a set of relational tables
contains all information available in another relational table.
Thus, both of these works only consider the setting in which there are no 
actions that can change the world, do not address a changing
world or planning, and do 
not present any methods that would make these extensions amenable to
their use in an open world planning algorithm, such as the image and
e-difference operations of \psiplan{} that are critical in the
computation of state update after an action, causal links and threat
resolution.  
}

\subsection{Complexity}\label{sec:rel-complexity}
Complexity of  propositional planning with incomplete information, with
and without sensing actions, has been addressed by many researchers (e.g.
\cite{haslum-jonsson-99}, \cite{eiter-etal-04}, \cite{baral-etal-00},
\cite{turner-02}). Results presented in these works, although
for different state and action languages, generally show that the complexity of
constructing conformant plans of polynomial length is greater than
planning with complete information, which is NP-complete. 
Though we have not proven the following formally, from the
results of this paper it appears that: (a) checking whether
a given plan (of polynomial length) solves a given problem in \psiplan{}
can be done in polynomial time and, thus, 
(b) determining whether there exists a plan (of polynomial length) that
solves a given problem in \psiplan{} is NP-complete.
These results do not contradict the results of Baral et al., nor those of
Turner \cite{turner-02}, as we explain below; the key to the reduced
complexity of \psiplan{}-based planning compared to the analysis in these
papers seems to be the absence of conditional effects.

Baral et al. \cite{baral-etal-00} present complexity results for a
variety of problems related to open world planning with  action
language \m{\cal A}. In particular, they show the problem of finding all
solution plans in presence of incomplete information and no sensing
belongs to the class \m{\Sigma_2P}. To keep the complexity of 
planning with incomplete 
information within the NP-completeness bounds, they propose a
0-approximation, which sacrifices completeness. 

In \psiplan{} as in 0-approximation of Baral et al., the set of possible worlds
is represented by a set of propositions that are known to be true.  However,
unlike the action language A used in Baral et al.'s work, \psiplan's action
language does not allow for conditional effects, and so all of an action's
effects are {\em guaranteed} to be true after the (executable) action
is performed. 
In contrast, in action language A determining the effect of the
action and thus the 
resulting set of possible worlds sometimes requires an analysis of possible
values of unknown propositions. 0-approximation does not involve such
analysis and thus loses such plans.

For example having no information at all and an action \m{a_0} with conditional
effect "\m{a_0} causes p if \m{\neg{}p}", a plan that consists of a
single action \m{a_0} 
achieves \m{p}, but it will be missed by 0-approximation. Without the
analysis of 
the result of performing \m{a_0} in two possible initial states
(corresponding to the two different values of p), it is impossible to
conclude that \m{p} is true after performing \m{a_0}. That is the
reason why 0-approximation will miss it. 
In \psiplan{}, there are no conditional effects, and so action \m{a_0}
from above 
cannot be represented.  Once executability of an action is determined, all
effects are guaranteed and the set of possible worlds is precisely described
by the single updated state of knowledge.  Thus, in \psiplan{} completeness of
conformant planning is preserved without an increase of complexity over
classical planning.

Turner  \cite{turner-02} presents a comprehensive complexity analysis
of a set of planning problems by using a very general framework for
describing  a planning problem. This framework represents actions as
state transition relations  and  integrates many action languages
including those with conditional effects, nondeterminism and concurrency.
As in Baral et al. his results on conformant planning consider actions
that may have conditional effects, and are more general than
\psiplan's. 

Haslum and Jonsson's paper \cite{haslum-jonsson-99} states a
PSPACE-completeness result for the problem of verifying existence of a
conformant plan of unbounded length with STRIPS-like actions. This
result is presented without proof and thus it is difficult to analyze
it for the case of polynomially bounded-length plans.

\section{Conclusions}\label{sec:conclusions}

Classical planning presupposes that complete and correct information
about the world is available at any point of planning (by having a
completely specified initial situation, and deterministic
actions). However, in a more realistic setting, the knowledge about
the initial state may be incomplete, the effects of actions may be
nondeterministic, or there may be other agents acting in the
world. These are some sources of uncertainty in planning.

In this paper we dropped one of the assumptions of classical
planning---the assumption of complete knowledge of the initial state
of the world--- thereby  considering the problem of open world planning.  We
have presented \psiplan{}, a language for representing and reasoning
in open world applications.
\psiplan{} uses \psiforms{} to represent  infinite sets of
clauses of negated literals. We have shown the following. \begin{itemize}

\item Our algorithm to determine entailment in \psiplan{} is sound
and complete and has polynomial complexity in the number of propositions in the state of knowledge under certain assumptions on the structure of \psiforms{} common for open world planning problems. Operations \emph{image} and \emph{e-difference} between \psiforms{}, which are crucial to planning with quantified propositions,  also have  polynomial complexity.

\item Updating the agent's state of knowledge after performing an
action has polynomial complexity in the number of propositions in the
state of the agent's knowledge.  In addition, the update procedure
correctly and completely describes the transition between possible
worlds due to the action.

\end{itemize}
Thus, \psiplan{} representation efficiently handles domains with an incomplete specification of the initial state without considering the set of all possible worlds, and does not require that the agent know  the set of all objects. 
\corr{1.13}{
We implemented a partial order planning algorithm  \psipop{} \cite{babaian-schmolze-00} for open worlds that uses \psiplan{} representation of state and actions. \psipop{} uses \psiplan{} calculus for reasoning about goal achievement.}
Since all of the \psiplan{} operations
used by \psipop{} have only polynomial complexity, we argue informally
that planning with \psiplan{} does not exceed the complexity of closed
world STRIPS style planning. \psigraph{} \cite{carlin-etal} is a GraphPlan-based planning algorithm which uses \psiplan{}.

Further evidence of the applicability of \psiplan{} representation for
planning in open worlds with a large or infinite number of objects,  
is the use of  \psipop{}'s extension to planning 
with sensing and interleaved execution at the core of the Writer's Aid \cite{waid-iui} -- a collaborative bibliography assistant.
Completeness and tractability of \psiplan's reasoning and its ability
to effectively handle initial information and goal statements
universally quantified over an infinite domain of objects ensured
effective and non-redundant operation of the system, which were
critically important in this application.

\corr{1.5}{
Experimental results from the initial implementation of \psipop{}, as well as 
an experimental assessment of the impact of various parameters of \psiforms{} on the performance of the entailment algorithm are presented in \cite{babaian-schmolze-TR}. Results from the initial implementation of \psigraph{} planner are presented in \cite{carlin-etal}. We are currently working on optimizing the performance of these two planners. A thorough experimental evaluation of \psipop{} and \psigraph{} is under way and we are planning on reporting it in a future paper.}

In the future, we will extend \psiplan{} to allow function symbols,
and we will publish already completed work that extends \psiplan{} 
to reasoning about the agent's knowledge goals and the effects of sensing
actions.

\section{Acknowledgements} 
We thank Barbara Grosz, Wendy Lucas, Wheeler Ruml   and the anonymous referees for their helpful comments on an earlier draft of this paper. 

\bibliographystyle{klunum} 
\bibliography{longtitles,planning-bibliography,all}

\begin{thebibliography}{00}

\bibitem{ambros-ingerson-steel-88}
Ambros-Ingerson, J.~A. and S. Steel: 1988, `Integrating Planning, Execution and
  Monitoring'.
\newblock In: {\em Proceedings of the Seventh National Conference on Artificial
  Intelligence (AAAI-88)}. St.~Paul, Minnesota, pp. 83--88.

\bibitem{babaian-thesis}
Babaian, T.: 2000, `Knowledge Representation and Open World Planning Using
  $\psi$-forms'.
\newblock Ph.D. thesis, Tufts University, Medford, MA.

\bibitem{waid-iui}
Babaian, T., B.~J. Grosz, and S.~M. Shieber: 2002, `A Writer's Collablrative
  Assistant'.
\newblock In: {\em Proceedings of Intelligent User Interfaces'02}. pp. 7--14.

\bibitem{babaian-schmolze-TR}
Babaian, T. and J. Schmolze, `Efficient Open World Reasoning and Planning'.
\newblock Unpublished paper, available at
  http://cis.bentley.edy/tbabaian/papers/p1.ps.

\bibitem{babaian-schmolze-00}
Babaian, T. and J. Schmolze: 2000, `PSIPLAN: open world planning with
  $\psi$-forms'.
\newblock In: {\em Artificial Intelligence Planning and Scheduling: Proceedings
  of the Fifth International Conference (AIPS'00)}. pp. 292--300.

\bibitem{baral-etal-00}
Baral, C., V. Kreinovich, and R. Trejo: 2000, `Computational complexity of
  planning and approximate planning in the presence of incompleteness'.
\newblock {\em Artificial Intelligence} {\bf 122}(1-2), 241--267.

\bibitem{baral-son-97}
Baral, C. and T.~C. Son: 1997, `Approximate Reasoning about Actions in Presence
  of Sensing and Incomplete Information'.
\newblock In: {\em International Logic Programming Symposium}. pp. 387--401.

\bibitem{ecp-99}
Biundo, S. and M. Fox (eds.): 1999, `Proceedings of the 5-th European
  Conference on Planning'.
\newblock Durham, England:, Springer-Verlag.

\bibitem{blum-furst-95}
Blum, A.~L. and M.~L. Furst: 1995, `Fast Planning Through Planning Graph
  Analysis'.
\newblock In: {\em Proceedings of the Fourteenth International Joint Conference
  on Artificial Intelligence (IJCAI-95)}. Nagoya, Japan, pp. 1636--1642.

\bibitem{bonet-geffner-00}
Bonet, B. and H. Geffner: 2000, `Planning with incomplete information as a
  heuristic search in belief space.'.
\newblock In: {\em Artificial Intelligence Planning and Scheduling: Proceedings
  of the Fifth International Conference (AIPS'00)}. Breckenridge, Colorado.

\bibitem{brafman-hoffman-04}
Brafman, R.~I. and J. Hoffmann: 2004, `Conformant Planning via Heuristic
  Forward Search: A New Approach.'.
\newblock In: {\em ICAPS}. pp. 355--364.

\bibitem{carlin-etal}
Carlin, A., J.~G. Schmolze, and T. Babaian: 2005, `Graphplan Based
  Conformant Planning with Limited Quantification'.
\newblock {\em Research on Computing Science.} {\bf 5} (Special Issue: Advances in Artificial Intelligence Theory), 65--75.

\bibitem{cimatti-roveri-99}
Cimatti, A. and M. Roveri: 1999, `Conformant Planning via Model Checking'.
\newblock in \cite{ecp-99}.

\bibitem{eiter-etal-00}
Eiter, T., W. Faber, N. Leone, G. Pfeifer, and A. Polleres: 2000, `Planning
  under Incomplete Knowledge'.
\newblock In: {\em Computational Logic}. pp. 807--821.

\bibitem{eiter-etal-04}
Eiter, T., W. Faber, N. Leone, G. Pfeifer, and A. Polleres: 2004, `A logic
  programming approach to knowledge-state planning: Semantics and complexity'.
\newblock {\em ACM Trans. Comput. Logic} {\bf 5}(2), 206--263.

\bibitem{etzioni-etal-97}
Etzioni, O., K. Golden, and D. Weld: 1997, `Sound and efficient closed-world
  reasoning for planning'.
\newblock {\em Artificial Intelligence} {\bf 89}(1--2), 113--148.

\bibitem{etzioni-etal-92}
Etzioni, O., S. Hanks, D. Weld, D. Draper, N. Lesh, and M. Williamson: 1992,
  `An Approach to Planning with Incomplete Information'.
\newblock In: {\em Proceedings of the Third International Conference on
  Principles of Knowledge Representation and Reasoning (KR-92)}. Cambridge, MA,
  pp. 115--125.

\bibitem{reiter-00}
Finzi, A., F. Pirri, and R. Reiter: 2000, `Open World Planning in the Situation
  Calculus'.
\newblock In: {\em Proceedings of the Seventeenth National Conference on
  Artificial Intelligence (AAAI-00)}. Austin, Texas.

\bibitem{friedman-weld-97}
Friedman, M. and D.~S. Weld: 1997, `{Efficiently Executing
  Information-Gathering Plans}'.
\newblock In: {\em Proceedings of the Fifthteenth International Joint
  Conference on Artificial Intelligence (IJCAI-97)}.

\bibitem{genesereth-nilsson}
Genesereth, M.~R. and N. Nilsson: 1986, {\em {Logical Foundations of Artificial
  Intelligence}}.
\newblock Morgan-Kaufmann Publishers.

\bibitem{golden-thesis}
Golden, K.: 1997, `Planning and knowledge representation for Softbots.'.
\newblock Ph.D. thesis, University of Washington.

\bibitem{golden-etal-94}
Golden, K., O. Etzioni, and D. Weld: 1994, `Omnipotence Without Omniscience:
  Efficient Sensor Management for Planning'.
\newblock In: {\em Proceedings of AAAI-94}.

\bibitem{golden-weld-96}
Golden, K. and D. Weld: 1996, `Representing Sensing Actions: The Middle Ground
  Revisited'.
\newblock In: {\em Proceedings of the Fifth International Conference on
  Principles of Knowledge Representation and Reasoning (KR-96)}. Cambridge, MA,
  pp. 174--185.

\bibitem{haslum-jonsson-99}
Haslum, P. and P. Jonsson: 1999, `Some results on the complexity of planning
  with incomplete information.'.
\newblock in \cite{ecp-99}.

\bibitem{krebsbach-etal-92}
Krebsbach, K., D. Olawsky, and M. Gini: 1992, `An Empiracal Study of Sensing
  and Defaulting in Planning'.
\newblock In: {\em Artificial Intelligence Planning Systems: Proceedings of the
  First International Conference (AIPS-92)}. College Park, MD, pp. 136--144.

\bibitem{levesque-96}
Levesque, H.: 1996, `What is planning in the presence of sensing?'.
\newblock In: {\em Proceedings of the Thirteenth National Conference on
  Artificial Intelligence (AAAI-96)}.

\bibitem{golog}
Levesque, H.~J., R. Reiter, Y. Lesperance, F. Lin, and R.~B. Scherl: 1997,
  `{GOLOG}: A Logic Programming Language for Dynamic Domains'.
\newblock {\em Journal of Logic Programming} {\bf 31}(1-3), 59--83.

\bibitem{levy-96}
Levy, A.: 1996, `{Obtaining Complete Answers from Incomplete Databases}'.
\newblock In: {\em Proceedings of the 22nd VLDB Conference}. Mumbai (Bombay),
  India.

\bibitem{lifschitz-99}
Lifschitz, V.: 1999, `Answer Set Planning.'.
\newblock In: {\em ICLP}. pp. 23--37.

\bibitem{liu-levesque-05}
Liu, Y. and H.~J. Levesque: 2005, `Tractable reasoning with incomplete
  first-order knowledge in dynamic systems with context-dependent actions'.
\newblock In: {\em IJCAI}. pp. 639--644.

\bibitem{mcallester-rosenblitt-91}
McAllester, D. and D. Rosenblitt: 1991, `Systematic Nonlinear Planning'.
\newblock In: {\em Proceedings of the Ninth National Conference on Artificial
  Intelligence (AAAI-91)}. Anaheim, California, pp. 634--639.

\bibitem{peot-smith-92}
Peot, M.~A. and D.~E. Smith: 1992, `Conditional Nonlinear Planning'.
\newblock In: {\em Artificial Intelligence Planning Systems: Proceedings of the
  First International Conference (AIPS-92)}. College Park, MD, pp. 189--197.

\bibitem{petrick-bacchus}
Petrick, R. and F. Bacchus: 2002, `A Knowledge-Based Approach to Planning with
  Incomplete Information and Sensing'.
\newblock In: {\em AI Planning and Scheduling (AIPS2002)}. pp. 212--222.

\bibitem{scherl-levesque-93}
Scherl, R.~B. and H.~J. Levesque: 1993, `{The Frame Problem and
  Knowledge-Producing Actions}'.
\newblock In: {\em Proceedings of the Eleventh National Conference on
  Artificial Intelligence (AAAI-93)}.

\bibitem{shirazi-amir-05}
Shirazi, A. and E. Amir: 2005, `First Order Logical Filtering'.
\newblock In: {\em IJCAI}. pp. 589--595.

\bibitem{smith-weld-98}
Smith, D. and D.~S. Weld: 1998, `Conformant {Graphplan}'.
\newblock In: {\em Proceedings of the Fifteenth National Conference on
  Artificial Intelligence (AAAI-98)}.

\bibitem{thielscher-05}
Thielscher, M.: 2005, `{FLUX}: A logic programming method for reasoning
  agents'.
\newblock {\em Theory and Practice of Logic Programming} {\bf 5}(4-5),
  533--565.

\bibitem{turner-02}
Turner, H.: 2002, `Polynomial-length planning spans the polynomial hierarchy'.
\newblock In: {\em Proceedings of Eighth European Conf. on Logics in Artificial
  Intelligence (JELIA'02)}.

\bibitem{weld-99}
Weld, D.~S.: 1999, `Recent Advances in {AI} Planning'.
\newblock {\em AI Magazine} {\bf 20}(2), 93--123.

\bibitem{weld-etal-98}
Weld, D.~S., C.~R. Anderson, and D.~E. Smith: 1998, `Extending {Graphplan} to
  Handle Uncertainty and Sensing Actions'.
\newblock In: {\em Proceedings of the Fifteenth National Conference on
  Artificial Intelligence (AAAI-98)}. Madison, Wisconsin.

\bibitem{winslett-88}
Winslett, M.: 1988, `Reasoning about action using a possible models approach'.
\newblock In: {\em Proceedings of the Seventh National Conference on Artificial
  Intelligence (AAAI-88)}.

\end{thebibliography}


\end{document}